\documentclass[conference,compsoc]{IEEEtran}
\ifCLASSOPTIONcompsoc
  \usepackage[nocompress]{cite}
\else
  \usepackage{cite}
\fi
\usepackage{tikz}
\usepackage{amsmath}
\usepackage{multirow}
\usepackage{booktabs}
\usepackage{subcaption}
\usepackage{amssymb}
\usepackage{makecell}
\usepackage{wasysym}
\usepackage{color}
\usepackage{listings}
\usepackage{url}
\usepackage{threeparttable}
\newcommand{\upp}[1]{\color[rgb]{0.117, 0.447, 0.999}#1}
\newcommand{\dow}[1]{\color[rgb]{0.753,0,0}#1}
\newcommand{\inc}[1]{\upp\mathbf{\blacktriangle #1\%}}
\newcommand{\dec}[1]{\dow\mathbf{\blacktriangledown #1\%}}
\lstdefinestyle{codestyle}{
    basicstyle=\ttfamily\footnotesize,
    numbers=left,
    numberstyle=\tiny\color{gray},
    stepnumber=1,
    numbersep=5pt,
    backgroundcolor=\color{black!5},
    keywordstyle=\color{blue},
    commentstyle=\color{green!50!black},
    stringstyle=\color{red!70!black},
    breaklines=true,
    breakatwhitespace=true,
    tabsize=2,
    showstringspaces=false,
    frame=single,
    rulecolor=\color{black!40},
    captionpos=b
}
\usepackage{enumitem}
\usepackage{hyperref}
\usepackage{marvosym}
\ifCLASSINFOpdf
\else
\fi
\begin{document}
%
\title{Are LLM-Enhanced Graph Neural Networks Robust against Poisoning Attacks?}

\author{
\IEEEauthorblockN{Yuhang Ma\IEEEauthorrefmark{2}\IEEEauthorrefmark{1},
Jie Wang\IEEEauthorrefmark{2}\IEEEauthorrefmark{1},
and Zheng Yan\IEEEauthorrefmark{2}\IEEEauthorrefmark{3}\textsuperscript{\Letter}}
\IEEEauthorblockA{\IEEEauthorrefmark{2}State Key Laboratory of Integrated Services Networks, School of Cyber Engineering, Xidian University, China}
\IEEEauthorblockA{\IEEEauthorrefmark{3}Hangzhou Institute of Technology, Xidian University, China}
\IEEEauthorblockA{yhm01041@gmail.com, jwang1997@stu.xidian.edu.cn, zyan@xidian.edu.cn}
}


\maketitle
{
\makeatletter
\long\def\@makefntext#1{\parindent 1em #1}
\makeatother
\footnotetext{\IEEEauthorrefmark{1} Yuhang Ma and Jie Wang contributed equally to this work.}
\footnotetext{\textsuperscript{\Letter} Zheng Yan is the corresponding author.}
}

\begin{abstract}
Large Language Models (LLMs) have advanced Graph Neural Networks (GNNs) by enriching node representations with semantic features, giving rise to LLM-enhanced GNNs that achieve notable performance gains. However, the robustness of these models against poisoning attacks, which manipulate both graph structures and textual attributes during training, remains unexplored. To bridge this gap, we propose a robustness assessment framework that systematically evaluates LLM-enhanced GNNs under poisoning attacks. Our framework enables comprehensive evaluation across multiple dimensions. Specifically, we assess 24 victim models by combining eight LLM- or Language Model (LM)-based feature enhancers with three representative GNN backbones. To ensure diversity in attack coverage, we incorporate six structural poisoning attacks (both targeted and non-targeted) and three textual poisoning attacks operating at the character, word, and sentence levels. Furthermore, we employ four real-world datasets, including one released after the emergence of LLMs, to avoid potential ground truth leakage during LLM pretraining, thereby ensuring fair evaluation. Extensive experiments show that LLM-enhanced GNNs exhibit significantly higher accuracy and lower Relative Drop in Accuracy (RDA) than a shallow embedding-based baseline across various attack settings. Our in-depth analysis identifies key factors that contribute to this robustness, such as the effective encoding of structural and label information in node representations. Based on these insights, we outline future research directions from both offensive and defensive perspectives, and propose a new combined attack along with a graph purification defense. To support future research, we release the source code of our framework at~\url{https://github.com/CyberAlSec/LLMEGNNRP}.
\end{abstract}


%
\IEEEpeerreviewmaketitle

\section{Introduction}

A Graph Neural Network (GNN) is a type of neural network designed to learn from graph-structured data, that is, data made up of nodes (vertices) and edges (links or relationships). It learns low-dimensional node representations (i.e., embeddings) that capture both local network topology and node features~\cite{zhang2020gnnguard} to perform various downstream tasks, including node classification, link prediction, and graph classification~\cite{kipf2017semi,zhang2018link,xu2018powerful}. Text-Attributed Graphs (TAGs)~\cite{yao2019graph} are data structures that GNNs can operate on. TAGs integrate textual information with graph structures, naturally modeling many real-world networks such as citation networks and social networks. In TAGs, nodes correspond to entities (e.g., papers or users), while their textual content (e.g., abstracts or profiles) serves as node attributes. A standard GNN pipeline for processing TAGs first encodes node textual attributes using shallow embedding techniques, such as Bag-of-Words (BoW)~\cite{harris1954distributional} or skip-gram~\cite{mikolov2013distributed}. The resulting embeddings are then combined with the graph structure for GNN training. However, these shallow embeddings fail to capture the rich semantics present in lengthy and complex texts, leading to suboptimal performance in real-world applications such as citation analysis and social recommendation.

In recent years, Large Language Models (LLMs) have demonstrated remarkable capabilities in understanding and reasoning over complex texts~\cite{wu2024legilimens}. Building on this success, researchers have begun exploring whether LLMs can enhance the performance of GNNs, giving rise to a new paradigm: LLM-enhanced GNNs. In this paradigm, an LLM or smaller Language Model (LM) acts as a feature enhancer to generate semantically rich node embeddings, which are then processed by GNNs for downstream tasks. Depending on how LLMs/LMs are integrated into the GNN pipeline, three main integration approaches have been proposed: (1) LLM-Explanation~\cite{he2024harnessing,chen2024exploring}, which leverages LLMs to extract key information from raw texts and then convert them into embeddings using LMs. (2) LLM-Embedding~\cite{liu2023one,zhu2024efficient}, which directly employs LLMs to generate embeddings from textual inputs. (3) LM-Embedding~\cite{chien2021node,duan2023simteg}, which uses smaller LMs for embedding generation. By leveraging the reasoning capabilities and extensive knowledge bases of LLMs/LMs, these approaches extract deep semantic features from textual attributes, significantly improving node embedding quality and GNN performance~\cite{li2024glbench}. Overall, this integration extends the applicability of GNNs to complex real-world scenarios that require both relational reasoning and textual understanding.

Despite the promising performance of LLM-enhanced GNNs in various graph learning tasks, their robustness against poisoning attacks remains largely unexplored. Poisoning attacks pose a serious threat because they can drastically degrade model performance by manipulating training data~\cite{kumar2020adversarial,tang2020transferring}. In LLM-enhanced GNNs, this threat could be amplified since the attack surface spans both graph structures and node textual attributes, making the model vulnerable. While a few recent studies~\cite{guo2024learning,zhang2024can} have made initial attempts to evaluate how LLMs influence GNN robustness, they do not adequately address poisoning attacks and have the following limitations:

\textbf{L1: Incomplete coverage of victim models and attack types.} Prior studies consider only a limited number of LLMs/LMs and GNN backbones, and restrict their evaluation to one or two attack types. Broader coverage is needed, as different LLMs/LMs and GNN backbones may exhibit varying levels of robustness, and structural and textual poisoning exploit complementary vulnerabilities. For instance, structural attacks manipulate graph topology in both targeted and non-targeted manners, while textual attacks corrupt node attributes at the character, word, and sentence levels. These diverse attack types may have varying impacts on model behavior. Without evaluating a wide range of models and attack types, it is difficult to obtain a comprehensive understanding of model robustness.

\textbf{L2: Lack of post-LLM dataset evaluation.} Existing studies evaluate robustness only on traditional graph benchmarks, without using datasets released after the emergence of LLMs. This is problematic because many commonly used datasets (e.g., Cora~\cite{mccallum2000automating}, Pubmed~\cite{sen2008collective}, and Ogbn-products~\cite{he2024harnessing}) are likely included in LLM/LM pretraining corpora, leading to potential ground truth leakage. Evaluation on such datasets may overestimate actual robustness and make it difficult to determine whether the observed improvements come from ground truth leakage or from the architectural integration of LLMs/LMs.

\textbf{L3: Insufficient discussion of future research directions.} Prior studies provide little guidance on future directions from both offensive and defensive perspectives. Advancing research in this field requires deep understanding of how attacks and defenses can co-evolve, including how adversaries can jointly exploit structural and textual vulnerabilities, and how robust defenses can be developed for emerging LLM-enhanced GNN models.

To address these limitations, in this paper, we propose a comprehensive robustness assessment framework for evaluating LLM-enhanced GNNs against poisoning attacks. 
\textbf{To address L1}, we integrate eight representative LLM/LM-based feature enhancers~\cite{he2024harnessing,chen2024exploring,touvron2023llama,openai-embeddings-api,LinqAIResearch2024,duan2023simteg,wang2022text,warner2024smarter} with three classic GNN backbones (i.e., GCN~\cite{kipf2017semi}, GAT~\cite{velivckovic2017graph}, and GraphSAGE~\cite{hamilton2017inductive}), yielding a total of 24 victim model combinations. These combinations span the three dominant integration approaches in the literature: LLM-Explanation, LLM-Embedding, and LM-Embedding. To ensure attack diversity, we employ both targeted~\cite{zugner2018adversarial,li2021adversarial,zhao2023black} and non-targeted~\cite{zugner_adversarial_2019,waniek2018hiding,zhu2024simple} structural poisoning attacks that perturb graph structures, as well as textual poisoning attacks that modify node textual attributes at multiple granularities, including character~\cite{gao2018black}, word~\cite{li2020bert}, and sentence~\cite{chen2021multi} levels.
\textbf{To address L2}, we include the up-to-date Tape-arxiv23~\cite{he2024harnessing} dataset, which was released after many LLMs/LMs and is therefore unlikely to be part of their pretraining corpora. This mitigates the risk of ground truth leakage and supports a fair robustness evaluation.
\textbf{For L3}, we discuss future research directions from the perspectives of both adversaries and defenders. In addition, we propose a new attack by combining structural and textual attacks, and design a graph purification defense based on LLM/LM-generated embeddings.

Specifically, the proposed robustness assessment framework operates as follows. We first prepare four real-world TAG datasets collected from citation networks and e-commerce platforms, including Cora~\cite{mccallum2000automating}, Pubmed~\cite{sen2008collective}, Ogbn-products~\cite{he2024harnessing}, and a newly released post-LLM dataset, Tape-arxiv23~\cite{he2024harnessing}. Then, we apply different types of structural and textual poisoning attacks to perturb graph structures and textual attributes, respectively. These attacks are conducted under a gray-box setting, where an adversary has access to input data but lacks knowledge about victim models, making it more practical than a white-box setting. We train the victim models using the perturbed data to obtain poisoned models. Finally, we assess robustness by comparing the performance of poisoned and clean models on test datasets using accuracy and Relative Drop in Accuracy (RDA), followed by an analysis of key factors that contribute to model robustness or vulnerability.

Our extensive evaluation and in-depth analysis yield several findings: 
(1) LLM-enhanced GNNs exhibit significantly stronger robustness against both structural and textual poisoning attacks compared to a shallow embedding-based baseline. This robustness improvement is particularly evident under high perturbation rates.
(2) The robustness gains of LLM-enhanced models are not systematically associated with ground truth leakage but primarily arise from the architectural integration of LLMs/LMs.
(3) LLM/LM-based feature enhancers produce high-quality embeddings that encode label and structural information more effectively than the baseline, thus enhancing robustness against poisoning attacks.
(4) The proposed combined attack that considers both structural and textual perturbations is more effective than single-type attacks. Furthermore, our graph purification defense leveraging LLM/LM-generated embeddings substantially enhances robustness and, in some cases, even surpasses performance on clean graphs.

To summarize, our main contributions are as follows:
\begin{itemize}
\item \textbf{New evaluation perspective.} We develop a robustness assessment framework to evaluate LLM-enhanced GNNs against poisoning attacks. To the best of our knowledge, this framework provides the most comprehensive coverage of victim models and attack types to date, and incorporates post-LLM dataset evaluation.
\item \textbf{Comprehensive evaluation and robustness insights.} We systematically evaluate three classes of LLM-enhanced GNNs under six structural and three textual poisoning attacks on four real-world TAG datasets. We further conduct fine-grained analyses of attack results to uncover key factors contributing to model robustness or vulnerability.
\item \textbf{Future research directions.} We discuss future directions from both offensive and defensive perspectives, and propose a new combined attack along with a graph purification defense.
\item \textbf{Open-source framework.} We release our framework as an open-source toolkit to facilitate future research on the robustness evaluation of LLM-enhanced GNNs against poisoning attacks.
\end{itemize}

\section{Background and Related Work}
In this section, we introduce the basics of TAGs, GNNs, LLMs/LMs, and adversarial attacks on graphs and texts. Furthermore, we briefly review recent advances in LLM-enhanced GNNs.

\subsection{Text-Attributed Graphs}
A TAG, denoted as $\mathcal G=(\mathcal{V}, \mathcal{E}, \mathcal{S})$, 
is a graph in which each node is associated with textual attributes. Here, $\mathcal{V}=\{v_1,\ldots,v_N\}$ denotes a node set, $\mathcal{E}=\{e_1,\ldots,e_M\}$ is an edge set, and $\mathcal{S}=\{s_1,\ldots,s_N\}$ represents the set of node-associated textual attributes. The adjacency matrix of the graph is given by $\mathbf{A}\in \mathbb{R}^{N \times N}$, where $\mathbf{A}_{ij} \in \{0,1\}$ indicates the existence of an edge between $v_i$ and $v_j$. TAGs integrate structured topological relations with unstructured textual semantics, effectively modeling real-world networks such as social, citation, and protein interaction networks. 

\subsection{Graph Neural Networks} 
GNNs learn meaningful node representations (i.e., embeddings) through recursive message passing and aggregation guided by graph structures. They have demonstrated superior performance on various tasks, including node classification, link prediction, and graph classification~\cite{kipf2017semi,zhang2018link,xu2018powerful}.
Below, we briefly introduce three representative GNNs. Graph Convolutional Network (GCN)\cite{kipf2017semi} extends convolution to non-Euclidean graphs by aggregating normalized neighbor features. Graph Attention Network (GAT)~\cite{velivckovic2017graph} employs an attention mechanism to assign learnable weights to neighbors, allowing the model to focus on key information during aggregation. GraphSAGE~\cite{hamilton2017inductive} achieves scalability and inductive learning by sampling fixed-size neighborhoods and applying learnable aggregators.

In this work, we focus on node classification since it is the most extensively studied task in both clean and adversarial settings~\cite{sun2022adversarial}, serving as the standard benchmark for evaluating GNN robustness. Moreover, robustness evaluation results obtained from node classification naturally generalize to other graph learning tasks, as they all rely on the same principles of message passing and feature aggregation~\cite{zhou2020graph}.
Formally, in node classification, each node $v_i \in \mathcal{V}$ is associated with a ground-truth label $y_i$ indicating its category. The textual attributes in $\mathcal{S}$ are encoded into a node feature matrix $\mathbf{X} \in \mathbb{R}^{N \times d}$ (e.g., using TF-IDF~\cite{salton1988term} or Bag-of-Words (BoW)~\cite{harris1954distributional}), where the $i$-th row $\mathbf{x}_i \in \mathbb{R}^d$ represents the feature vector of node $v_i$. Given a labeled training subset $\mathcal{V}_L\subset\mathcal{V}$ and its unlabeled complement $\mathcal{V}_U=\mathcal{V}\setminus\mathcal{V}_L$, the goal is to train a GNN model $f_{\text{GNN}}(\mathbf{A}, \mathbf{X})$ that accurately predicts the labels of nodes in $\mathcal{V}_U$.

\subsection{Graph Adversarial Attacks}
\label{graph_adversarial}
Graph adversarial attacks intentionally manipulate graph-structured data to compromise the performance, integrity, or behavior of GNN models, either by corrupting the training process through poisoning attacks or by generating malicious inputs during inference through evasion attacks. In this work, we focus on poisoning attacks, as they are more damaging than evasion attacks and can severely degrade model performance~\cite{zhang2024can}. 
Poisoning attacks may target node features or graph structures. The latter, referred to as structural poisoning attacks, has proven more effective than the former and thus becomes the primary focus of this work~\cite{jin2020graph,geisler2021robustness}. 
In structural poisoning attacks, an adversary modifies the graph structure, such as by adding or removing edges within a limited attack budget. Based on attack specificity, structural poisoning attacks can be categorized into two types. Targeted attacks aim to manipulate the edges connected to a specific node so that the model produces an incorrect prediction for that node. In contrast, non-targeted attacks attempt to degrade the overall performance of the model across the entire test set.

\subsection{LLMs and LMs} 
LLMs and LMs are closely related but differ in scale and usage. Following~\cite{li2024survey}, we define LLMs as large-scale language models with billions of parameters, pretrained on massive text corpora, such as LLaMA~\cite{touvron2023llama} and GPT-4~\cite{achiam2023gpt}. In contrast, LMs refer to those earlier pretrained models with moderate parameter sizes, typically in the millions, such as BERT~\cite{koroteev2021bert} and RoBERTa~\cite{liu2019roberta}. While LMs can be easily fine-tuned on task-specific data, fine-tuning LLMs is challenging due to their huge parameter sizes. Instead, LLMs are typically adapted to new tasks through prompting. Owing to their in-context learning capabilities, they can generate task-specific outputs without parameter updates. \textit{Given the relatively small size of GNNs, both LLMs and LMs can be regarded as large-scale language models.} For simplicity, we refer all LLM/LM-enhanced GNNs to LLM-enhanced GNNs. 
In the context of TAGs, LLMs/LMs can be used to encode textual attributes, formulated as $\mathbf{x}_{i} = f_{\text{LLM/LM}}(s_{i})$, where $s_{i}$ is the textual attribute of node $v_i$, and $f_{\text{LLM/LM}}$ is an LLM/LM. The model $f_\text{LLM/LM}$ encodes $s_{i}$ into a $d$-dimensional vector $\mathbf{x}_{i} \in \mathbb{R}^d$.

\subsection{Textual Adversarial Attacks} 
Textual adversarial attacks aim to mislead language models by intentionally modifying textual data. Depending on perturbation granularity, they can occur at the character level (e.g., inserting typos, replacing or swapping characters)~\cite{gao2018black}, word level (e.g., substituting words with synonyms, homophones, or misspellings)~\cite{li2020bert}, and sentence level (e.g., reordering phrases, paraphrasing, or inserting distracting contents)~\cite{chen2021multi}. Based on the attack stage, textual attacks can also be classified into evasion and poisoning attacks. In the context of LLMs and LMs, which are pretrained and contain a large number of parameters, evasion attacks are more practical than poisoning attacks. Therefore, in this paper, we do not directly poison LLMs/LMs. Instead, we launch evasion attacks against them by generating perturbed textual attributes, which are then used to train LLM-enhanced GNNs. This approach reduces the cost of attacks while maintaining their effectiveness.

\subsection{LLM-Enhanced GNNs} \label{background_llm_gnn}
LLMs can encode textual attributes into semantically rich embeddings, which are then combined with graph structures and processed by GNNs. Depending on whether the LLM produces additional textual information, LLM-enhanced GNNs fall into explanation-based (i.e., LLM-Explanation) and embedding-based approaches. Embedding-based approaches can be further divided into LLM-Embedding and LM-Embedding methods, depending on whether the embeddings are generated by an LLM or a smaller LM. In this paper, we systematically investigate the robustness of these three types of LLM-enhanced GNNs against poisoning attacks.

\textbf{LLM-Explanation:} This type of method leverages the reasoning capabilities and knowledge of LLMs to extract high-level semantic information from raw textual attributes. The extracted information is then converted into embeddings by an LM. Formally, $\mathbf{x}_{i} = f_{\text{LM}}(f_{\text{LLM}}(p, s_{i}),s_{i})$, where $f_{\text{LLM}}$ generates augmented texts (e.g., explanations) by receiving a carefully-designed prompt $p$ and the original text $s_i$. The augmented texts are then processed by $f_{\text{LM}}$ to produce enhanced embedding $\mathbf{x}_{i}$ for GNN training. 
For instance, TAPE~\cite{he2024harnessing} employs GPT-3.5~\cite{brown2020language} to generate explanations and pseudo-labels, and fine-tunes DeBERTa-base~\cite{he2020deberta} on both the original and augmented texts to enrich node embeddings. Similarly, KEA~\cite{chen2024exploring} extracts knowledge entities with GPT-3.5 and encodes them using several LMs.

\textbf{LLM-Embedding:} This type of method directly uses LLMs to encode textual attributes into embeddings, which serve as initial node embeddings for GNN training. Formally, it is expressed as $\mathbf{x}_{i} = f_{\text{LLM}}(s_{i})$. LLM-Embedding methods rely on embedding-accessible LLMs, either open-source models with exposed embedding layers~\cite{LinqAIResearch2024} or API-based models that provide embeddings through an interface~\cite{openai-embeddings-api}. For example, OFA~\cite{liu2023one} employs LLaMA-2~\cite{touvron2023llama} to map node and edge descriptions into a shared, task-agnostic embedding space that can be reused across multiple downstream tasks. ENGINE~\cite{zhu2024efficient} uses a tunable adapter to incorporate structural and textual information into LLaMA-generated embeddings.

\textbf{LM-Embedding:} This type of method fine-tunes an LM with textual attributes and uses the resulting embeddings for GNN training. This process can be represented as $\mathbf{x}_{i} = f_{\text{LM}}(s_{i})$. For example, GIANT~\cite{chien2021node} fine-tunes BERT~\cite{koroteev2021bert} using an eXtreme Multi-label Classification (XMC) formalism, yielding embeddings that outperform BoW~\cite{harris1954distributional} and vanilla BERT~\cite{devlin2019bert} on node classification. SimTeG~\cite{duan2023simteg} improves embedding quality by applying Parameter-Efficient Fine-Tuning (PEFT) to the pretrained E5-Large model~\cite{wang2022text}.

\section{Threat Model}
\label{threat_model}
\textbf{Adversary's Goal.} We consider that an adversary seeks to perform poisoning attacks by manipulating input data, including node textual attributes and the graph structure, which are used to train a victim model (i.e., an LLM-enhanced GNN). The objective is either to reduce classification accuracy on a targeted subset of nodes or to degrade overall model performance. Such performance degradation can benefit the adversary by altering node classification outcomes in decision-critical applications. For instance, in e-commerce graphs, the adversary may cause fraudulent nodes to be classified as legitimate, thereby evading fraud detection for financial gain.

\textbf{Adversary's Knowledge.} Following prior works~\cite{zugner2018adversarial,li2021adversarial,zugner_adversarial_2019}, we adopt a gray-box setting, where the adversary has full access to the graph structure, node textual attributes, and the labels of a subset $\mathcal{V}_L$, but lacks knowledge of the victim model's architecture or parameters~\footnote{Note that some works also conduct experiments under the assumption of partial graph knowledge, which typically reduces attack effectiveness. Since our goal is to evaluate model robustness, we adopt the full-knowledge assumption to launch strong attacks.}. This setting reflects practical scenarios in domains such as citation networks, where citation links (i.e., graph structure), paper contents (i.e., node textual attributes), and research topics (i.e., node labels) are publicly available through platforms like Google Scholar, while the victim model is inaccessible due to intellectual property restrictions.

\textbf{Adversary's Capability.} We consider that the adversary can launch both structural and textual poisoning attacks. For structural attacks, the adversary perturbs the graph structure (e.g., adding or removing edges) using established GNN poisoning techniques~\cite{zugner2018adversarial,li2021adversarial,zhao2023black,zugner_adversarial_2019,waniek2018hiding,zhu2024simple}. For textual attacks, instead of directly poisoning LLMs or LMs, which is costly due to their large scales, the adversary applies textual evasion techniques~\cite{gao2018black,li2020bert,chen2021multi} to generate perturbed texts and then inject them into the training data. These perturbed texts preserve semantic consistency while embedding malicious patterns.

\begin{figure*}[tbp]
\centering
\includegraphics[width=\textwidth]{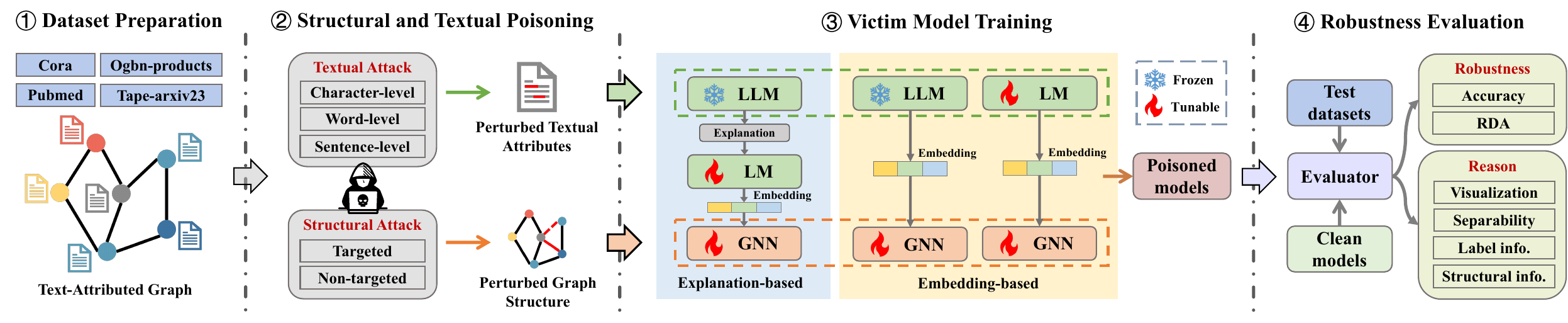}
\vspace{-4mm}
\caption{An overview of the proposed robustness assessment framework.}
\label{fig:workflow}
\end{figure*}

\section{Robustness Assessment Framework}
In this section, we first provide an overview of the proposed robustness assessment framework and then describe each component in detail.

\subsection{Overview}
As illustrated in Fig.~\ref{fig:workflow}, our framework consists of four main phases: dataset preparation, structural and textual poisoning, victim model training, and robustness evaluation. In the dataset preparation phase, we collect four real-world TAG datasets from citation networks and e-commerce platforms. Notably, one dataset (i.e., Tape-arxiv23) was released after the emergence of many LLMs/LMs, which helps mitigate potential ground truth leakage and ensures a fair robustness evaluation. In the structural and textual poisoning phase, we apply six structural attacks (three targeted and three non-targeted) to perturb graph structures, along with three textual attacks operating at the character, word, and sentence levels to modify node textual attributes. This diverse attack set enables a thorough robustness evaluation of LLM-enhanced GNN models. During the victim model training phase, we train models based on the perturbed graph structures or modified textual attributes to obtain poisoned models. Finally, in the robustness evaluation phase, we quantitatively assess model robustness by comparing the performance of poisoned and clean models on test datasets using two metrics: accuracy and Relative Drop in Accuracy (RDA). We further analyze key factors that affect robustness from multiple perspectives, including embedding visualization, embedding separability, and preservation of label and structural information.

\subsection{Dataset Preparation}
In this phase, we prepare datasets for subsequent poisoning. Four real-world TAG datasets are selected to ensure diversity in graph scales, domains, and node classes. The \textbf{Cora}~\cite{mccallum2000automating} dataset contains 2,708 scientific publications categorized into seven research areas, where each paper cites or is cited by at least one other publication. 
The \textbf{Pubmed}~\cite{sen2008collective} dataset contains 19,717 diabetes-related publications classified into three categories.
For the two large-scale datasets, \textbf{Ogbn-products}~\cite{he2024harnessing} and \textbf{Tape-arxiv23}~\cite{he2024harnessing}, we adopt a node sampling strategy~\cite{hamilton2017inductive} to obtain their subsets, enabling efficient execution of poisoning attacks.
The sampled Ogbn-products dataset is an Amazon product co-purchasing network with nodes representing products, edges indicating co-purchases, and a classification task over 47 top-level product categories.
The sampled Tape-arxiv23 dataset is a directed citation network of all computer science arXiv papers published in 2023 or later, where the task is to predict 40 subject areas. As this dataset extends beyond the knowledge cutoff of widely-used LLMs/LMs (e.g., GPT-3.5's cutoff is September 2021), it is unlikely to be included in their pretraining corpora. This temporal separation enables a fair robustness evaluation by mitigating potential ground truth leakage. Table~\ref{tab:datasets} summarizes the statistics of the four datasets, which collectively support a broad robustness evaluation of LLM-enhanced GNNs in realistic scenarios.

\begin{table}[!t]
\renewcommand\arraystretch{1.2}
\caption{Dataset descriptions.}
\label{tab:datasets}
\resizebox{\linewidth}{!}{
\centering
\begin{tabular}{@{}lcccccc@{}}
		\hline
		\textbf{Dataset}               & \textbf{\#Nodes} & \textbf{\#Edges} & \textbf{\#Classes} & \textbf{\#Features} & \textbf{Method} & \textbf{Domain} \\ \hline
		Cora                 & 2,708  & 5,429  & 7       & 1,433  & BoW     & Citation \\
		Pubmed               & 19,717 & 44,338 & 3       & 500   & TF-IDF  & Citation \\
        Ogbn-products (subset)& 12,011 & 21,987 & 47      & 100   & BoW     & E-commerce \\ 
		Tape-arxiv23 (subset)   & 13,167 & 23,735 & 40      & 300   & Word2Vec& Citation \\ 
        \hline
	\end{tabular}}
\end{table}

\subsection{Structural and Textual Poisoning}
LLM-enhanced GNNs take two types of inputs for training: graph structure and node textual attributes. Accordingly, in this phase, we consider both structural and textual poisoning attacks to perturb these inputs. The specific attack methods for each modality are described below.

\textbf{Structural attacks:} We choose six structural attacks that differ in attack specificity: 
\begin{itemize}
    \item \textbf{Targeted attacks} aim to mislead GNNs on specific nodes. Nettack~\cite{zugner2018adversarial} is a gradient-based gray-box poisoning attack that induces misclassification by introducing minimal and unnoticeable perturbations to the graph structure. SGA~\cite{li2021adversarial} leverages simplified gradient estimation on the local subgraph of a target node, combined with a multi-stage attack procedure, to efficiently generate targeted structural perturbations on large graphs. NAG-R~\cite{zhao2023black} applies Nesterov momentum to score edges in the target node's local subgraph, enabling budgeted and stealthy edge rewiring.
    \item \textbf{Non-targeted attacks} are designed to degrade the overall performance of GNNs. Mettack~\cite{zugner_adversarial_2019} is a meta-learning-based poisoning attack that perturbs the graph structure by optimizing edge insertions and deletions through meta-gradients. DICE~\cite{waniek2018hiding} is a simple label-based poisoning attack that degrades GNN performance by removing intra-class edges and adding inter-class edges, thereby reducing homophily and disrupting message aggregation. PGA~\cite{zhu2024simple} identifies vulnerable target nodes via a hierarchical selection policy, then selects cost-effective anchor nodes and performs greedy edge modifications to concentrate the attack budget on partial graphs.
\end{itemize}

\textbf{Textual attacks:} We choose three textual attacks that perturb texts at different granularities:
\begin{itemize}
    \item \textbf{Character-level attacks} perform fine-grained edits (e.g., insertion, deletion, substitution, and transposition) on characters to disrupt tokenization and model predictions. DeepWordBug~\cite{gao2018black} (in short DWord) uses greedy character-level operations to generate adversarial typos.
    \item \textbf{Word-level attacks} replace high-impact words with context-aware candidates (e.g., synonyms or suggestions from LMs), thereby balancing semantic preservation with attack effectiveness. Bert-Attack~\cite{li2020bert} (in short BertAtk) leverages masked language modeling to substitute or insert linguistically plausible words that alter the prediction of a victim model.
    \item \textbf{Sentence-level attacks} perform paraphrasing or sentence rewriting to induce substantial representation shifts while preserving human-level meaning. MAYA~\cite{chen2021multi} is a multi-granularity adversarial attack that generates perturbations at the constituent, sentence, and word levels. It iteratively selects paraphrases that maximize the victim model's prediction error.
\end{itemize}

We select the above attack methods since they are widely used baselines in the literature. These methods cover different granularities of adversarial manipulation and are efficient, reproducible, and practical, making them suitable for our systematic robustness evaluation.

\subsection{Victim Model Training} \label{victim_training}
In this phase, we train victim models using the perturbed graph structures or modified node textual attributes to obtain poisoned models. Each victim model comprises an LLM/LM-based feature enhancer and a GNN backbone. Following the review in Section~\ref{background_llm_gnn}, we select eight feature enhancers based on their methodological diversity, state-of-the-art performance, and reproducibility, spanning three types of LLM-GNN integration approaches.
\begin{itemize}
    \item \textbf{LLM-Explanation:} We choose TAPE~\cite{he2024harnessing} and KEA~\cite{chen2024exploring}, two pioneering methods that leverage GPT-3.5 to generate explanations or keyphrases, which are then encoded by LMs such as DeBERTa-base~\cite{he2020deberta} or E5-Large~\cite{wang2022text}. 
    \item \textbf{LLM-Embedding:} We adopt three LLMs for embedding generation, including LLaMA-2~\cite{touvron2023llama} (in short LLaMA, recognized for its efficient inference and extensive open-source ecosystem), OpenAI's text-embedding-3-large~\cite{openai-embeddings-api} (in short TE3L, chosen for its high-precision multilingual embeddings and strong benchmark performance), and Linq-Embed-Mistral~\cite{LinqAIResearch2024} (in short Linq, ranking highly in embedding quality through careful data curation). 
    \item \textbf{LM-Embedding:} We select SimTeG~\cite{duan2023simteg}, E5-Large~\cite{wang2022text}, and ModernBert~\cite{warner2024smarter}, which provide a strong quality-scalability balance for embedding generation.
\end{itemize}

The details of LLMs/LMs used in the selected feature enhancers are summarized in Table~\ref{tab:LLMs}. We integrate three classic GNN backbones, i.e., \textbf{GCN}~\cite{kipf2017semi}, \textbf{GAT}~\cite{velivckovic2017graph}, and \textbf{GraphSAGE}~\cite{hamilton2017inductive} (in short \textbf{SAGE}), with these enhancers, resulting in a total of $3 \times 8=24$ victim model combinations. While more advanced LLMs (e.g., GPT-4~\cite{achiam2023gpt}) and GNN architectures (e.g., GIN~\cite{xu2018how} and GAT-v2~\cite{brody2022attentive}) exist, we focus on model combinations that have been widely adopted and evaluated in prior work~\cite{he2024harnessing,chen2024exploring,duan2023simteg} to ensure reproducibility and fair comparison with established baselines. This broad coverage enables a thorough robustness evaluation.

\begin{table}[!tbp]
\renewcommand\arraystretch{1.2}
\caption{Number of parameters and vendors of the LLMs/LMs used in our experiments.}
\label{tab:LLMs}
\resizebox{\linewidth}{!}{
\centering
\begin{tabular}{@{}clcc@{}}
        \toprule
        \textbf{Model Type} & \textbf{LLMs/LMs} & \textbf{\#Parameters} & \textbf{Vendor} \\ 
        \midrule
        \multirow{2}{*}{API-based} 
         & GPT-3.5-Turbo & - & OpenAI \\
        & Text-embedding-3-large & - & OpenAI \\
        \midrule
        \multirow{5}{*}{Open-source} 
         & DeBERTa-base & 140M & Microsoft \\
         & E5-Large & 300M & Microsoft \\
         & ModernBert-large & 400M & Answer.AI \\
         & LLaMA-2-7b-hf & 7B & Meta \\
         & Linq-Embed-Mistral & 7B & LinqAlpha \\
        \bottomrule
\end{tabular}}
\end{table}

\subsection{Robustness Evaluation}
In this phase, we evaluate robustness by comparing the performance of poisoned and clean models on test datasets using two metrics:
\textbf{Accuracy (ACC)} measures the proportion of correctly classified samples on the test dataset. It reflects absolute robustness, i.e., the extent to which a model retains its predictive performance under attacks. Higher ACC indicates better robustness.
\textbf{Relative Drop in Accuracy (RDA)} quantifies the relative decrease in accuracy under attacks compared to the clean setting, i.e., $(ACC_{clean} - ACC_{attack})/ACC_{clean}$. It reflects relative robustness by measuring a model's sensitivity to attacks. Lower RDA indicates better robustness. These two metrics are complementary and assess robustness from different perspectives.
In addition, we analyze the underlying causes that contribute to the robustness or vulnerability of victim models from the following dimensions:

\textbf{Embedding Visualization.} To qualitatively evaluate the quality of embeddings generated by different LLM/LM-based feature enhancers, we visualize them using the t-SNE technique~\cite{maaten2008visualizing}.

\textbf{Embedding Separability.} To quantitatively evaluate embedding quality, we analyze how well they form distinct clusters (i.e., separability) using the Davies-Bouldin Index (DBI)~\cite{davies2009cluster} and the Silhouette Score~\cite{rousseeuw1987silhouettes}. DBI measures the average similarity between each cluster and its most similar cluster, with lower values indicating better clustering. In contrast, the Silhouette Score measures both intra-cluster cohesion and inter-cluster separation, with larger values indicating better clustering quality.

\textbf{Preservation of Label Information.}
We evaluate how well the embeddings preserve label information using embedding homophily and embedding-label mutual information~\cite{hjelm2018learning} (their formulas are given in Appendix~\ref{appendix_formula}). Specifically, embedding homophily quantifies the proportion of a node's $k$-nearest neighbors in the embedding space that share the same label, reflecting structural-label consistency. Higher homophily indicates that nodes are more likely to be surrounded by neighbors with the same label. Embedding-label mutual information quantifies the amount of information shared between embeddings and labels, with larger values implying stronger preservation of label semantics in the embeddings. 

\textbf{Preservation of Structural Information.}
We evaluate how well the embeddings preserve local graph structure using two metrics: embedding-structural mutual information~\cite{zhao2023deep} and neighbor consistency~\cite{wang2020unifying} (their formulas are given in Appendix~\ref{appendix_formula}). Embedding-structural mutual information measures the dependency between node embeddings and local structural properties, including degree, clustering coefficient, PageRank score, and average neighbor degree; higher values indicate that more structural information is captured. Neighbor consistency measures the stability of embedding similarity between a node and its immediate neighbors, with higher values reflecting greater robustness to local perturbations.

\begin{table*}[!t]
\renewcommand\arraystretch{1.2}
\caption{Robustness comparison against the \textbf{Nettack} attack at a high perturbation level. The highest results are highlighted with \colorbox[HTML]{B8E7E1}{\textbf{bold}}, while the second-best results are marked with \colorbox[HTML]{D4FAFC}{\underline{underline}} (applied similarly in the following tables).}
\label{tab:high-nettack}
\resizebox{1.0\textwidth}{!}{
\begin{tabular}{lcccccccccccccccccccccccc}
\toprule
\multirow{4}{*}{\textbf{Model}} & \multicolumn{6}{c}{\textbf{Cora}} & \multicolumn{6}{c}{\textbf{Pubmed}} & \multicolumn{6}{c}{\textbf{Ogbn-products}} & \multicolumn{6}{c}{\textbf{Tape-arxiv23}}  \\
\cmidrule(lr){2-7} \cmidrule(lr){8-13} \cmidrule(lr){14-19} \cmidrule(lr){20-25} 
& \multicolumn{2}{c}{GCN} & \multicolumn{2}{c}{GAT} &  \multicolumn{2}{c}{SAGE} & \multicolumn{2}{c}{GCN} & \multicolumn{2}{c}{GAT} &  \multicolumn{2}{c}{SAGE} & \multicolumn{2}{c}{GCN} & \multicolumn{2}{c}{GAT} &  \multicolumn{2}{c}{SAGE} & \multicolumn{2}{c}{GCN} & \multicolumn{2}{c}{GAT} &  \multicolumn{2}{c}{SAGE} \\
\cmidrule(lr){2-3} \cmidrule(lr){4-5} \cmidrule(lr){6-7} \cmidrule(lr){8-9} \cmidrule(lr){10-11} \cmidrule(lr){12-13} \cmidrule(lr){14-15} \cmidrule(lr){16-17} \cmidrule(lr){18-19}\cmidrule(lr){20-21}\cmidrule(lr){22-23}\cmidrule(lr){24-25}
& ACC $\uparrow$ & RDA $\downarrow$  & ACC $\uparrow$ & RDA $\downarrow$  & ACC $\uparrow$ & RDA $\downarrow$  & ACC $\uparrow$ & RDA $\downarrow$ & ACC $\uparrow$ & RDA $\downarrow$ & ACC $\uparrow$ & RDA $\downarrow$ & ACC $\uparrow$ & RDA $\downarrow$ & ACC $\uparrow$ & RDA $\downarrow$ & ACC $\uparrow$ & RDA $\downarrow$ & ACC $\uparrow$ & RDA $\downarrow$ & ACC $\uparrow$ & RDA $\downarrow$ & ACC $\uparrow$ & RDA $\downarrow$ \\
\midrule
Shallow emb. & 58.15 & 31.96 & 54.03 & 33.35 & 65.46 & 20.24 & 70.93 & 20.74 & 79.39 & 11.18 & 81.65 & 6.96 & 75.12 & 11.28 & 74.66 & 11.02 & 68.96 & 13.28 & 56.39 & 17.67 & 51.77 & 19.44 & 61.26 & 12.11 \\
\midrule
TAPE & 65.41 & \colorbox[HTML]{D4FAFC}{\underline{21.30}} & \colorbox[HTML]{B8E7E1}{\textbf{73.51}} & \colorbox[HTML]{B8E7E1}{\textbf{10.43}} & \colorbox[HTML]{D4FAFC}{\underline{73.00}} & \colorbox[HTML]{D4FAFC}{\underline{11.61}} & \colorbox[HTML]{B8E7E1}{\textbf{84.76}} & \colorbox[HTML]{B8E7E1}{\textbf{6.18}} & \colorbox[HTML]{D4FAFC}{\underline{88.52}} & \colorbox[HTML]{D4FAFC}{\underline{3.28}} & 90.02 & 0.95 & \colorbox[HTML]{D4FAFC}{\underline{85.51}} & \colorbox[HTML]{D4FAFC}{\underline{3.42}} & \colorbox[HTML]{D4FAFC}{\underline{86.64}} & \colorbox[HTML]{B8E7E1}{\textbf{1.72}} & 87.07 & \colorbox[HTML]{D4FAFC}{\underline{1.29}} & \colorbox[HTML]{D4FAFC}{\underline{67.30}} & \colorbox[HTML]{D4FAFC}{\underline{12.50}} & \colorbox[HTML]{B8E7E1}{\textbf{64.31}} & \colorbox[HTML]{B8E7E1}{\textbf{15.68}} & \colorbox[HTML]{B8E7E1}{\textbf{77.39}} & 4.07 \\
KEA &\colorbox[HTML]{B8E7E1}{\textbf{67.56}} & \colorbox[HTML]{B8E7E1}{\textbf{20.61}} & \colorbox[HTML]{D4FAFC}{\underline{69.58}} & 18.46 & \colorbox[HTML]{B8E7E1}{\textbf{74.50}} & 13.50 & 80.16 & 11.27 & 79.28 & 11.30 & 88.09 & 2.95 & 83.36 & 5.50 & 83.91 & 4.25 & 86.84 & 2.61 & 62.95 & 14.53 & 53.26 & 25.55 & 71.84 & 6.62 \\

\midrule
LLaMA &60.55 & 30.22 & 65.74 & 23.09 & 67.29 & 17.01 & 79.61 & 12.09 & 80.15 & 8.90 & 88.20 & 1.67 & 80.10 & 9.19 & 80.91 & 4.55 & 84.14 & 5.12 & 62.18 & 18.26 & 58.33 & 21.10 & 70.95 & 9.49 \\

TE3L & 61.05 & 31.89 & 52.11 & 36.96 & 70.14 & 19.17 & 78.97 & 12.28 & 82.08 & 8.93 & 88.41 & 3.06 & 82.65 & 6.84 & 83.82 & 5.23 & \colorbox[HTML]{B8E7E1}{\textbf{88.11}} & 2.17 & 63.27 & 14.98 & 56.68 & 26.02 & 73.31 & 3.35 \\
Linq &  \colorbox[HTML]{D4FAFC}{\underline{65.41}} & 25.51 & 63.71 & 23.45 & 72.46 & 14.67 & 79.19 & 12.56 & 83.37 & 6.38 & 88.31 & 2.83 & 83.80 & 4.69 & 82.32 & 5.43 & \colorbox[HTML]{D4FAFC}{\underline{88.02}} & 1.85 & 65.76 & 14.10 & 56.42 & 21.10 & 75.02 & 4.63 \\

\midrule
SimTeg & 62.26 & 28.28 & 63.37 & 24.47 & 71.71 & 16.10 & 79.51 & 12.82 & 81.66 & 8.74 & 87.87 & 2.97 & 81.66 & 7.12 & 82.85 & 4.02 & 86.07 & 3.57 & 63.91 & 13.91 & 57.91 & 22.57 & 74.14 & 2.77 \\

E5-Large & 62.86 & 27.99 & 67.29 & \colorbox[HTML]{D4FAFC}{\underline{18.04}} & 71.18 & \colorbox[HTML]{B8E7E1}{\textbf{11.02}} & \colorbox[HTML]{D4FAFC}{\underline{83.25}} & 8.61 & \colorbox[HTML]{B8E7E1}{\textbf{91.20}} & \colorbox[HTML]{B8E7E1}{\textbf{0.71}} & \colorbox[HTML]{D4FAFC}{\underline{92.06}} & \colorbox[HTML]{D4FAFC}{\underline{0.35}} & \colorbox[HTML]{B8E7E1}{\textbf{85.84}} & \colorbox[HTML]{B8E7E1}{\textbf{3.31}} & \colorbox[HTML]{B8E7E1}{\textbf{87.03}} & \colorbox[HTML]{D4FAFC}{\underline{1.81}} & 87.64 & \colorbox[HTML]{B8E7E1}{\textbf{1.23}} & \colorbox[HTML]{B8E7E1}{\textbf{67.49}} & 13.22 & \colorbox[HTML]{D4FAFC}{\underline{62.28}} & \colorbox[HTML]{D4FAFC}{\underline{17.92}} & \colorbox[HTML]{D4FAFC}{\underline{76.75}} & \colorbox[HTML]{B8E7E1}{\textbf{0.98}} \\

ModernBert & 59.20 & 25.51 & 60.00 & 23.75 & 61.83 & 16.20 & 82.51 & \colorbox[HTML]{D4FAFC}{\underline{8.56}} & 84.22 & 4.02 & \colorbox[HTML]{B8E7E1}{\textbf{92.49}} & \colorbox[HTML]{B8E7E1}{\textbf{-0.35}} & 83.47 & 5.37 & 80.80 & 3.83 & 86.84 & 2.04 & 65.75 & \colorbox[HTML]{B8E7E1}{\textbf{9.94}} & 51.09 & 31.43 & 75.32 & \colorbox[HTML]{D4FAFC}{\underline{2.55}} \\

\bottomrule
\end{tabular}
}
\end{table*}

\section{Assessment and Analysis}
In this section, we assess the robustness of LLM-enhanced GNNs against poisoning attacks. We first introduce the evaluation setup and then answer the following research questions: 
\begin{itemize}
    \item \textbf{RQ1:} How robust are LLM-enhanced GNNs against structural poisoning attacks, including targeted and non-targeted attacks?
    \item \textbf{RQ2:} How robust are LLM-enhanced GNNs against textual poisoning attacks at the levels of character, word, and sentence? 
\end{itemize}

To gain deep insights, we further provide a fine-grained analysis of structural and textual attack results.

\subsection{Evaluation Setup} \label{evaluation_setup}
\textbf{Baseline.}
To examine whether LLM/LM-augmented node embeddings enhance GNN robustness, we adopt shallow embeddings as a baseline. These embeddings are derived from traditional methods such as BoW~\cite{harris1954distributional}, TF-IDF~\cite{salton1988term}, and Word2Vec~\cite{rong2014word2vec} (collectively referred to as \textbf{Shallow emb.}). The specific shallow embedding method used for each dataset is listed in Table~\ref{tab:datasets}.

\textbf{Implementation Details.}
All models were implemented in Python using PyTorch on a machine equipped with an RTX 4070 Ti Super. GNN models were built with PyTorch Geometric (PyG), while LLM/LM models were accessed through official APIs or HuggingFace Transformers~\cite{wolf2020transformers}. Structural attacks were implemented via the DeepRobust~\cite{li2020deeprobust} toolbox, and textual attacks with OpenAttack~\cite{zeng2020openattack}. Following~\cite{jin2020graph}, we randomly selected 10\% of nodes for training, 10\% for validation, and the remaining 80\% for testing in each dataset. To ensure strong performance in node classification tasks, we conducted a grid search to determine suitable GNN hyperparameters. Specifically, all GNNs are configured with two layers, a hidden dimension of 256, dropout of 0.5, and a learning rate of 0.001. For GAT, we configured the first layer with eight attention heads and the second layer with a single head. Each experiment was repeated five times with different random seeds, and we report the average results. The details of each attack setting are provided in the corresponding sections.

\begin{figure*}[!t]
  \centering
  \subfloat[ACC under the targeted attacks at a low perturbation level.]{\includegraphics[width=\columnwidth]{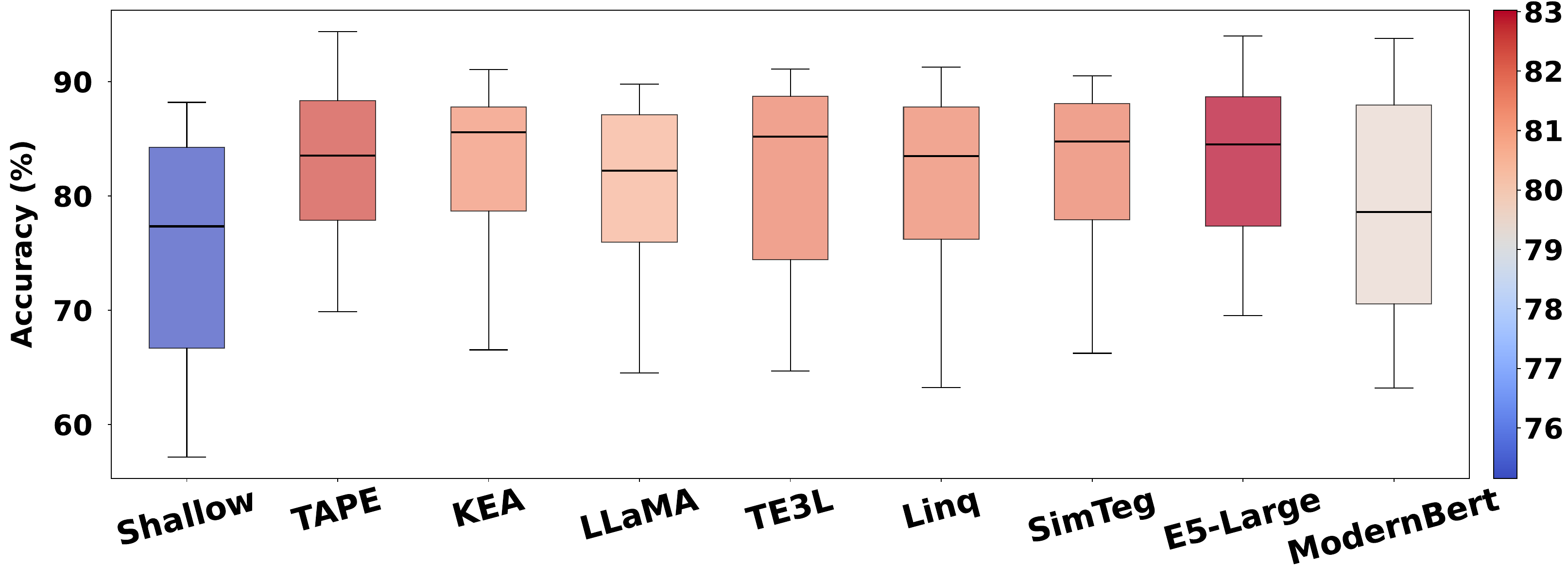}
  \label{fig:taraccboxl}}
  \hspace{5pt}
  \subfloat[ACC under the targeted attacks at a high perturbation level.]{\includegraphics[width=\columnwidth]{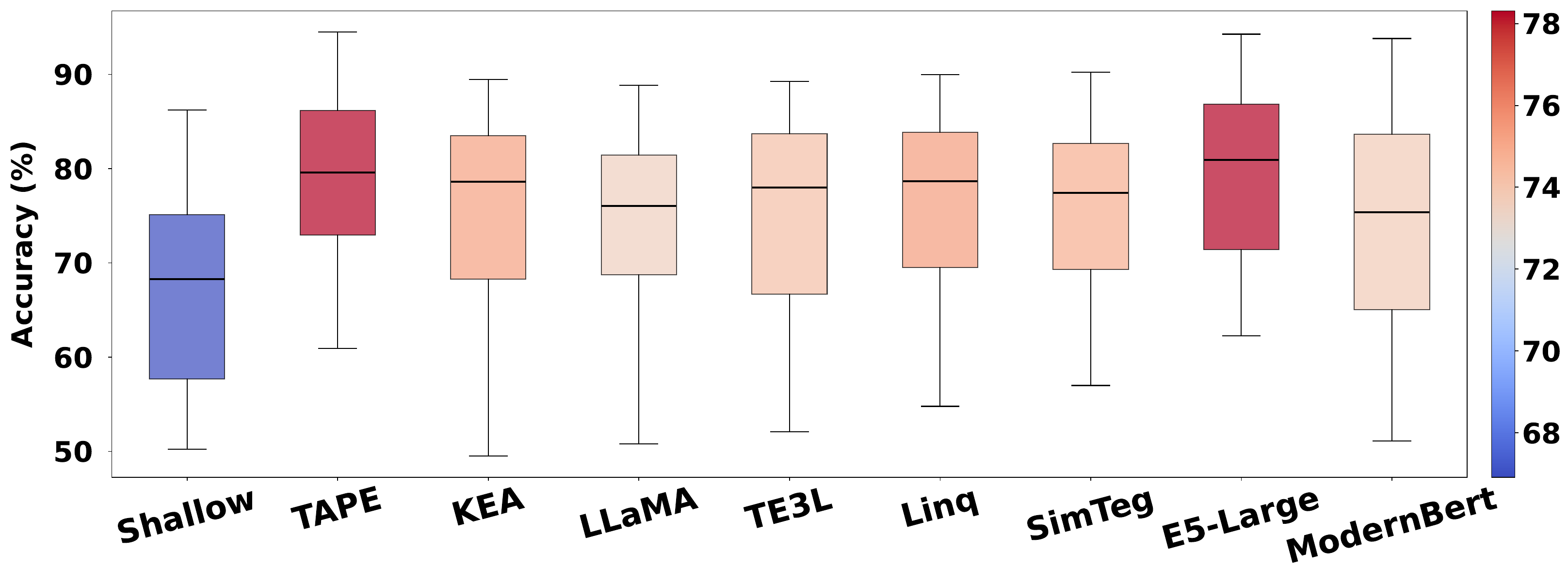}
  \label{fig:taraccboxh}}
  \vspace{5pt}
  \subfloat[RDA under the targeted attacks at a low perturbation level.]{\includegraphics[width=\columnwidth]{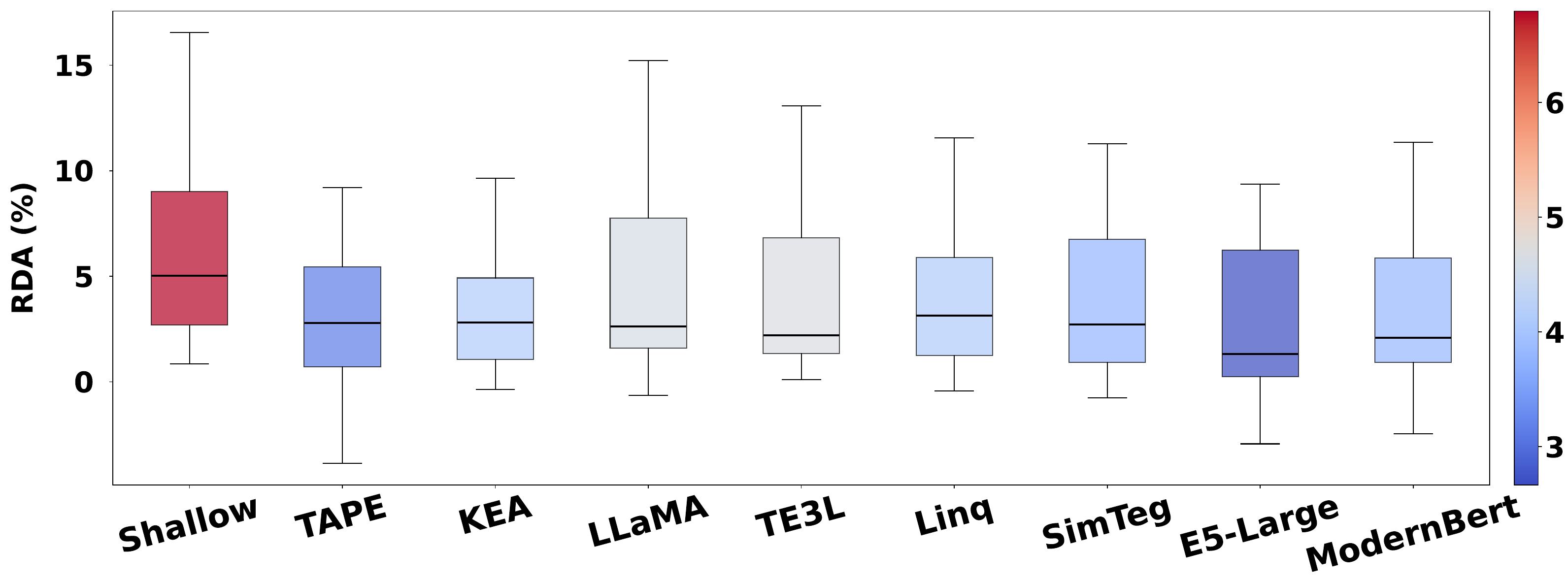}
  \label{fig:tarrdaboxl}}
  \hspace{5pt}
  \subfloat[RDA under the targeted attacks at a high perturbation level.]{\includegraphics[width=\columnwidth]{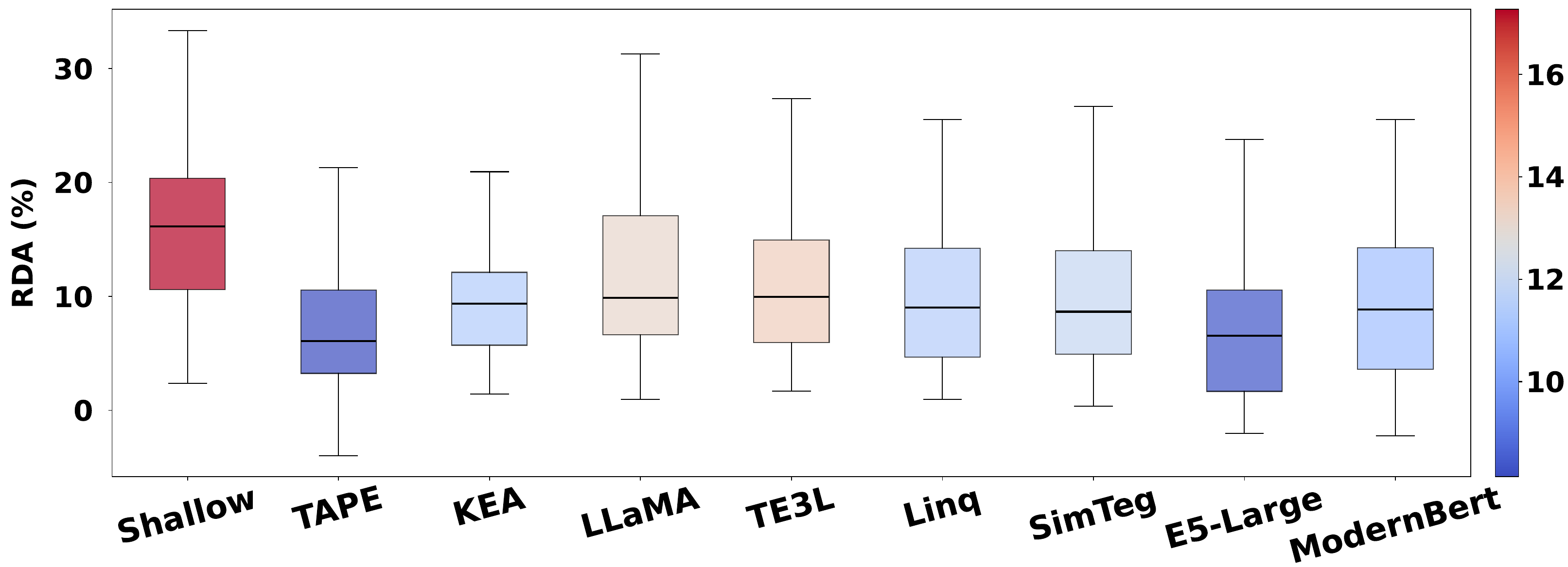}
  \label{fig:tarrdaboxh}}
  \caption{ACC and RDA under the targeted structural attacks with different perturbation levels. In each box plot, the color denotes the average score and the central line indicates the median.}
  \label{fig:boxplots_targeted}
\end{figure*}

\subsection{Results on Targeted Structural Attacks} \label{evaluation_targeted}
\textbf{Attack Settings.} We employed Nettack~\cite{zugner2018adversarial}, SGA~\cite{li2021adversarial}, and NAG-R~\cite{zhao2023black} to perform targeted structural poisoning attacks on four datasets: Cora, Pubmed, Ogbn-products, and Tape-arxiv23. Following~\cite{zhu2019robust}, we varied the number of perturbations per targeted node between 1 and 5, representing low and high perturbation levels. Target nodes were selected from the test set with degrees greater than 10. For Cora, all target nodes were used due to its small size. For the larger datasets, we reduced attack runtime by sampling 10\% of target nodes for Pubmed and Ogbn-products, whereas 50\% for Tape-arxiv23 due to the limited number of high-degree nodes. The perturbed graph structures combined with clean textual attributes were then used to train various LLM-enhanced GNNs. Finally, we evaluated the robustness of each model on the target nodes in terms of ACC and RDA.

\textbf{Accuracy on Clean Graphs.}
As shown in Table~\ref{tab:targeted-clean} in Appendix~\ref{appendix_clean}, LLM-enhanced GNN models outperform the baseline in almost all cases, even on the newly released Tape-arxiv23. For instance, with the GCN backbone, the maximum accuracy improvements over the baseline are 4.89\%, 1.91\%, 4.85\%, and 13.55\% on Cora, Pubmed, Ogbn-products, and Tape-arxiv23, respectively. We observe that LLMs achieve state-of-the-art performance on small-scale datasets by leveraging extensive pretrained knowledge. However, as the volume of training data grows, fine-tuned LMs exhibit better performance. Among them, E5-Large is the most effective model, achieving the best or second-best accuracy across nearly all GNN backbones on Pubmed and Ogbn-products. Overall, these results highlight the effectiveness of incorporating LLMs/LMs into graph learning and confirm that high-quality node embeddings are critical for improving GNN accuracy.

\textbf{Robustness against Targeted Structural Attacks.}
Due to space constraints, we only report results for Nettack at a high perturbation level and provide the complete results in Appendix~\ref{appendix_structural}. As shown in Tables~\ref{tab:high-nettack},~\ref{tab:high-sga}, and~\ref{tab:high-nag}, LLM-enhanced GNN models consistently exhibit superior robustness against targeted structural attacks compared to the baseline. For instance, under the Nettack attack with high perturbation, GCN with shallow embeddings achieves only 58.15\%, 70.93\%, 75.12\%, and 56.39\% accuracy on Cora, Pubmed, Ogbn-products, and Tape-arxiv23, respectively. However, the best LLM-enhanced models significantly outperform it, reaching 67.56\%, 84.76\%, 85.84\%, and 67.49\% on the same datasets. Similar improvements are observed for the RDA metric, highlighting the inherent vulnerability of shallow embeddings to structural perturbations and the relative robustness provided by the semantically rich embeddings derived from LLMs/LMs. 

Fig.~\ref{fig:boxplots_targeted} summarizes the ACC and RDA results under the three targeted structural attacks across different GNN backbones and datasets. Overall, TAPE and E5-Large exhibit the strongest robustness. Among the three LLM-GNN integration approaches, LLM-Explanation methods achieve the best, indicating that LLMs can extract robust features by summarizing textual attributes. In contrast, LLM-Embedding methods show dataset sensitivity. For example, while Linq sometimes approaches top performance, its effectiveness varies across datasets. Furthermore, Fig.~\ref{fig:taraccboxl} and~\ref{fig:taraccboxh}, along with Fig.~\ref{fig:tarrdaboxl} and~\ref{fig:tarrdaboxh} show that all models suffer noticeable performance degradation as perturbation levels increase. This is expected since higher perturbation levels introduce more structural distortions, increasing the difficulty of accurate prediction. Notably, the performance gap between LLM-enhanced models and the baseline widens at higher perturbation levels, indicating that the robustness advantage of LLM-enhanced models is more pronounced under severe attacks. A possible explanation is that heavily distorted graph structures impair the message-passing mechanism of GNNs, while embeddings enhanced by LLMs or LMs retain discriminative semantic features through the encoding of node textual attributes.

\textbf{\underline{Key Takeaway 1:}} LLM-enhanced GNNs demonstrate stronger robustness against targeted structural attacks compared to the shallow embedding-based baseline.

\textbf{\underline{Key Takeaway 2:}} The robustness advantage of LLM-enhanced models over the baseline becomes more pronounced as perturbations increase.

\begin{table*}[!tbp]
\renewcommand\arraystretch{1.2}
\caption{Robustness comparison against the \textbf{Mettack} attack at a high perturbation level.}
\label{tab:high-mettack}
\resizebox{1.0\textwidth}{!}{
\begin{tabular}{lcccccccccccccccccccccccc}
\toprule
\multirow{4}{*}{\textbf{Model}} & \multicolumn{6}{c}{\textbf{Cora}} & \multicolumn{6}{c}{\textbf{Pubmed}} & \multicolumn{6}{c}{\textbf{Ogbn-products}} & \multicolumn{6}{c}{\textbf{Tape-arxiv23}}  \\
\cmidrule(lr){2-7} \cmidrule(lr){8-13} \cmidrule(lr){14-19} \cmidrule(lr){20-25} 
& \multicolumn{2}{c}{GCN} & \multicolumn{2}{c}{GAT} &  \multicolumn{2}{c}{SAGE} & \multicolumn{2}{c}{GCN} & \multicolumn{2}{c}{GAT} & \multicolumn{2}{c}{SAGE} & \multicolumn{2}{c}{GCN} & \multicolumn{2}{c}{GAT} & \multicolumn{2}{c}{SAGE} & \multicolumn{2}{c}{GCN} & \multicolumn{2}{c}{GAT} & \multicolumn{2}{c}{SAGE} \\
\cmidrule(lr){2-3} \cmidrule(lr){4-5} \cmidrule(lr){6-7} \cmidrule(lr){8-9} \cmidrule(lr){10-11} \cmidrule(lr){12-13} \cmidrule(lr){14-15} \cmidrule(lr){16-17} \cmidrule(lr){18-19}\cmidrule(lr){20-21}\cmidrule(lr){22-23}\cmidrule(lr){24-25}
& ACC $\uparrow$ & RDA $\downarrow$  & ACC $\uparrow$ & RDA $\downarrow$  & ACC $\uparrow$ & RDA $\downarrow$  & ACC $\uparrow$ & RDA $\downarrow$ & ACC $\uparrow$ & RDA $\downarrow$ & ACC $\uparrow$ & RDA $\downarrow$ & ACC $\uparrow$ & RDA $\downarrow$ & ACC $\uparrow$ & RDA $\downarrow$ & ACC $\uparrow$ & RDA $\downarrow$ & ACC $\uparrow$ & RDA $\downarrow$ & ACC $\uparrow$ & RDA $\downarrow$ & ACC $\uparrow$ & RDA $\downarrow$ \\
\midrule
Shallow emb. &74.60 & 7.92 & 73.57 & 8.03 & 75.77 & 5.32 & 74.12 & 14.02 & 72.25 & 15.49 & 80.13 & 6.14 & 63.78 & 3.29 & 66.39 & 4.32 & 59.47 & 4.53 & 51.03 & 11.59 & 50.43 & 9.09 & 51.60 & 8.38 \\
\midrule
TAPE &78.02 & \colorbox[HTML]{D4FAFC}{\underline{4.55}} & 76.16 & \colorbox[HTML]{D4FAFC}{\underline{5.37}} & 78.17 & 4.46 & \colorbox[HTML]{B8E7E1}{\textbf{84.91}} & \colorbox[HTML]{D4FAFC}{\underline{8.82}} & \colorbox[HTML]{B8E7E1}{\textbf{83.63}} & \colorbox[HTML]{B8E7E1}{\textbf{6.84}} & \colorbox[HTML]{B8E7E1}{\textbf{94.13}} & \colorbox[HTML]{D4FAFC}{\underline{0.43}} & \colorbox[HTML]{B8E7E1}{\textbf{84.26}} & \colorbox[HTML]{B8E7E1}{\textbf{1.84}} & \colorbox[HTML]{B8E7E1}{\textbf{84.20}} & \colorbox[HTML]{B8E7E1}{\textbf{1.36}} & \colorbox[HTML]{B8E7E1}{\textbf{85.85}} & \colorbox[HTML]{B8E7E1}{\textbf{0.23}} & \colorbox[HTML]{B8E7E1}{\textbf{71.33}} & \colorbox[HTML]{D4FAFC}{\underline{4.22}} & \colorbox[HTML]{B8E7E1}{\textbf{69.29}} & \colorbox[HTML]{B8E7E1}{\textbf{6.92}} & \colorbox[HTML]{B8E7E1}{\textbf{73.69}} & \colorbox[HTML]{B8E7E1}{\textbf{0.87}} \\

KEA &77.02 & 8.53 & 76.13 & 9.87 & \colorbox[HTML]{B8E7E1}{\textbf{82.24}} & \colorbox[HTML]{D4FAFC}{\underline{3.08}} & 77.72 & 12.64 & 75.33 & 12.96 & 84.97 & 4.86 & 81.98 & 2.28 & 80.98 & 2.62 & 83.82 & 0.53 & 62.08 & 9.83 & 60.82 & 12.95 & 64.41 & 7.03 \\

\midrule
LLaMA &77.36 & 5.24 & 76.64 & 6.08 & 77.65 & 4.41 & 73.05 & 16.87 & 73.94 & 12.98 & 82.84 & 5.84 & 81.80 & 2.27 & 74.81 & 6.98 & 82.85 & 0.70 & 64.79 & 9.40 & 59.03 & 16.20 & 65.33 & 8.12 \\

TE3L &78.85 & 5.73 & 76.96 & 7.28 & 80.43 & 4.12 & 76.54 & 13.77 & 75.94 & 12.80 & 85.11 & 4.41 & 82.16 & 2.28 & 81.54 & \colorbox[HTML]{D4FAFC}{\underline{1.56}} & 84.24 & 0.37 & 63.48 & 9.53 & 62.05 & 9.81 & 65.55 & 6.17 \\

Linq &\colorbox[HTML]{D4FAFC}{\underline{79.55}} & 5.62 & 77.54 & 6.09 & 80.89 & 4.18 & 76.74 & 14.44 & 77.29 & 11.31 & 88.48 & 3.09 & 82.57 & 1.99 & 79.42 & 3.42 & 84.52 & \colorbox[HTML]{D4FAFC}{\underline{0.26}} & 67.19 & 7.09 & 64.19 & 10.25 & 67.80 & 6.52 \\

\midrule
 SimTeg &78.50 & 5.75 & \colorbox[HTML]{D4FAFC}{\underline{78.16}} & 5.71 & 80.21 & 4.25 & 77.79 & 12.87 & 74.85 & 19.57 & 86.60 & 6.46 & 81.47 & 2.21 & 80.66 & 2.69 & 83.21 & 0.44 & 66.02 & 8.32 & 64.82 & 8.67 & 67.48 & 5.83 \\

E5-Large &\colorbox[HTML]{B8E7E1}{\textbf{80.60}} & \colorbox[HTML]{B8E7E1}{\textbf{4.51}} & \colorbox[HTML]{B8E7E1}{\textbf{78.52}} & \colorbox[HTML]{B8E7E1}{\textbf{4.24}} & \colorbox[HTML]{D4FAFC}{\underline{81.62}} & \colorbox[HTML]{B8E7E1}{\textbf{2.59}} & \colorbox[HTML]{D4FAFC}{\underline{84.71}} & 8.97 & \colorbox[HTML]{D4FAFC}{\underline{82.60}} & \colorbox[HTML]{D4FAFC}{\underline{8.07}} & \colorbox[HTML]{D4FAFC}{\underline{93.94}} & \colorbox[HTML]{B8E7E1}{\textbf{0.21}} & \colorbox[HTML]{D4FAFC}{\underline{83.23}} & 2.04 & \colorbox[HTML]{D4FAFC}{\underline{82.48}} & 1.86 & \colorbox[HTML]{D4FAFC}{\underline{84.70}} & 0.34 & \colorbox[HTML]{D4FAFC}{\underline{70.04}} & \colorbox[HTML]{B8E7E1}{\textbf{4.16}} & \colorbox[HTML]{D4FAFC}{\underline{66.13}} & \colorbox[HTML]{D4FAFC}{\underline{7.73}} & \colorbox[HTML]{D4FAFC}{\underline{71.91}} & \colorbox[HTML]{D4FAFC}{\underline{0.88}} \\

ModernBert &72.39 & 6.85 & 69.07 & 8.38 & 70.22 & 6.10 & 83.56 & \colorbox[HTML]{B8E7E1}{\textbf{7.53}} & 79.20 & 16.03 & 92.79 & 1.14 & 82.05 & \colorbox[HTML]{D4FAFC}{\underline{1.91}} & 78.78 & 2.56 & 83.63 & 0.32 & 66.32 & 5.70 & 60.51 & 11.65 & 68.14 & 3.31 \\

\bottomrule
\end{tabular}}
\end{table*}

\begin{figure*}[!t]
  \centering
  \subfloat[ACC under the non-targeted attacks at a low perturbation level.]{\includegraphics[width=\columnwidth]{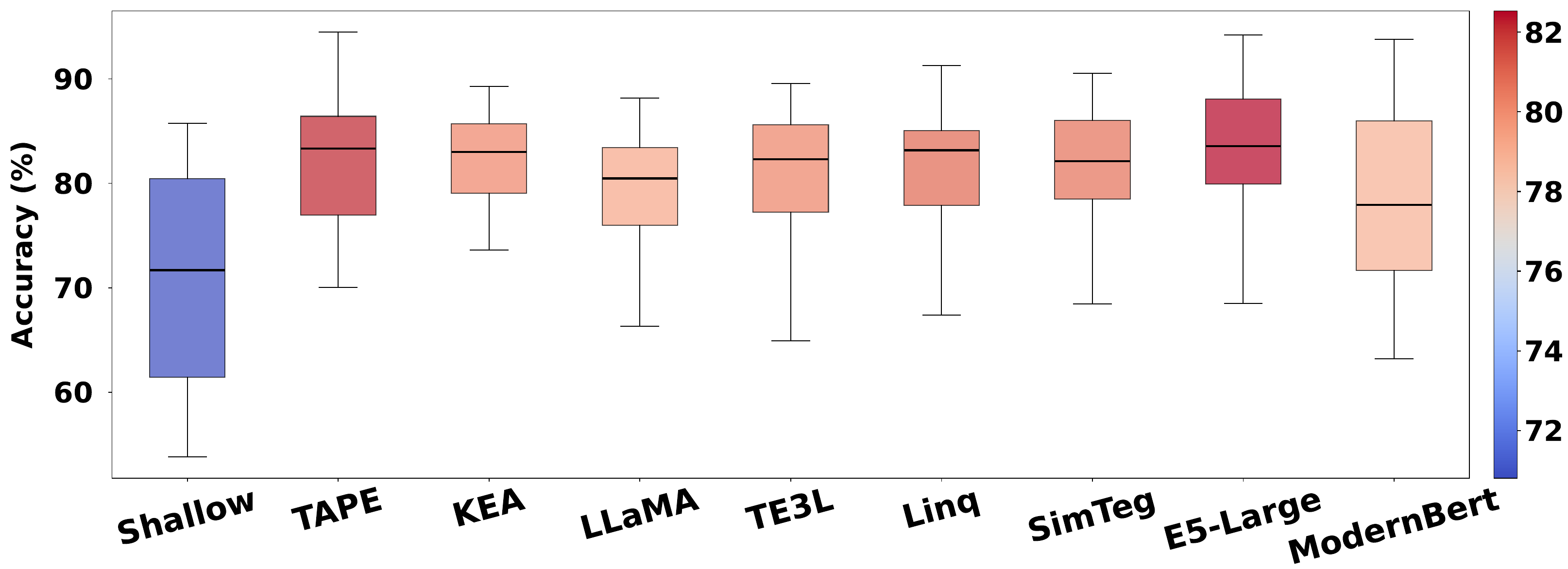}
  \label{fig:globaccboxl}}
  \hspace{5pt}
  \subfloat[ACC under the non-targeted attacks at a high perturbation level.]{\includegraphics[width=\columnwidth]{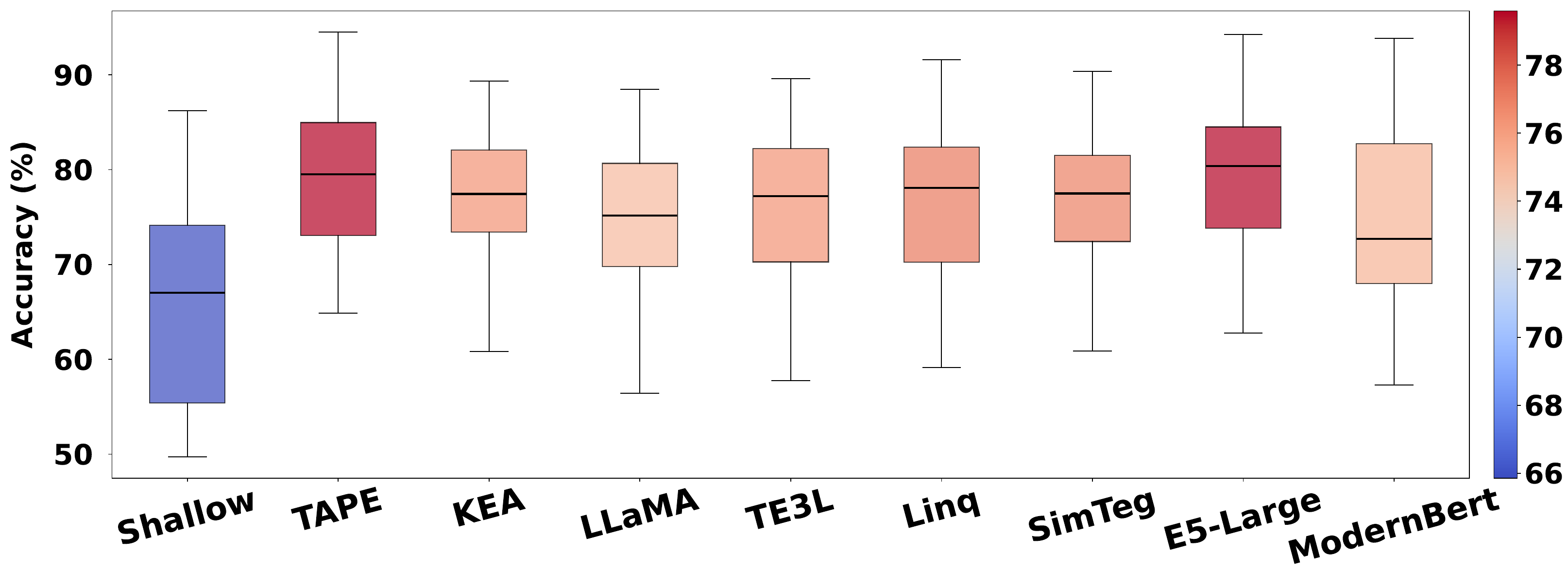}
  \label{fig:globaccboxh}}
  \vspace{5pt}
  \subfloat[RDA under the non-targeted attacks at a low perturbation level.]{\includegraphics[width=\columnwidth]{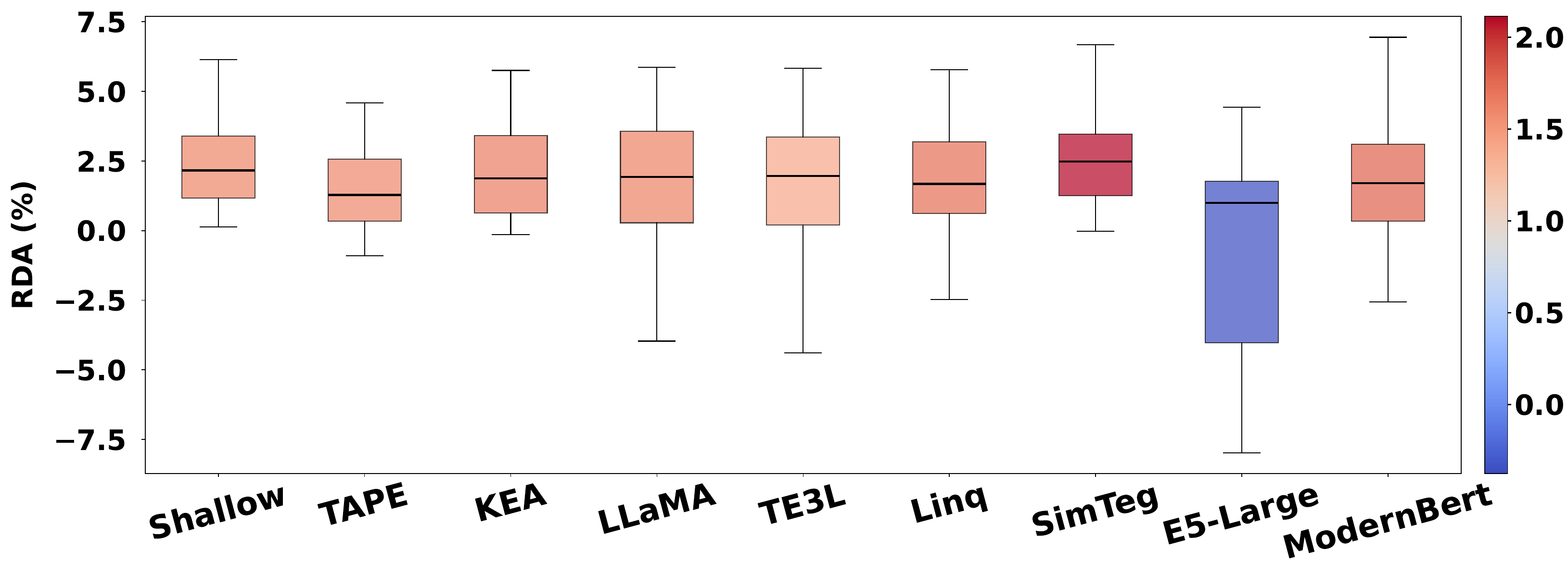}
  \label{fig:globrdaboxl}}
  \hspace{5pt}
  \subfloat[RDA under the non-targeted attacks at a high perturbation level.]{\includegraphics[width=\columnwidth]{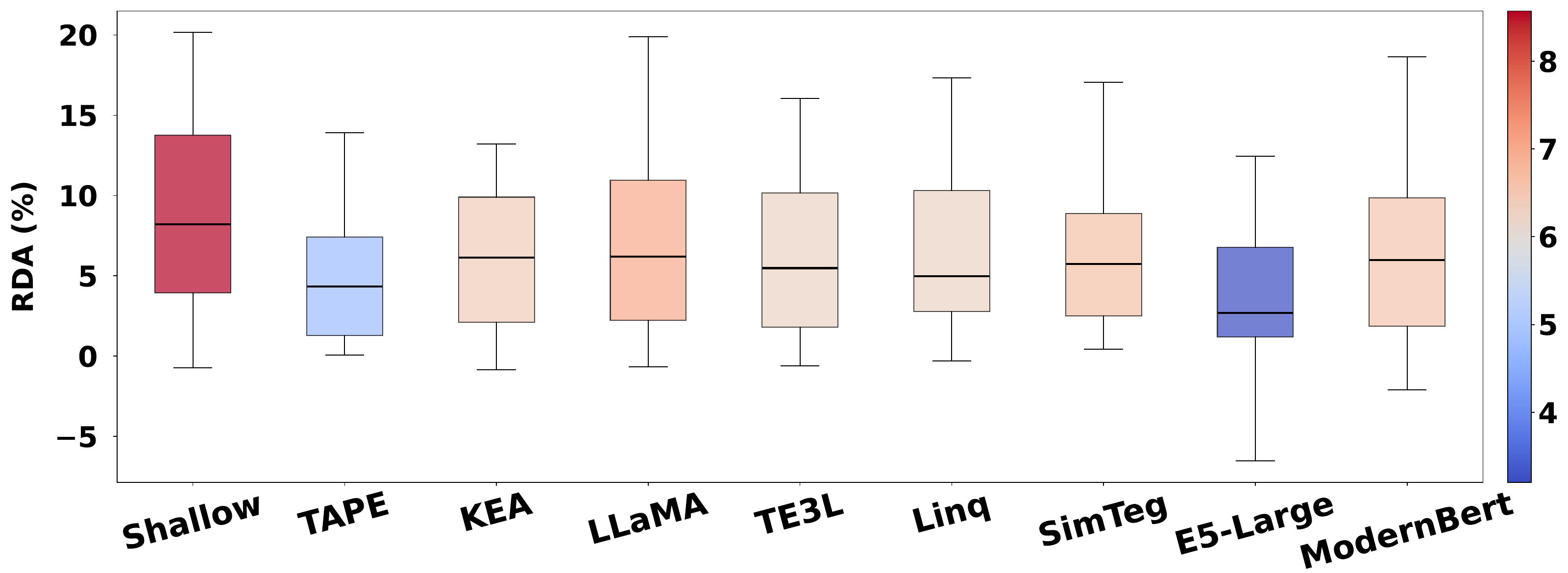}
  \label{fig:globrdaboxh}}
  \caption{ACC and RDA under the non-targeted structural attacks with different perturbation levels. In each box plot, the color denotes the average score and the central line indicates the median.}
  \label{fig:boxplots_nontargeted}
\end{figure*}

\subsection{Results on Non-Targeted Structural Attacks} \label{evaluation_nontargeted}

\textbf{Attack Settings.} We employed Mettack~\cite{zugner_adversarial_2019}, DICE~\cite{waniek2018hiding}, and PGA~\cite{zhu2024simple} to perform non-targeted structural poisoning attacks on the four datasets. Following~\cite{zhang2024can}, Mettack and PGA were applied with global perturbation rates (i.e., the proportion of modified edges relative to the total number of edges in the original graph) of 5\% and 20\%, corresponding to low and high perturbation levels. Since DICE generally causes minor performance degradation, we deployed it under more aggressive budgets of 10\% and 40\%. The perturbed graph structures combined with clean textual attributes were then used to train LLM-enhanced GNNs. Finally, we evaluated the robustness of each model on the whole test set in terms of ACC and RDA.

\textbf{Accuracy on Clean Graphs.} 
As shown in Table~\ref{tab:global-clean} in Appendix~\ref{appendix_clean}, LLM-enhanced GNN models exhibit significant advantages over the baseline, indicating that their enriched node embeddings contribute a lot to performance gains. For instance, with the GCN backbone, the maximum accuracy improvements over the baseline are 4.18\%, 8.02\%, 30.17\%, and 29.02\% on Cora, Pubmed, Ogbn-products, and Tape-arxiv23, respectively. We observe that LLM-Explanation and LM-Embedding approaches often achieve the best accuracy, likely because their LMs are fine-tuned on specific datasets, making them compatible to perform node classification on those datasets. Additionally, the baseline performs poorly on the newly published dataset, i.e., Tape-arxiv23, whereas LLM-enhanced GNNs maintain strong performance, indicating that pretrained knowledge in LLMs/LMs provides a certain degree of generality. Overall, these findings further highlight the great potential of integrating LLMs/LMs into GNNs.

\textbf{Robustness against Non-Targeted Structural Attacks.}
Due to space constraints, we only report results for Mettack at a high perturbation level and provide the complete results in Appendix~\ref{appendix_structural}. As shown in Tables~\ref{tab:high-mettack},~\ref{tab:high-dice}, and~\ref{tab:high-pga}, LLM-enhanced GNN models demonstrate strong robustness against non-targeted structural attacks. For instance, under the Mettack attack with high perturbation, the best-performing LLM-enhanced GCN models achieve accuracy improvements of 8.04\%, 14.56\%, 32.11\%, and 39.78\% over the baseline on Cora, Pubmed, Ogbn-products, and Tape-arxiv23, respectively. Interestingly, in some cases, LLM-enhanced models yield negative RDA values, meaning that they achieve higher accuracy under attacks than on clean graphs. This phenomenon can be explained by the fact that LLM/LM-enhanced models capture rich semantic information beyond the graph structure. When structural perturbations are introduced, they may serve as a form of regularization that reduces overfitting and improves generalization. In contrast, the baseline, which relies primarily on graph topology, cannot benefit from such effects. We also notice that ModernBert-enhanced GNNs exhibit lower robustness than the baseline on Cora but outperform it on the other datasets, likely due to the difficulty of fine-tuning large LMs on small-scale data.

Fig.~\ref{fig:boxplots_nontargeted} summarizes the ACC and RDA results under the three non-targeted attacks across different GNN backbones and datasets. Overall, all LLM-enhanced GNN models demonstrate a clear advantage over the baseline, with TAPE and E5-Large exhibiting the strongest robustness, whereas ModernBert shows inconsistent robustness across different settings. These results confirm that LLM/LM-enhanced embeddings, particularly those produced by well-fine-tuned LMs, provide substantial robustness against non-targeted structural perturbations.

\textbf{\underline{Key Takeaway 3:}} LLM-enhanced GNNs show significant robustness improvements over the baseline under non-targeted structural attacks in nearly all settings.

\begin{figure}[!t]
  \centering
  \captionsetup[sub]{aboveskip=0pt,belowskip=0pt,skip=4pt}
  \begin{subfigure}{0.48\columnwidth}
    \includegraphics[width=\linewidth]{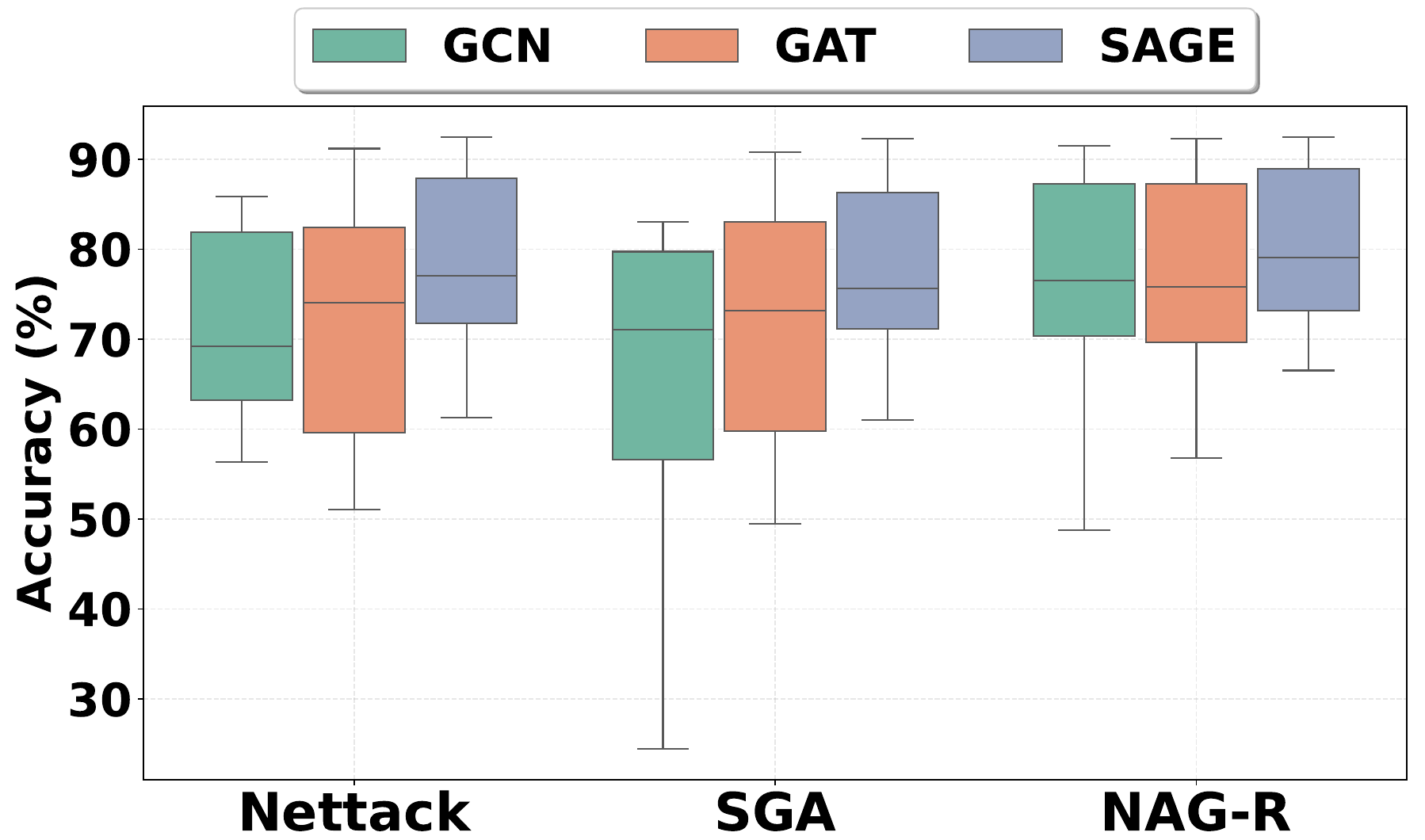}
    \caption{ACC under the targeted structural attacks.}
    \label{fig:targetgnnacc}
  \end{subfigure}\hfill
  \begin{subfigure}{0.48\columnwidth}
    \includegraphics[width=\linewidth]{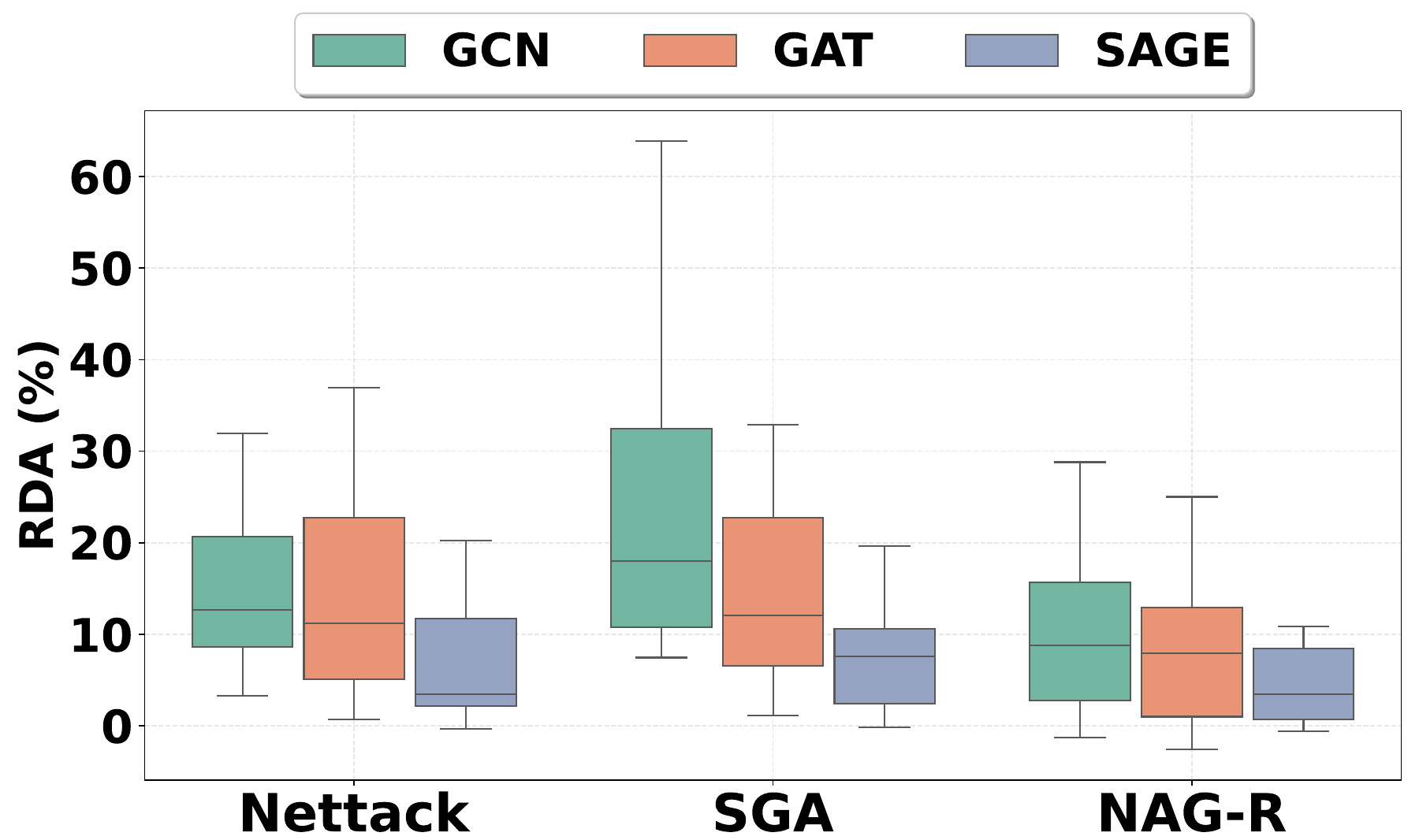}
    \caption{RDA under the targeted structural attacks.}
    \label{fig:targetgnnrda}
  \end{subfigure}
  \begin{subfigure}{0.48\columnwidth}
    \includegraphics[width=\linewidth]{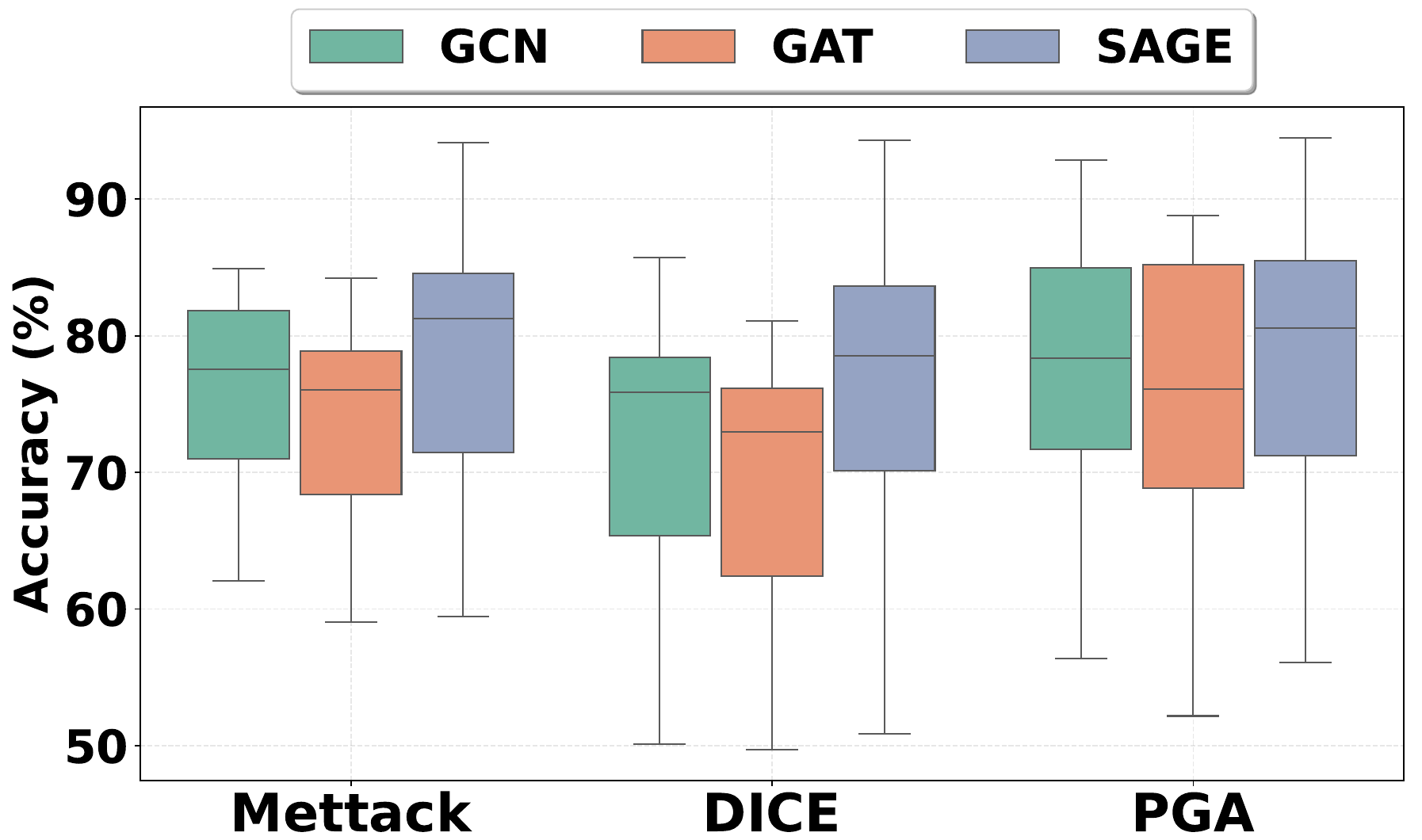}
    \caption{ACC under the non-targeted structural attacks.}
    \label{fig:globalgnnacc}
  \end{subfigure}\hfill
  \begin{subfigure}{0.48\columnwidth}
    \includegraphics[width=\linewidth]{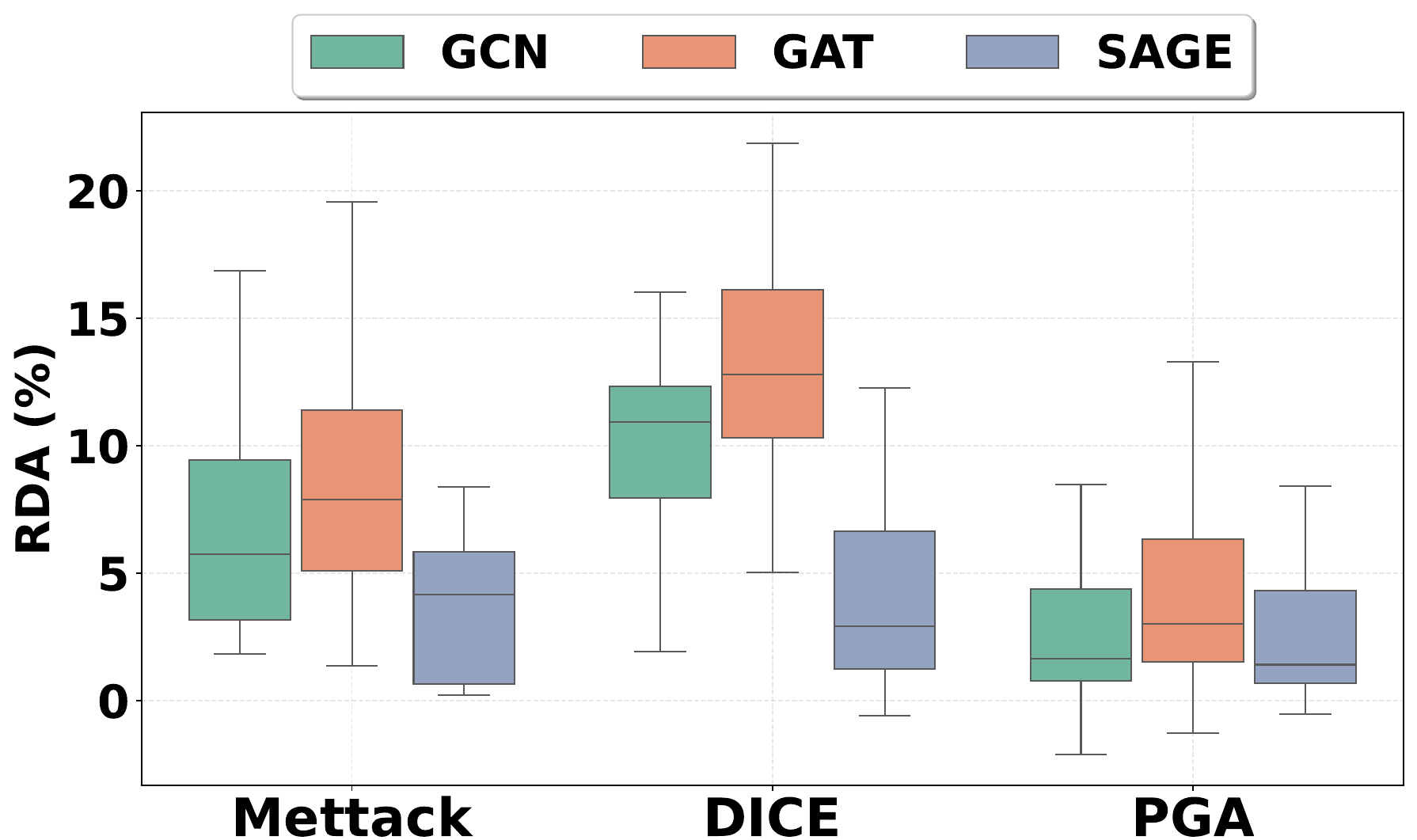}
    \caption{RDA under the non-targeted structural attacks.}
    \label{fig:globalgnnrda}
  \end{subfigure}
  \caption{ACC and RDA under the structural attacks with different GNN models.}
  \label{fig:boxplots_gnn}
\end{figure}

\begin{figure}[!t]
  \centering
    \includegraphics[width=\linewidth]{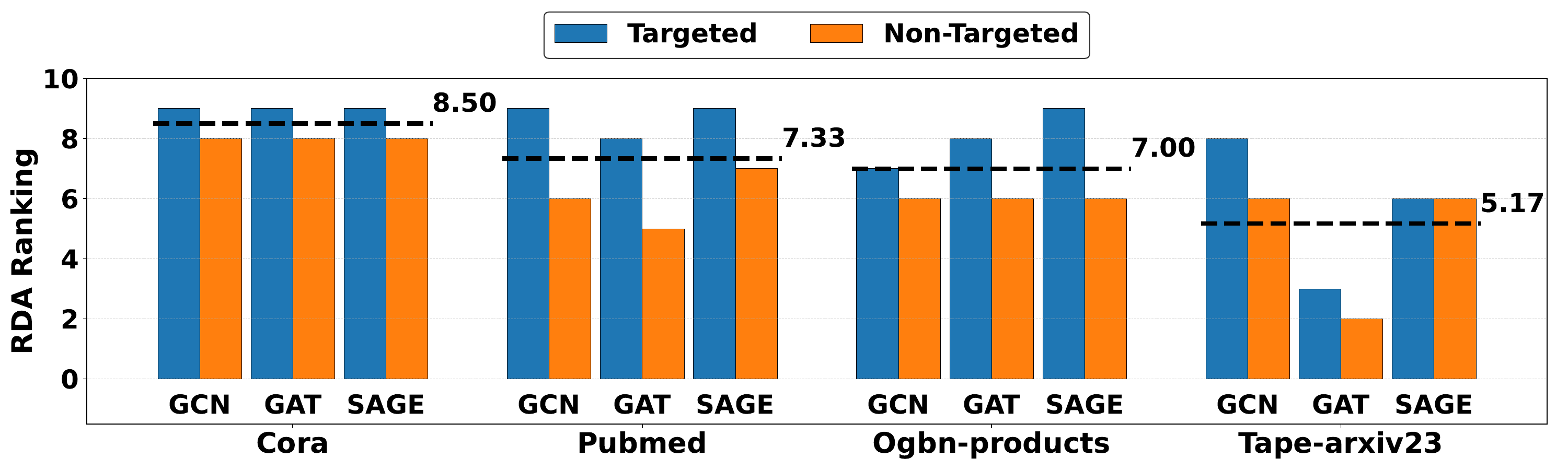}
\caption{RDA ranking of the shallow embedding-based baseline. A lower ranking indicates that the baseline outperforms some LLM-enhanced models. The black dashed line represents the average ranking.}
\label{fig:rdarank}
\vspace{-1mm}
\end{figure}

Regarding the robustness of different GNN backbones, Fig.~\ref{fig:boxplots_gnn} shows that the GraphSAGE backbone consistently exhibits stronger robustness than GCN and GAT. One possible explanation is that GraphSAGE preserves strong self-representations through feature concatenation and employs mean aggregation, which reduces the influence of adversarially perturbed neighbors.

\begin{figure}[!t]
  \centering
\includegraphics[width=\linewidth]{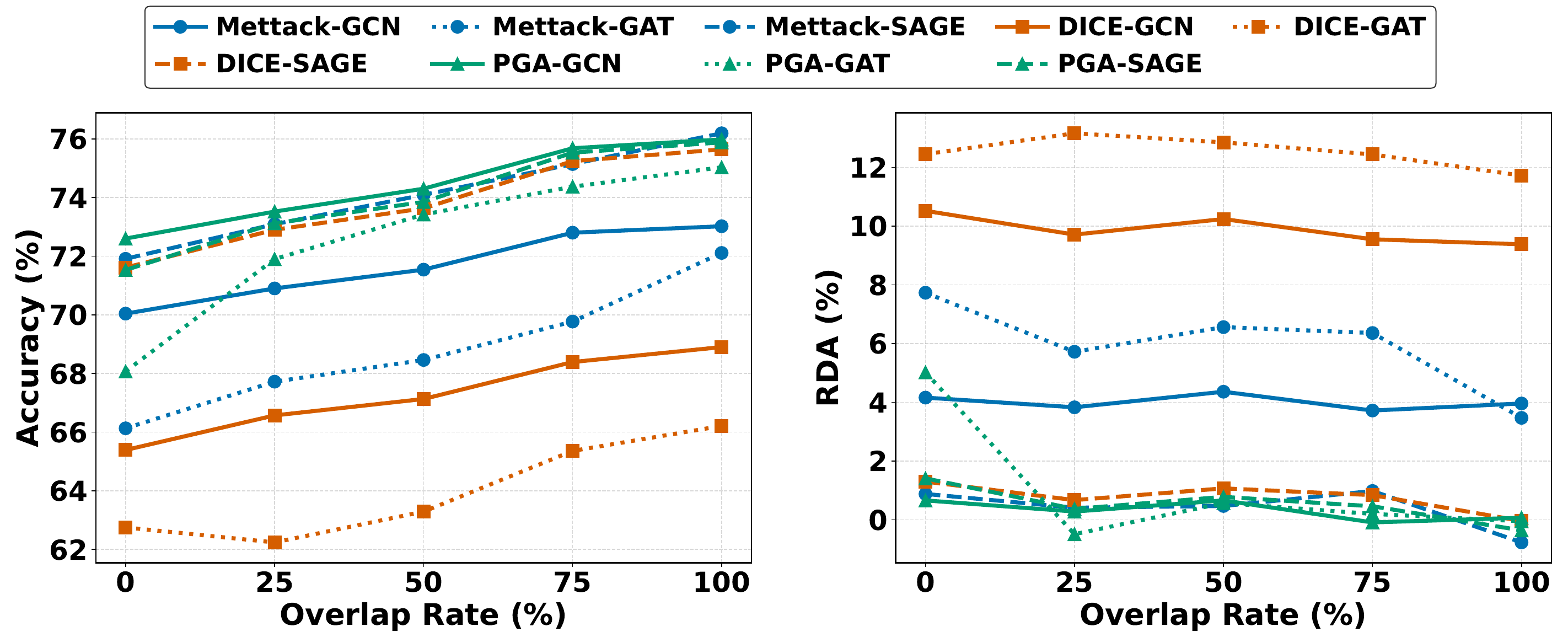}
\caption{\label{fig:overlap} Impact of training-testing data overlap on model robustness.}
\vspace{-1mm}
\end{figure}

In addition, Fig.~\ref{fig:rdarank} presents the RDA ranking of the shallow embedding-based baseline. Interestingly, the baseline performs considerably better on Tape-arxiv23 (a dataset released after the rise of LLMs) than on other datasets. One possible explanation is that LLMs/LMs may have been pretrained on Cora, Pubmed, and Ogbn-products, leading to ground truth leakage and giving LLM-enhanced models an unintended robustness advantage. To further examine whether the observed robustness gains stem from such leakage, we fine-tuned E5-Large under different overlap rates (0\%, 25\%, 50\%, 75\%, and 100\%), where the overlap rate denotes the percentage of training data derived from testing data. We then evaluated E5-Large-enhanced GNNs on Tape-arxiv23 under three non-targeted structural attacks. As shown in Fig.~\ref{fig:overlap}, increasing the overlap rate consistently improves absolute robustness (accuracy), while the relative robustness (RDA) remains largely stable across overlap rates, and robustness advantages persist even at 0\% overlap. These findings suggest that although ground truth leakage can improve absolute robustness, the overall robustness gains are not systematically driven by leakage but primarily arise from the architectural integration of LLMs/LMs.

\textbf{\underline{Key Takeaway 4:}} Among the GNN backbones, GraphSAGE exhibits the strongest robustness.

\textbf{\underline{Key Takeaway 5:}} Ground truth leakage improves absolute robustness, but the robustness gains mainly come from LLM/LM integration.

\subsection{Analysis of Structural Attacks} \label{analysis_structural}
Sections~\ref{evaluation_targeted} and~\ref{evaluation_nontargeted} have demonstrated the robustness of LLM-enhanced GNNs against structural attacks. To understand the underlying reasons, we conduct an in-depth analysis from several perspectives. Due to space constraints, we primarily present results on the Pubmed dataset.

\begin{figure}[!t]
  \centering
  \captionsetup[sub]{aboveskip=0pt,belowskip=0pt,skip=4pt}
  \subfloat[Cora]{\includegraphics[width=0.48\columnwidth]{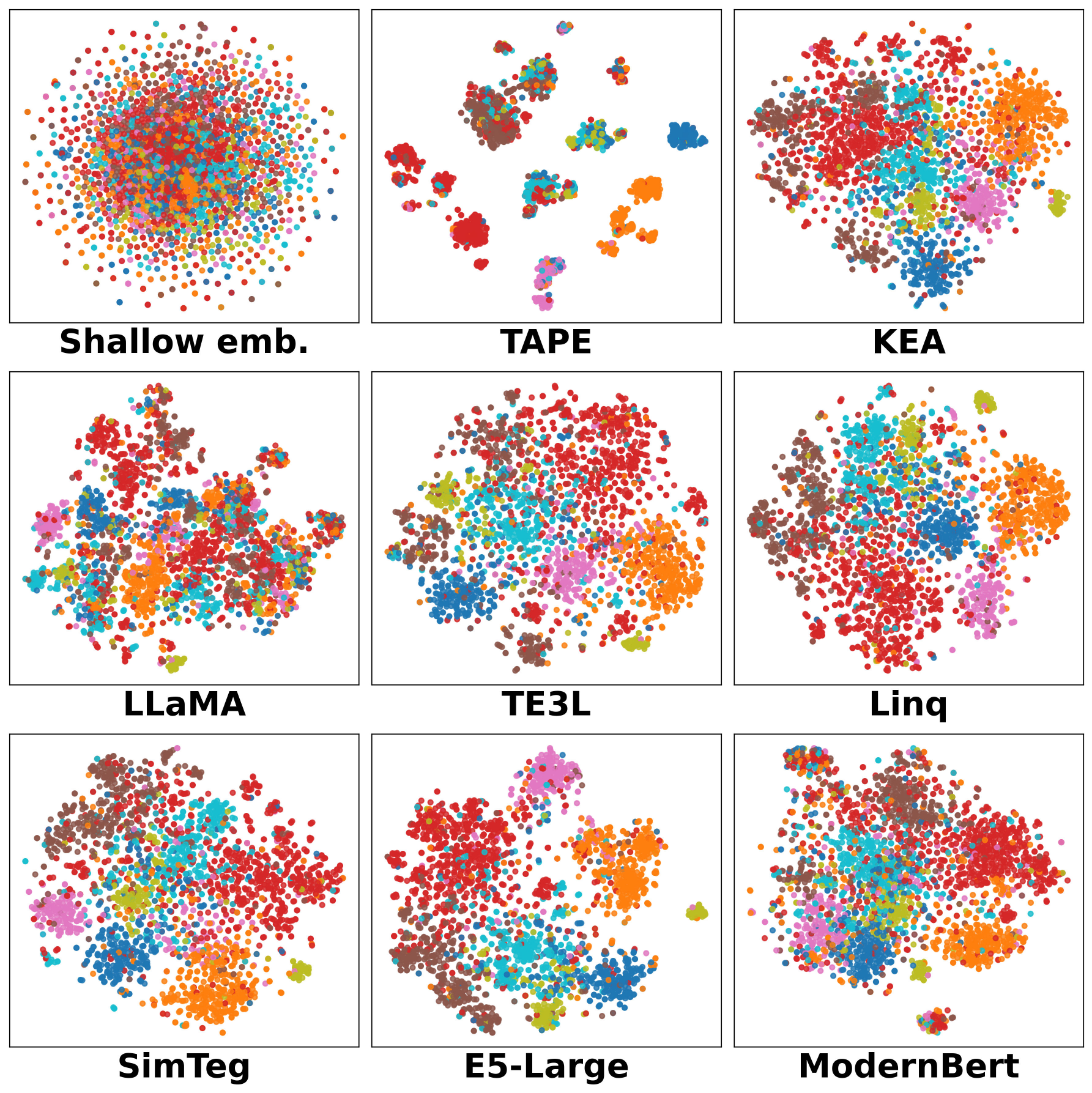}
\label{fig:tsnecora}}
 \subfloat[Pubmed]{\includegraphics[width=0.48\columnwidth]{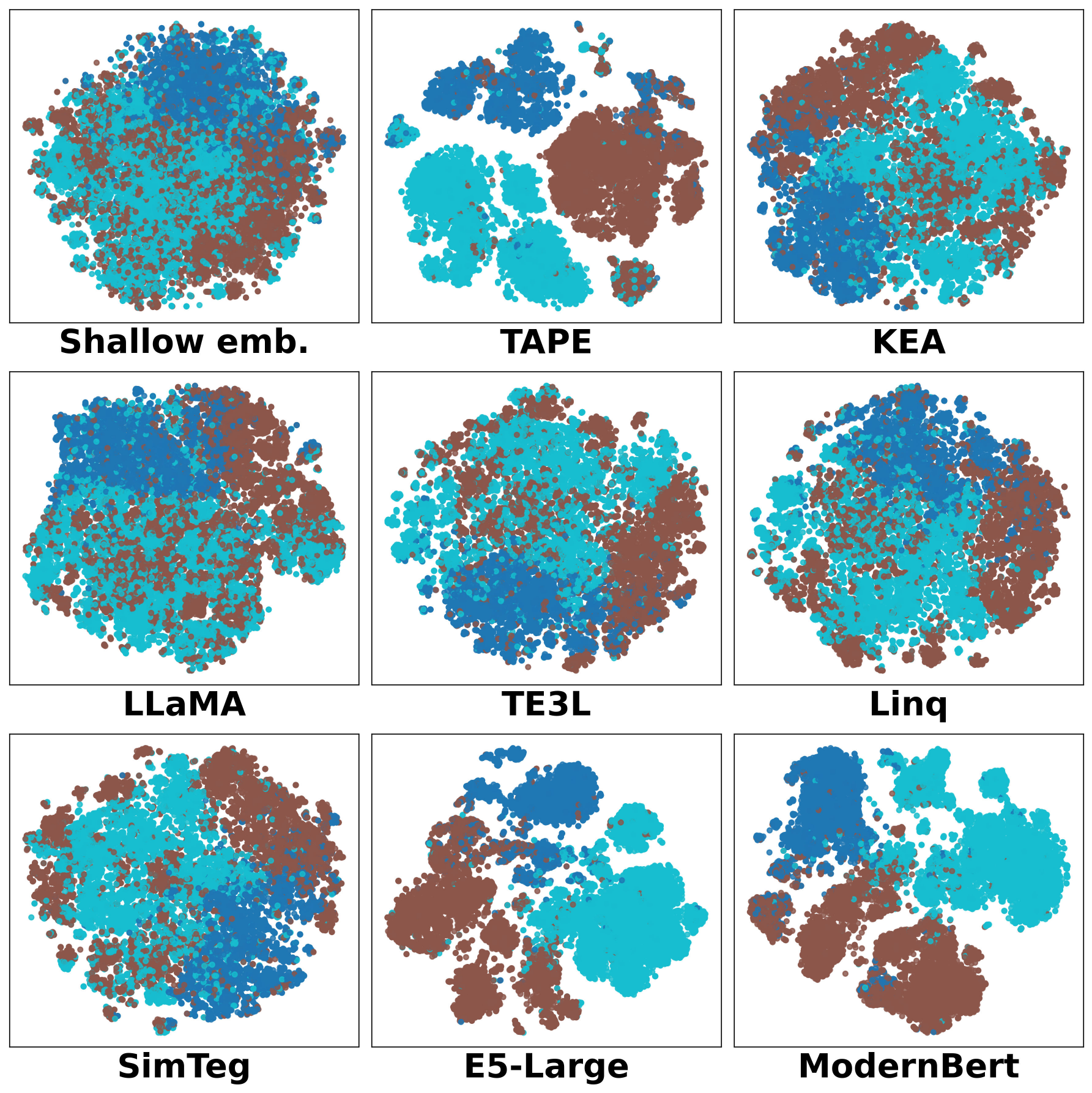}
\label{fig:tsnepubmed}}
\caption{\label{fig:tSNE} t-SNE visualizations of node embeddings on two datasets. Different colors refer to different node categories.}
\end{figure}

\textbf{Embedding Visualization.}
Fig.~\ref{fig:tSNE} shows that embeddings produced by LLMs/LMs, particularly TAPE and E5-Large, are more distinguishable in node categories than the baseline, indicating higher embedding quality. These results are consistent with our earlier findings that TAPE and E5-Large exhibit the strongest robustness. Therefore, \textit{we can conclude that model robustness is highly related to embedding quality}.

\begin{table}[!tbp]
\renewcommand\arraystretch{1.2}
\caption{Embedding quality and robustness analysis under structural attacks on the Pubmed dataset.}
\label{tab:structural-pubmed}
\resizebox{1.0\linewidth}{!}{
\begin{threeparttable}
\begin{tabular}{lcccccccc}
\toprule
\multirow{2}{*}{\textbf{Model}} &  \multirow{2}{*}{\textbf{DBI $\downarrow$}} & \multirow{2}{*}{\textbf{Sil $\uparrow$}} & \multirow{2}{*}{\textbf{Hom $\uparrow$}} & \multirow{2}{*}{\textbf{ELMI $\uparrow$}} & \multicolumn{2}{c}{\textbf{ESMI $\uparrow$}} & \multicolumn{2}{c}{\textbf{NCon $\uparrow$}} \\
\cmidrule(lr){6-7} \cmidrule(lr){8-9}
 &  &  &  &  & cle & poi & cle & poi \\
\midrule
Shallow emb. & 8.29 & 1.37 & 75.73 & 1.14 & 0.28 & 0.25 & 39.87 & 36.02 \\
\midrule
TAPE & \colorbox[HTML]{B8E7E1}{\textbf{0.57}} & \colorbox[HTML]{B8E7E1}{\textbf{64.29}} & 91.82 & \colorbox[HTML]{B8E7E1}{\textbf{57.06}} & \colorbox[HTML]{D4FAFC}{\underline{1.55}} & \colorbox[HTML]{B8E7E1}{\textbf{1.57}} & \colorbox[HTML]{D4FAFC}{\underline{95.14}} & \colorbox[HTML]{D4FAFC}{\underline{91.81}} \\
KEA & 6.19 & 2.12 & 83.12 & 2.33 & 0.33 & 0.2 & 88.47 & 87.65 \\
\midrule
LLaMA & 5.43 & 1.18 & 76.15 & 3.65 & 0.63 & 0.58 & 78.96 & 77.03 \\
TE3L & 5.52 & 2.55 & 82.29 & 2.80 & 0.47 & 0.45 & 67.51 & 62.74 \\
Linq & 5.55 & 2.47 & 82.69 & 2.40 & 0.42 & 0.14 & 67.35 & 63.08 \\
\midrule
SimTeg & 5.28 & 2.85 & 83.49 & 2.84 & 0.52 & 0.49 & 88.15 & 86.48 \\
E5-Large & \colorbox[HTML]{D4FAFC}{\underline{0.63}} & \colorbox[HTML]{D4FAFC}{\underline{60.22}} & \colorbox[HTML]{B8E7E1}{\textbf{93.39}} & \colorbox[HTML]{D4FAFC}{\underline{53.37}} & \colorbox[HTML]{B8E7E1}{\textbf{1.66}} & \colorbox[HTML]{D4FAFC}{\underline{1.49}} & 92.68 & 88.16 \\
ModernBert & 1.94 & 18.89 & \colorbox[HTML]{D4FAFC}{\underline{93.10}} & 16.74 & 0.65 & 0.65 & \colorbox[HTML]{B8E7E1}{\textbf{99.58}} & \colorbox[HTML]{B8E7E1}{\textbf{99.51}} \\
\bottomrule
\end{tabular}

\begin{tablenotes}[para,flushleft]
    \item[] DBI: Davies-Bouldin Index; Sil: Silhouette Score; Hom: Embedding Homophily: ELMI: Embedding-Label Mutual Information; ESMI: Embedding-Structural Mutual Information; NCon: Neighbor Consistency; cle: Clean; poi: Poisoned.
\end{tablenotes}
\end{threeparttable}
}
\end{table}

\textbf{Embedding Separability.} 
Table~\ref{tab:structural-pubmed} shows that LLM/LM-based feature enhancers consistently deliver better clustering quality than the baseline, suggesting that their generated embeddings are more separable than shallow embeddings. This observation also explains why LLM-enhanced GNNs exhibit strong robustness to structural perturbations, as separable embeddings reduce the likelihood of perturbed nodes being misclassified into incorrect categories. 

\textbf{Preservation of Label Information.} 
As shown in Table~\ref{tab:structural-pubmed}, embeddings generated by LLMs/LMs preserve more label information than shallow embeddings. Consequently, GNNs equipped with these semantically rich embeddings can perform node classification effectively. This result also suggests that LLMs/LMs can serve as annotators, providing pseudo-labels to guide GNN training in scenarios with limited supervision.

\textbf{Preservation of Structural Information.}
Table~\ref{tab:structural-pubmed} reports embedding-structural mutual information and neighbor consistency before and after the Mettack attack at a high perturbation level. On clean graphs, LLM/LM-based enhancers retain more structural information than the baseline. Under the attack, all models exhibit reduced values, indicating that Mettack destroys structural properties; however, LLM/LM-based enhancers still maintain higher levels of structural preservation. \textit{These findings suggest that preserving structural properties is key to robustness.}

\textbf{\underline{Key Takeaway 6:}} LLM/LM-based feature enhancers produce high-quality embeddings that capture label and structural information more effectively than the baseline, contributing to their strong robustness.

\textbf{\underline{Key Takeaway 7:}} The success of structural attacks stems from their ability to destroy structural properties (e.g., neighbor consistency).

\subsection{Results on Textual Attacks} \label{evaluation_textual}
\textbf{Attack Settings.} We employed three textual attacks: DWord~\cite{gao2018black}, BertAtk~\cite{li2020bert}, and MAYA~\cite{chen2021multi}, which operate at the character, word, and sentence levels, respectively, to perturb textual attributes in the training sets of the four datasets. The perturbed texts were encoded as embeddings using either LLMs/LMs or shallow embedding methods and then combined with clean graph structures to train GNN models. Finally, model robustness was evaluated on the full test set in terms of ACC and RDA.

\textbf{Robustness against Textual Attacks.} 
We present the results under the MAYA attack, which is the most disruptive among the three attacks, and provide the results under DWord and BertAtk in Appendix~\ref{appendix_textual}. From Tables~\ref{tab:text-maya},~\ref{tab:text-dword}, and~\ref{tab:text-bertatk}, we draw the following observations: (1) LLM-enhanced GNN models are still more robust than the baseline. This is largely because the scale of LLMs/LMs makes them hard to be attacked successfully, thus making the generated embeddings robust. (2) Textual attacks are not as effective as structural attacks. All models exhibit relatively lower RDA under textual perturbations compared to structural ones. For example, the maximum RDA for the baseline is 7.93\% under textual attacks, while it is 15.49\% under the Mettack attack (a non-targeted structural attack). This suggests that the impact of textual perturbations might be mitigated by GNNs owing to their message passing mechanisms. This result is consistent with prior studies that structural attacks are more effective than feature attacks~\cite{zugner2018adversarial,li2021adversarial}. (3) In several cases, RDA takes negative values, indicating that models may even benefit from textual perturbations. For example, under the DWord attack, ModernBert-enhanced SAGE achieves an RDA value of -3.44 on the Cora dataset, indicating that its performance is significantly better than that on clean graphs. This inspires us that introducing reasonable perturbations may serve as a form of data augmentation for training robust GNN models. (4) The choice of GNN backbones has limited impact. Unlike structural attacks, which primarily disrupt the message-passing mechanism of GNNs, textual perturbations alter semantic features captured during textual encoding, leaving backbone choice less consequential.

\textbf{\underline{Key Takeaway 8:}} Textual attacks are substantially less effective than structural ones, with sentence-level attacks being the most effective. LLM-enhanced GNNs still exhibit greater robustness than the baseline under these attacks.

\textbf{\underline{Key Takeaway 9:}} In several cases, LLM-enhanced GNNs even benefit from textual perturbations.

\textbf{\underline{Key Takeaway 10:}} The type of GNN backbones has a minor impact on model robustness under textual attacks.

\begin{table*}[!t]
\renewcommand\arraystretch{1.2}
\caption{Robustness comparison against the \textbf{MAYA} attack.}
\label{tab:text-maya}
\resizebox{1.0\textwidth}{!}{
\begin{tabular}{ccccccccccccccccccccccccc}
\toprule
\multirow{4}{*}{\textbf{Model}} & \multicolumn{6}{c}{\textbf{Cora}} & \multicolumn{6}{c}{\textbf{Pubmed}} & \multicolumn{6}{c}{\textbf{Ogbn-products}} & \multicolumn{6}{c}{\textbf{Tape-arxiv23}}  \\
\cmidrule(lr){2-7} \cmidrule(lr){8-13} \cmidrule(lr){14-19} \cmidrule(lr){20-25} 
& \multicolumn{2}{c}{GCN} & \multicolumn{2}{c}{GAT} &  \multicolumn{2}{c}{SAGE} & \multicolumn{2}{c}{GCN} & \multicolumn{2}{c}{GAT} &  \multicolumn{2}{c}{SAGE} & \multicolumn{2}{c}{GCN} & \multicolumn{2}{c}{GAT} &  \multicolumn{2}{c}{SAGE} & \multicolumn{2}{c}{GCN} & \multicolumn{2}{c}{GAT} &  \multicolumn{2}{c}{SAGE} \\
\cmidrule(lr){2-3} \cmidrule(lr){4-5} \cmidrule(lr){6-7} \cmidrule(lr){8-9} \cmidrule(lr){10-11} \cmidrule(lr){12-13} \cmidrule(lr){14-15} \cmidrule(lr){16-17} \cmidrule(lr){18-19}\cmidrule(lr){20-21}\cmidrule(lr){22-23}\cmidrule(lr){24-25}
& ACC $\uparrow$ & RDA $\downarrow$  & ACC $\uparrow$ & RDA $\downarrow$  & ACC $\uparrow$ & RDA $\downarrow$  & ACC $\uparrow$ & RDA $\downarrow$ & ACC $\uparrow$ & RDA $\downarrow$ & ACC $\uparrow$ & RDA $\downarrow$ & ACC $\uparrow$ & RDA $\downarrow$ & ACC $\uparrow$ & RDA $\downarrow$ & ACC $\uparrow$ & RDA $\downarrow$ & ACC $\uparrow$ & RDA $\downarrow$ & ACC $\uparrow$ & RDA $\downarrow$ & ACC $\uparrow$ & RDA $\downarrow$ \\
\midrule
Shallow emb. &80.54 & 0.59 & 78.11 & 2.35 & 79.75 & 0.35 & 85.54 & 0.78 & 85.26 & \colorbox[HTML]{D4FAFC}{\underline{0.27}} & 83.81 & \colorbox[HTML]{D4FAFC}{\underline{1.83}} & 65.03 & 1.39 & 67.96 & 2.06 & 60.82 & 2.36 & 53.14 & 7.93 & 51.56 & 7.05 & 52.74 & 6.36 \\

\midrule
TAPE &82.55 & \colorbox[HTML]{B8E7E1}{\textbf{-0.99}} & 80.33 & 0.19 & 80.20 & 1.98 & \colorbox[HTML]{B8E7E1}{\textbf{91.75}} & 1.47 & \colorbox[HTML]{B8E7E1}{\textbf{88.96}} & 0.90 & \colorbox[HTML]{D4FAFC}{\underline{92.30}} & 2.37 & \colorbox[HTML]{B8E7E1}{\textbf{85.48}} & 0.42 & \colorbox[HTML]{B8E7E1}{\textbf{85.23}} & 0.15 & \colorbox[HTML]{B8E7E1}{\textbf{85.44}} & 0.71 & \colorbox[HTML]{D4FAFC}{\underline{71.33}} & 4.22 & \colorbox[HTML]{B8E7E1}{\textbf{71.50}} & 3.95 & \colorbox[HTML]{D4FAFC}{\underline{70.18}} & 5.60 \\

KEA &\colorbox[HTML]{D4FAFC}{\underline{84.13}} & 0.08 & \colorbox[HTML]{B8E7E1}{\textbf{84.30}} & 0.20 & 83.97 & 1.04 & 87.83 & 1.28 & 86.33 & \colorbox[HTML]{B8E7E1}{\textbf{0.25}} & 86.83 & 2.78 & 83.73 & \colorbox[HTML]{D4FAFC}{\underline{0.19}} & 82.69 & 0.57 & 83.72 & \colorbox[HTML]{D4FAFC}{\underline{0.65}} & 65.77 & 4.47 & 66.26 & 5.17 & 65.76 & 5.08 \\

\midrule
LLaMA &81.66 & -0.02 & 80.56 & 1.27 & 79.73 & 1.85 & 87.05 & 0.93 & 84.42 & 0.65 & 86.30 & 1.91 & 82.75 & 1.14 & 80.55 & \colorbox[HTML]{B8E7E1}{\textbf{-0.16}} & 82.30 & 1.35 & 68.72 & 3.90 & 68.44 & \colorbox[HTML]{D4FAFC}{\underline{2.84}} & 67.79 & \colorbox[HTML]{D4FAFC}{\underline{4.66}} \\

TE3L &83.35 & 0.35 & 81.92 & 1.30 & 83.59 & 0.36 & 87.74 & 1.15 & 86.49 & 0.69 & 87.00 & 2.29 & 83.57 & 0.61 & \colorbox[HTML]{D4FAFC}{\underline{82.82}} & \colorbox[HTML]{D4FAFC}{\underline{0.01}} & \colorbox[HTML]{D4FAFC}{\underline{83.98}} & 0.67 & 67.44 & 3.89 & 65.64 & 4.59 & 66.09 & 5.40 \\

Linq &\colorbox[HTML]{B8E7E1}{\textbf{84.18}} & 0.13 & 82.63 & -0.07 & \colorbox[HTML]{D4FAFC}{\underline{84.11}} & 0.37 & 88.81 & 0.98 & 86.37 & 0.90 & 88.56 & 3.00 & \colorbox[HTML]{D4FAFC}{\underline{84.13}} & \colorbox[HTML]{B8E7E1}{\textbf{0.14}} & 81.74 & 0.60 & 83.89 & 1.00 & 70.25 & \colorbox[HTML]{D4FAFC}{\underline{2.86}} & 69.30 & 3.10 & 68.57 & 5.46 \\

\midrule
SimTeg &83.28 & 0.01 & \colorbox[HTML]{D4FAFC}{\underline{83.58}} & \colorbox[HTML]{D4FAFC}{\underline{-0.83}} & 83.66 & 0.13 & \colorbox[HTML]{D4FAFC}{\underline{91.04}} & \colorbox[HTML]{B8E7E1}{\textbf{-1.97}} & \colorbox[HTML]{D4FAFC}{\underline{88.41}} & 5.00 & \colorbox[HTML]{B8E7E1}{\textbf{92.90}} & \colorbox[HTML]{B8E7E1}{\textbf{-0.35}} & 82.76 & 0.66 & 82.16 & 0.88 & 83.05 & \colorbox[HTML]{B8E7E1}{\textbf{0.63}} & \colorbox[HTML]{B8E7E1}{\textbf{71.89}} & \colorbox[HTML]{B8E7E1}{\textbf{0.17}} & \colorbox[HTML]{D4FAFC}{\underline{71.20}} & \colorbox[HTML]{B8E7E1}{\textbf{-0.32}} & \colorbox[HTML]{B8E7E1}{\textbf{71.71}} & \colorbox[HTML]{B8E7E1}{\textbf{-0.07}} \\

E5-Large &83.38 & 1.22 & 81.96 & 0.05 & \colorbox[HTML]{B8E7E1}{\textbf{84.16}} & \colorbox[HTML]{D4FAFC}{\underline{-0.44}} & 88.73 & 4.65 & 86.58 & 3.64 & 88.81 & 5.66 & 84.03 & 1.09 & 82.46 & 1.88 & 83.85 & 1.34 & 69.69 & 4.64 & 67.85 & 5.33 & 68.32 & 5.83 \\

ModernBert &77.77 & \colorbox[HTML]{D4FAFC}{\underline{-0.08}} & 77.01 & \colorbox[HTML]{B8E7E1}{\textbf{-2.15}} & 75.18 & \colorbox[HTML]{B8E7E1}{\textbf{-0.53}} & 89.70 & \colorbox[HTML]{D4FAFC}{\underline{0.73}} & 85.66 & 9.18 & 89.63 & 4.51 & 82.66 & 1.18 & 79.60 & 1.55 & 81.84 & 2.46 & 65.94 & 6.24 & 64.25 & 6.19 & 65.92 & 6.46 \\

\bottomrule
\end{tabular}}
\vspace{-2mm}
\end{table*}

\subsection{Analysis of Textual Attacks} \label{analysis_textual}
We analyze why LLM-enhanced GNNs are robust to textual attacks, focusing on the three strongest LLM/LM-based feature enhancers (i.e., TAPE, SimTeg, E5-Large) and the weakest one (i.e., ModernBert) on the Pubmed dataset, using embedding visualization, the Davies-Bouldin Index (DBI), embedding Homophily (Hom), and Embedding-Structural Mutual Information (ESMI).

\begin{figure}[!t]
  \centering
    \includegraphics[width=\linewidth]{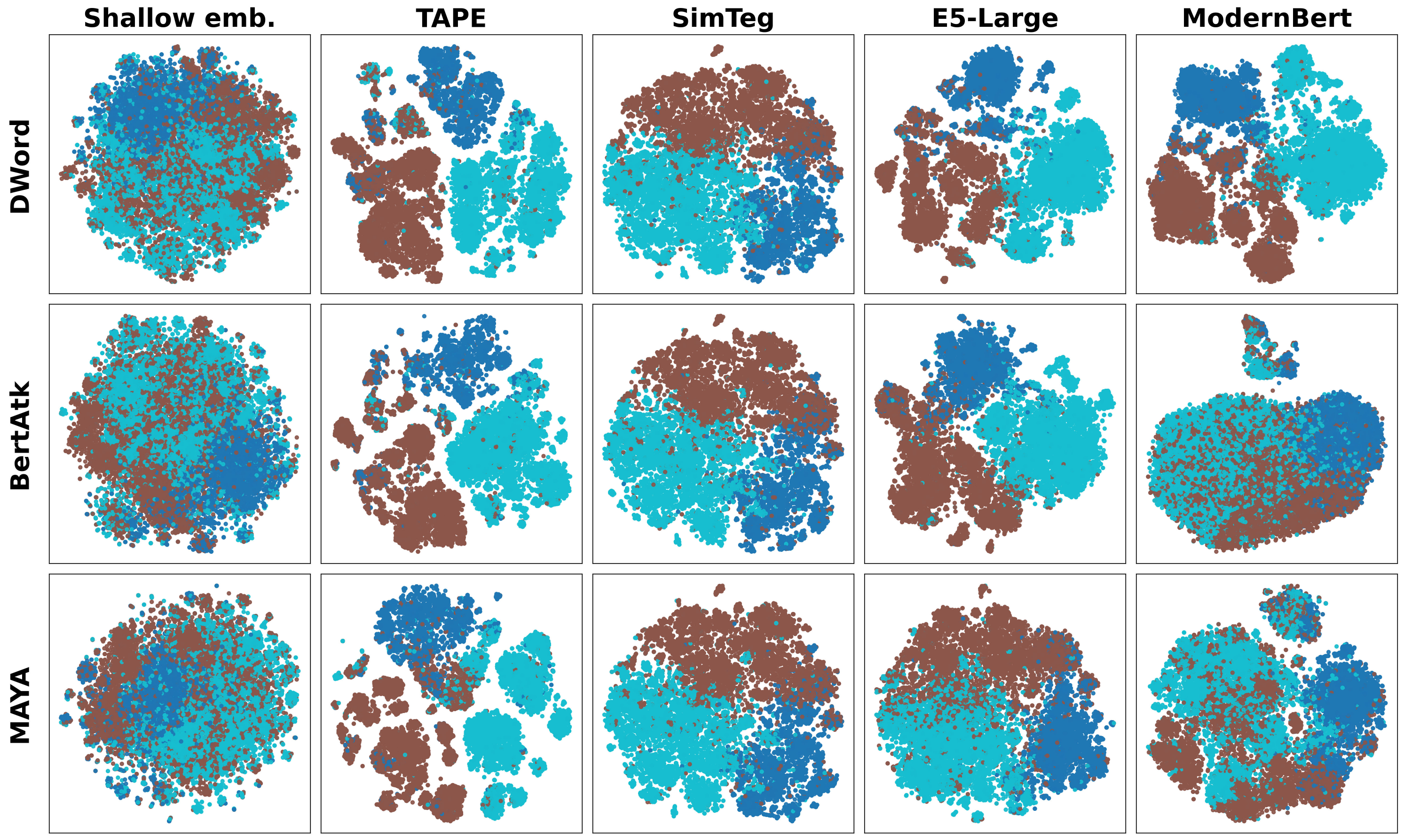}
\caption{t-SNE visualizations of node embeddings on the Pubmed dataset under textual attacks.}
\label{fig:tsnepubmedtext}
\vspace{-2mm}
\end{figure}

\textbf{Embedding Visualization.} As shown in Fig.~\ref{fig:tsnepubmedtext}, embeddings generated by TAPE, SimTeg, and E5-Large form more separable clusters than the others, consistent with their strong robustness. In contrast, shallow embeddings and ModernBert exhibit poor separation, making them vulnerable to attacks.

\begin{figure}[!t]
\centering
\includegraphics[width=\linewidth]{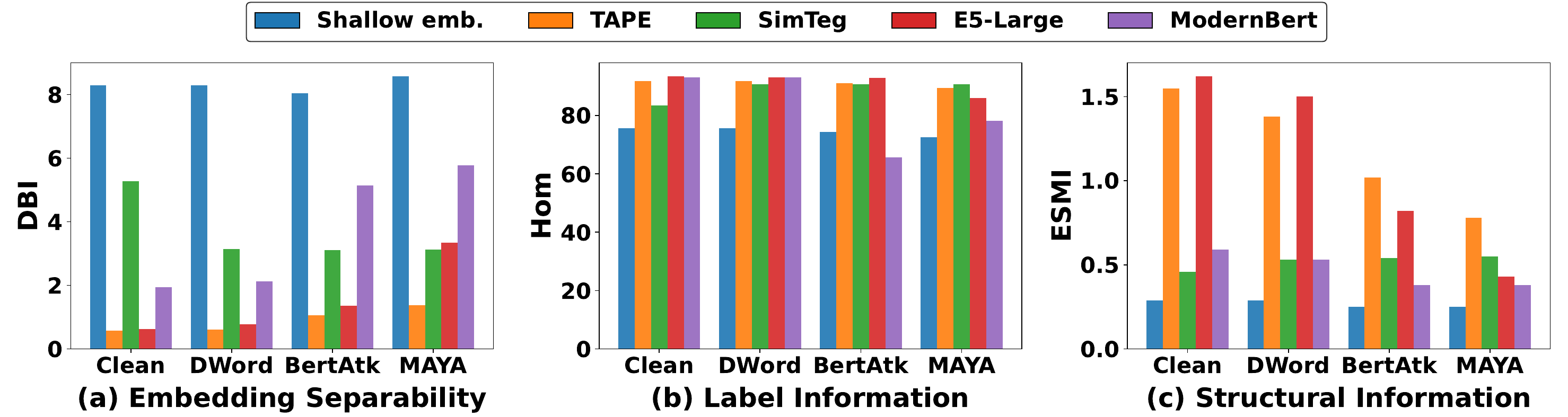}
\caption{Robustness analysis under textual attacks on the Pubmed dataset.}
\label{fig:textRDA}
\vspace{-1mm}
\end{figure}

\textbf{Embedding Separability.}
As shown in Fig.~\ref{fig:textRDA}, the shallow embeddings consistently yield the highest DBI, indicating their poor quality. In contrast, TAPE achieves the lowest DBI, which aligns with its superior robustness. Notably, the DBI value of TAPE remains low and stable across textual attacks, suggesting that perturbations to textual attributes do not fundamentally disrupt its embedding separability. This explains TAPE's robustness: its embedding space remains well-structured, reducing the likelihood of misclassification.

\textbf{Preservation of Label Information.}
As shown in Fig.~\ref{fig:textRDA}, embeddings generated by LLMs/LMs preserve significantly higher homophily compared to shallow embeddings, indicating their effective encoding of label semantics. Additionally, homophily remains high in both clean and perturbed settings, implying that textual attacks do not significantly affect the semantics of textual attributes.

\textbf{Preservation of Structural Information.}
As shown in Fig.~\ref{fig:textRDA}, ESMI drops significantly for some models, showing that textual perturbations can partially weaken the encoding of structural information. However, this decline is much smaller than that from structural attacks, explaining the diminished effectiveness of textual attacks. 

\textbf{\underline{Key Takeaway 11:}} Key embedding properties of several models remain stable across clean and perturbed settings, confirming the limited effectiveness of textual attacks.

\begin{table*}[!t]
\renewcommand\arraystretch{1.2}
\caption{Accuracy (\%) under a combined structural and textual attack.}
\label{tab:text-struct}
\resizebox{1.0\textwidth}{!}{
\begin{tabular}{@{}cccccccccccccc@{}}
\toprule
\multirow{2}{*}{{\textbf{Structural}}} 
 &
 \multirow{2}{*}{{\textbf{Textual}}} 

   &
  \multicolumn{3}{c}{\textbf{Cora}} &
  \multicolumn{3}{c}{\textbf{Pubmed}} & \multicolumn{3}{c}{\textbf{Ogbn-products}} & \multicolumn{3}{c}{\textbf{Tape-arxiv23}}\\ 
  \cmidrule(l){3-5} \cmidrule(l){6-8} \cmidrule(l){9-11} \cmidrule(l){12-14} &
   &
   Shallow emb. & TAPE & E5-Large &
   Shallow emb. & TAPE & E5-Large &
   Shallow emb. & TAPE & E5-Large &
   Shallow emb. & TAPE & E5-Large \\ \midrule

\multirow{4}{*}{{Mettack}} 
&
  Non-textual &
  74.60 &78.02 &80.60&
  74.12 &84.91 &84.71&
  63.78 &84.26 &83.23&
  51.03 &71.33 &70.04\\
&
  DWord &
$74.90_{\inc{0.30}}$ & $77.90_{\dec{0.12}}$ & $80.63_{\inc{0.03}}$ & $73.94_{\dec{0.18}}$ & $84.57_{\dec{0.34}}$ & $84.59_{\dec{0.12}}$ & $64.10_{\inc{0.32}}$ & $84.32_{\inc{0.06}}$ & $82.97_{\dec{0.26}}$ & $51.07_{\inc{0.04}}$ & $71.42_{\inc{0.09}}$ & $69.71_{\dec{0.33}}$ \\
  &
  BertAtk &
$72.63_{\dec{1.97}}$ & $77.39_{\dec{0.63}}$ & $78.16_{\dec{2.44}}$ & $73.00_{\dec{1.12}}$ & $82.52_{\dec{2.39}}$ & $81.65_{\dec{3.06}}$ & $61.76_{\dec{2.02}}$ & $83.98_{\dec{0.28}}$ & $82.50_{\dec{0.73}}$ & $49.83_{\dec{1.20}}$ & $69.68_{\dec{1.65}}$ & $68.96_{\dec{1.08}}$ \\
    &
  MAYA &
$74.35_{\dec{0.25}}$ & $78.40_{\inc{0.38}}$ & $80.32_{\dec{0.28}}$ & $73.39_{\dec{0.73}}$ & $81.24_{\dec{3.67}}$ & $77.08_{\dec{7.63}}$ & $62.24_{\dec{1.54}}$ & $84.01_{\dec{0.25}}$ & $82.52_{\dec{0.71}}$ & $49.11_{\dec{1.92}}$ & $68.71_{\dec{2.62}}$ & $65.47_{\dec{4.57}}$ \\

\midrule
\multirow{4}{*}{{DICE}}
&
  Non-textual &
    68.63  &70.82  &76.31 &
  74.36  &85.73  &85.48 &
  55.41  &81.03  &79.83 &
  50.10  &67.13  &65.39\\

&
  DWord &
$68.63_{\dec{0.00}}$ & $70.52_{\dec{0.30}}$ & $75.48_{\dec{0.83}}$ & $74.34_{\dec{0.02}}$ & $85.52_{\dec{0.21}}$ & $85.26_{\dec{0.22}}$ & $55.02_{\dec{0.39}}$ & $81.17_{\inc{0.14}}$ & $79.51_{\dec{0.32}}$ & $50.83_{\inc{0.73}}$ & $66.93_{\dec{0.20}}$ & $65.39_{\dec{0.00}}$ \\

  &
  BertAtk &
$66.45_{\dec{2.18}}$ & $72.33_{\inc{1.51}}$ & $73.36_{\dec{2.95}}$ & $73.66_{\dec{0.70}}$ & $84.47_{\dec{1.26}}$ & $84.22_{\dec{1.26}}$ & $53.22_{\dec{2.19}}$ & $80.78_{\dec{0.25}}$ & $79.15_{\dec{0.68}}$ & $48.95_{\dec{1.15}}$ & $65.88_{\dec{1.25}}$ & $64.54_{\dec{0.85}}$ \\

    &
  MAYA &
$67.60_{\dec{1.03}}$ & $71.30_{\inc{0.48}}$ & $75.03_{\dec{1.28}}$ & $73.34_{\dec{1.02}}$ & $83.70_{\dec{2.03}}$ & $78.54_{\dec{6.94}}$ & $53.98_{\dec{1.43}}$ & $80.93_{\dec{0.10}}$ & $78.90_{\dec{0.93}}$ & $47.70_{\dec{2.40}}$ & $64.06_{\dec{3.07}}$ & $61.70_{\dec{3.69}}$ \\

   \midrule
\multirow{4}{*}{{PGA}}
&
  Non-textual &
  71.89  &75.71  &78.55 &
  86.22  &92.82  &92.84 &
  65.24  &84.57  &83.74 &
  56.39  &72.89  &72.60\\
&
  DWord &
$71.23_{\dec{0.66}}$ & $75.25_{\dec{0.46}}$ & $78.47_{\dec{0.08}}$ & $85.92_{\dec{0.30}}$ & $92.54_{\dec{0.28}}$ & $92.91_{\inc{0.07}}$ & $64.90_{\dec{0.31}}$ & $84.52_{\dec{0.05}}$ & $83.29_{\dec{0.45}}$ & $55.91_{\dec{0.48}}$ & $72.85_{\dec{0.04}}$ & $71.78_{\dec{0.82}}$ \\

  &
  BertAtk &
$69.83_{\dec{2.06}}$ & $76.11_{\inc{0.40}}$ & $75.90_{\dec{2.65}}$ & $85.80_{\dec{0.42}}$ & $91.74_{\dec{1.08}}$ & $92.41_{\dec{0.43}}$ & $63.30_{\dec{1.94}}$ & $84.33_{\dec{0.24}}$ & $82.84_{\dec{0.90}}$ & $55.04_{\dec{1.35}}$ & $70.47_{\dec{2.42}}$ & $71.12_{\dec{1.48}}$ \\

    &
  MAYA &
$71.55_{\dec{0.34}}$ & $75.80_{\inc{0.09}}$ & $77.87_{\dec{0.68}}$ & $85.72_{\dec{0.50}}$ & $91.46_{\dec{1.36}}$ & $88.32_{\dec{4.52}}$ & $64.26_{\dec{0.98}}$ & $84.47_{\dec{0.10}}$ & $82.88_{\dec{0.86}}$ & $53.24_{\dec{3.15}}$ & $69.11_{\dec{3.78}}$ & $68.43_{\dec{4.17}}$ \\

   \midrule   
\end{tabular}
}
\begin{tablenotes}[para,flushleft]
\item[]\begin{minipage}{\textwidth}
The symbols $\dec{}$ and $\inc{}$ denote the percentage decrease and percentage increase, respectively, relative to non-targeted structural attacks without additional textual perturbations.
\end{minipage}
\end{tablenotes}
\end{table*}

\begin{table*}[!t]
\renewcommand\arraystretch{1.2}
\caption{Accuracy (\%) after purifying perturbed graph structures.}
\label{tab:cleangraph}
\resizebox{1.0\textwidth}{!}{
\begin{tabular}{@{}cccccccccccccccccc@{}}
\toprule
\multirow{2}{*}{{\textbf{Structural}}} 
 &
 \multirow{2}{*}{{\textbf{Textual}}} 
   &
  \multicolumn{4}{c}{\textbf{Cora}} &
  \multicolumn{4}{c}{\textbf{Pubmed}} & \multicolumn{4}{c}{\textbf{Ogbn-products}} & \multicolumn{4}{c}{\textbf{Tape-arxiv23}}\\ 
  \cmidrule(l){3-6} \cmidrule(l){7-10} \cmidrule(l){11-14} \cmidrule(l){15-18} &
   &
   RGCN &SimpGCN & TAPE & E5-Large &
   RGCN &SimpGCN & TAPE & E5-Large &
   RGCN &SimpGCN & TAPE & E5-Large &
   RGCN &SimpGCN & TAPE & E5-Large \\ \midrule
 &
  clean &
  81.02 & 81.02 &81.74  &84.41 &
  86.21 & 86.21 &93.12  &87.15 &
  65.95 & 65.95 &85.85  &84.96 &
  57.72 & 57.72 &74.47  &73.08\\
\midrule
\multirow{2}{*}{{Mettack}} 
&
  attack &
  74.60 & 74.60 &78.02 &80.60&
  74.12 & 74.12 &84.91 &84.71&
  63.78 & 63.78 &84.26 &83.23&
  51.03 & 51.03 &71.33 &70.04\\
  &
  defense &
$77.40_{\inc{2.80}}$ & $42.91_{\dec{31.69}}$ & $77.89_{\dec{0.13}}$ & $81.65_{\inc{1.05}}$ &
$71.51_{\dec{2.61}}$ & $84.28_{\inc{10.16}}$ & $94.23_{\inc{9.32}}$ & $94.03_{\inc{9.32}}$ & 
$45.39_{\dec{18.39}}$ & $58.45_{\dec{5.33}}$ & $86.26_{\inc{2.00}}$ & $84.30_{\inc{1.07}}$ & 
$42.89_{\dec{8.14}}$ & $47.37_{\dec{3.66}}$ & $74.90_{\inc{3.57}}$ & $73.09_{\inc{3.05}}$ \\
\midrule
\multirow{2}{*}{{DICE}}
&
  attack &
  68.63 & 68.63 &70.82  &76.31 &
  74.36 & 74.36 &85.73  &85.48 &
  55.41 & 55.41 &81.03  &79.83 &
  50.10 & 50.10 &67.13  &65.39\\
  &
  defense &
$71.33_{\inc{2.70}}$ & $40.36_{\dec{28.27}}$ & $71.68_{\inc{0.86}}$ & $77.26_{\inc{0.95}}$ & 
$75.72_{\inc{1.36}}$ & $85.74_{\inc{11.38}}$ & $94.21_{\inc{8.48}}$ & $93.85_{\inc{8.37}}$ & 
$40.15_{\dec{15.26}}$ & $51.68_{\dec{3.73}}$ & $84.87_{\inc{3.84}}$ & $83.65_{\inc{3.82}}$ & 
$43.19_{\dec{6.91}}$ & $49.22_{\dec{0.88}}$ & $74.73_{\inc{7.60}}$ & $72.06_{\inc{6.67}}$ \\
   \midrule
\multirow{2}{*}{{PGA}}
&
  attack &
  71.89 &71.89 &75.71  &78.55 &
  85.87 &85.87 &92.82  &92.84 &
  65.24 &65.24 &84.57  &83.74 &
  56.39 &56.39 &72.89  &72.60\\
  &
  defense &
$73.96_{\inc{2.07}}$ & $40.54_{\dec{31.35}}$ & $75.97_{\inc{0.26}}$ & $79.74_{\inc{1.19}}$ & 
$86.28_{\inc{0.41}}$ & $88.12_{\inc{2.25}}$ &  $94.20_{\inc{1.38}}$ & $94.15_{\inc{1.31}}$ & 
$47.84_{\dec{17.40}}$ & $59.78_{\dec{5.46}}$ &$86.07_{\inc{1.50}}$ & $84.23_{\inc{0.49}}$ & 
$43.58_{\dec{12.81}}$ & $48.60_{\dec{7.79}}$ & $74.77_{\inc{1.88}}$ & $73.08_{\inc{0.48}}$ \\

\midrule
\end{tabular}
}

\begin{tablenotes}[para,flushleft]
\item[]\begin{minipage}{\textwidth}
The symbols $\dec{}$ and $\inc{}$ denote the percentage decrease and percentage increase, respectively, relative to non-targeted structural attacks without defense.
    \end{minipage}
\end{tablenotes}
\vspace{-1mm}
\end{table*}

\section{Future Research Directions} \label{future_direction}
In this section, we outline several future research directions from both offensive and defensive perspectives.

\textbf{Offensive Perspectives.}
(1) \textit{LLMs as knowledge augmenters:} Most existing attacks assume a gray-box setting, where the adversary has access to both the full graph structure and node attributes. However, this assumption does not hold in privacy-sensitive applications. For example, in social networks, complete friendship links are often inaccessible due to privacy policies, while node textual attributes (e.g., usernames and profile descriptions) are relatively easy to obtain. In such cases, the adversary only possesses partial structural information, which reduces the effectiveness of conventional attacks and remains largely underexplored in prior work. Motivated by embedding visualizations, we observe that LLM/LM-generated embeddings exhibit strong clustering patterns, where nodes with similar semantic content are positioned closely in the embedding space. This property allows the adversary to leverage LLMs as link predictors, inferring missing edges based on embedding proximity to complement partial structural knowledge. Such attacks would be more practical than conventional ones.

(2) \textit{More stealthy structural attacks:} 
As discussed in Section~\ref{analysis_structural}, structural attacks substantially alter the structural properties of original graphs, creating detectable anomalies that defenders can exploit. A promising direction for future research is to design attacks that preserve key structural properties (e.g., clustering coefficient) by incorporating them as constraints, thereby improving stealthiness. However, attack efficiency must be carefully balanced to ensure that such attacks remain practical.

(3) \textit{More effective textual attacks:} 
Textual attacks are stealthy since they do not significantly affect textual semantics and structural properties. However, they are relatively ineffective and are often mitigated by GNNs themselves. Therefore, developing textual attacks that are both effective and efficient remains an important direction for future research, particularly given the high computational cost of using LLMs/LMs as surrogate models.

(4) \textit{Combining textual and structural attacks:} 
Adversaries can fully exploit the expanded attack surface of LLM-enhanced GNNs. One straightforward strategy is to launch combined textual and structural attacks. Table~\ref{tab:text-struct} reports model accuracy under high-perturbation non-targeted structural attacks combined with three types of textual attacks. The results show that, in most cases, combining structural and textual attacks increases overall attack effectiveness. Interestingly, TAPE sometimes benefits from additional textual perturbations, demonstrating its strong robustness. Among the attack combinations, sentence-level attacks with structural perturbations are generally the most effective. These findings provide preliminary evidence that joint structural-textual attacks are a promising direction for enhancing attack effectiveness.
Beyond simple combinations, adversaries could adopt a coordinated strategy that greedily selects structural or textual perturbations to maximize the loss of a surrogate model, enabling adaptive selection of the most impactful perturbations.

\textbf{Defensive Perspectives.} 
(1) \textit{Robust model architectures:} Future work should explore architectures that naturally exhibit robustness. Our findings suggest that fine-tuned LMs and GraphSAGE exhibit greater resistance to attacks compared with LLMs and other GNN backbones, highlighting the potential of domain-specific fine-tuning and preserving self-representations.

(2) \textit{Graph purification:} 
Section~\ref{analysis_structural} shows that structural attacks influence the structural properties (e.g., neighbor consistency) of original graphs. Therefore, one can use pre-processing strategies based on neighbor consistency to filter suspicious edges. However, existing methods typically rely on shallow embeddings for such filtration, which provide limited defense performance. We propose a graph purification defense using LLM/LM-generated embeddings, which encode more label semantics and structural information than the shallow embeddings. Specifically, we compute the cosine similarity between the embeddings of each edge's endpoints and remove edges with similarity below a predefined threshold. To validate the effectiveness of this defense, we compare it with an input-level defense~\cite{jin2021node} and a model-level defense~\cite{zhu2019robust}.
Table~\ref{tab:cleangraph} presents the results using GCN under non-targeted structural attacks. We find that our defense method is highly effective: models with purification achieve significantly higher accuracy on Pubmed and Tape-arxiv23 than those without defense and, in some cases, even outperform models trained on clean graphs. In contrast, the baseline defenses show inconsistent effectiveness across different datasets, suggesting limited generalizability.
In summary, graph purification using LLM/LM-generated embeddings provides a simple yet effective defense with good generality. However, its effectiveness depends on threshold selection, which correlates with structural properties, e.g., higher-homophily datasets typically require lower thresholds.

(3) \textit{Perturbations as data augmentation:} Section~\ref{evaluation_textual} shows that introducing textual perturbations sometimes improves model performance, suggesting that controlled adversarial perturbations may serve as effective data augmentation. Future work could explore incorporating such perturbations into model training to enhance generalization and robustness against unseen attacks.

\section{Conclusion}
In this paper, we proposed a robustness assessment framework for evaluating LLM-enhanced GNNs against poisoning attacks. We systematically evaluated 24 LLM-enhanced GNNs under six structural and three textual poisoning attacks across four real-world TAG datasets. Our evaluation revealed that LLM-enhanced GNNs consistently outperform a shallow embedding-based baseline and exhibit strong robustness across diverse attack settings. We further explained our assessment results through embedding visualization and analyses of key embedding properties. Finally, we discussed several promising research directions from both offensive and defensive perspectives.

\section*{Acknowledgment}
This work is supported in part by the ``Pioneer'' and ``Leading Goose'' R\&D Program of Zhejiang under Grant No. 2026C01020; in part by the National Natural Science Foundation of China under Grant U23A20300; in part by the Concept Verification Funding of Hangzhou Institute of Technology of Xidian University under Grant GNYZ2024XX007; and in part by the 111 Center under Grant B16037.

\section*{Ethics Considerations}
Our research on the robustness of LLM-enhanced GNNs is driven by the primary goal of improving their security. We have carefully considered the ethical implications of this work to ensure responsible conduct.

\textbf{Stakeholders and Potential Impact.}
The stakeholders of this research include:
(1) Developers: Our robustness assessment framework is a valuable tool for developers to proactively identify vulnerabilities and design secure models. However, these vulnerabilities could be exploited by malicious actors.
(2) End users: Users of applications built on these models benefit from secure and reliable AI services. However, if the vulnerabilities we identified in this work are exploited before adequate defenses are deployed, end users may face risks such as incorrect predictions in critical applications.
(3) AI security research community: Our framework provides a standardized approach to assess the robustness of LLM-enhanced GNN models against poisoning attacks, which promotes reproducible research and accelerates the development of defense methods.
(4) Society at large: By strengthening the security of LLM-enhanced GNNs, our research can foster public trust in LLM-driven GNN applications.

\textbf{Mitigation.}
We acknowledge that publishing attack strategies carries risks. To mitigate these, we have taken the following key measures:
(1) Coupled attacks and defenses: We present potential new poisoning attacks only for research purposes, and they are discussed with corresponding defense mechanisms (see Section~\ref{future_direction}). Our objective is to inform the community about risks so that defenses can be developed, not to provide tools for malicious exploitation.
(2) Use of public data: All datasets used in this work are publicly available academic benchmarks. We have not used any personally identifiable information or sensitive data.
(3) Community engagement: We commit to actively monitoring community feedback and adjusting our disclosure strategies as needed to minimize the potential for misuse.

By following these principles, we aim to ensure that our research not only identifies vulnerabilities but also empowers the community to build strong defenses, supporting the safe and trustworthy use of LLM-enhanced GNNs.

\section*{LLM Usage Considerations}
(1) \textbf{Originality:} LLMs were used for editorial purposes in this manuscript, and all outputs were inspected by the authors to ensure accuracy and originality.
(2) \textbf{Transparency:} LLMs were used as feature enhancers for GNNs in this work, generating node embeddings from node textual attributes. All LLMs are open-sourced models or can be accessed through official APIs, as introduced in Section~\ref{victim_training}. Our research aims to assess the robustness of LLM-enhanced GNNs; thus, the core contribution is our robustness assessment framework. LLMs do not contribute to the design of attack strategies, evaluation framework, or result analysis; instead, they serve as components within the victim models being studied.
(3) \textbf{Responsibility:} All datasets are public academic benchmarks without sensitive information. The LLMs used are widely adopted in prior work, with the largest containing about 7 billion parameters. They were used solely for inference (e.g., embedding generation), minimizing computational cost and ensuring accessibility on standard academic GPUs. For example, generating embeddings for the Pubmed dataset with the LLaMA-2-7b-hf model on an RTX 4070 Ti Super took about one hour.






%
\bibliographystyle{IEEEtran}
\bibliography{IEEEabrv,reference}

\begin{thebibliography}{10}
\providecommand{\url}[1]{#1}
\csname url@samestyle\endcsname
\providecommand{\newblock}{\relax}
\providecommand{\bibinfo}[2]{#2}
\providecommand{\BIBentrySTDinterwordspacing}{\spaceskip=0pt\relax}
\providecommand{\BIBentryALTinterwordstretchfactor}{4}
\providecommand{\BIBentryALTinterwordspacing}{\spaceskip=\fontdimen2\font plus
\BIBentryALTinterwordstretchfactor\fontdimen3\font minus \fontdimen4\font\relax}
\providecommand{\BIBforeignlanguage}[2]{{%
\expandafter\ifx\csname l@#1\endcsname\relax
\typeout{** WARNING: IEEEtran.bst: No hyphenation pattern has been}%
\typeout{** loaded for the language `#1'. Using the pattern for}%
\typeout{** the default language instead.}%
\else
\language=\csname l@#1\endcsname
\fi
#2}}
\providecommand{\BIBdecl}{\relax}
\BIBdecl

\bibitem{zhang2020gnnguard}
X.~Zhang and M.~Zitnik, ``Gnnguard: Defending graph neural networks against adversarial attacks,'' \emph{Adv. Neural Inf. Process. Syst.}, vol.~33, pp. 9263--9275, 2020.

\bibitem{kipf2017semi}
T.~N. Kipf and M.~Welling, ``Semi-supervised classification with graph convolutional networks,'' in \emph{Proc. Int. Conf. Learn. Representations}, 2017.

\bibitem{zhang2018link}
M.~Zhang and Y.~Chen, ``Link prediction based on graph neural networks,'' \emph{Adv. Neural Inf. Process. Syst.}, vol.~31, 2018.

\bibitem{xu2018powerful}
K.~Xu, W.~Hu, J.~Leskovec, and S.~Jegelka, ``How powerful are graph neural networks?'' \emph{arXiv preprint arXiv:1810.00826}, 2018.

\bibitem{yao2019graph}
L.~Yao, C.~Mao, and Y.~Luo, ``Graph convolutional networks for text classification,'' in \emph{Proc. AAAI Conf. Artif. Intell.}, vol.~33, no.~01, 2019, pp. 7370--7377.

\bibitem{harris1954distributional}
Z.~S. Harris, ``Distributional structure,'' \emph{Word}, vol.~10, no. 2-3, pp. 146--162, 1954.

\bibitem{mikolov2013distributed}
T.~Mikolov, I.~Sutskever, K.~Chen, G.~S. Corrado, and J.~Dean, ``Distributed representations of words and phrases and their compositionality,'' \emph{Adv. Neural Inf. Process. Syst.}, vol.~26, 2013.

\bibitem{wu2024legilimens}
J.~Wu, J.~Deng, S.~Pang, Y.~Chen, J.~Xu, X.~Li, and W.~Xu, ``Legilimens: Practical and unified content moderation for large language model services,'' in \emph{Proc. ACM SIGSAC Conf. Comput. Commun. Secur.}, 2024, pp. 1151--1165.

\bibitem{he2024harnessing}
X.~He, X.~Bresson, T.~Laurent, A.~Perold, Y.~LeCun, and B.~Hooi, ``Harnessing explanations: Llm-to-lm interpreter for enhanced text-attributed graph representation learning,'' in \emph{Proc. Int. Conf. Learn. Representations}, 2024.

\bibitem{chen2024exploring}
Z.~Chen, H.~Mao, H.~Li, W.~Jin, H.~Wen, X.~Wei, S.~Wang, D.~Yin, W.~Fan, H.~Liu \emph{et~al.}, ``Exploring the potential of large language models (llms) in learning on graphs,'' \emph{ACM SIGKDD Explorations Newslett.}, vol.~25, no.~2, pp. 42--61, 2024.

\bibitem{liu2023one}
H.~Liu, J.~Feng, L.~Kong, N.~Liang, D.~Tao, Y.~Chen, and M.~Zhang, ``One for all: Towards training one graph model for all classification tasks,'' \emph{arXiv preprint arXiv:2310.00149}, 2023.

\bibitem{zhu2024efficient}
Y.~Zhu, Y.~Wang, H.~Shi, and S.~Tang, ``Efficient tuning and inference for large language models on textual graphs,'' in \emph{Proc. Int. Joint Conf. Artif. Intell.}, 2024, pp. 5734--5742.

\bibitem{chien2021node}
E.~Chien, W.-C. Chang, C.-J. Hsieh, H.-F. Yu, J.~Zhang, O.~Milenkovic, and I.~S. Dhillon, ``Node feature extraction by self-supervised multi-scale neighborhood prediction,'' \emph{arXiv preprint arXiv:2111.00064}, 2021.

\bibitem{duan2023simteg}
K.~Duan, Q.~Liu, T.-S. Chua, S.~Yan, W.~T. Ooi, Q.~Xie, and J.~He, ``Simteg: A frustratingly simple approach improves textual graph learning,'' \emph{arXiv preprint arXiv:2308.02565}, 2023.

\bibitem{li2024glbench}
Y.~Li, P.~Wang, X.~Zhu, A.~Chen, H.~Jiang, D.~Cai, V.~W. Chan, and J.~Li, ``Glbench: A comprehensive benchmark for graph with large language models,'' \emph{Adv. Neural Inf. Process. Syst.}, vol.~37, pp. 42\,349--42\,368, 2024.

\bibitem{kumar2020adversarial}
R.~S.~S. Kumar, M.~Nystr{\"o}m, J.~Lambert, A.~Marshall, M.~Goertzel, A.~Comissoneru, M.~Swann, and S.~Xia, ``Adversarial machine learning-industry perspectives,'' in \emph{Proc. IEEE Secur. Privacy Workshops}, 2020, pp. 69--75.

\bibitem{tang2020transferring}
X.~Tang, Y.~Li, Y.~Sun, H.~Yao, P.~Mitra, and S.~Wang, ``Transferring robustness for graph neural network against poisoning attacks,'' in \emph{Proc. Int. Conf. Web Search Data Mining}, 2020, pp. 600--608.

\bibitem{guo2024learning}
K.~Guo, Z.~Liu, Z.~Chen, H.~Wen, W.~Jin, J.~Tang, and Y.~Chang, ``Learning on graphs with large language models (llms): A deep dive into model robustness,'' \emph{arXiv preprint arXiv:2407.12068}, 2024.

\bibitem{zhang2024can}
Z.~Zhang, X.~Wang, H.~Zhou, Y.~Yu, M.~Zhang, C.~Yang, and C.~Shi, ``Can large language models improve the adversarial robustness of graph neural networks?'' in \emph{Proc. ACM SIGKDD Conf. Knowl. Discovery Data Mining}, 2025, pp. 2008--2019.

\bibitem{mccallum2000automating}
A.~K. McCallum, K.~Nigam, J.~Rennie, and K.~Seymore, ``Automating the construction of internet portals with machine learning,'' \emph{Inform. Retrieval}, vol.~3, pp. 127--163, 2000.

\bibitem{sen2008collective}
P.~Sen, G.~Namata, M.~Bilgic, L.~Getoor, B.~Galligher, and T.~Eliassi-Rad, ``Collective classification in network data,'' \emph{AI Magazine}, vol.~29, no.~3, pp. 93--93, 2008.

\bibitem{touvron2023llama}
H.~Touvron, T.~Lavril, G.~Izacard, X.~Martinet, M.-A. Lachaux, T.~Lacroix, B.~Rozi{\`e}re, N.~Goyal, E.~Hambro, F.~Azhar \emph{et~al.}, ``Llama: Open and efficient foundation language models,'' \emph{arXiv preprint arXiv:2302.13971}, 2023.

\bibitem{openai-embeddings-api}
\BIBentryALTinterwordspacing
OpenAI. (2024) Embeddings {API} reference. Accessed: Oct. 11, 2025. [Online]. Available: \url{https://platform.openai.com/docs/api-reference/embeddings}
\BIBentrySTDinterwordspacing

\bibitem{LinqAIResearch2024}
\BIBentryALTinterwordspacing
J.~Kim, S.~Lee, J.~Kwon, S.~Gu, Y.~Kim, M.~Cho, J.~yong Sohn, and C.~Choi. (2024) {Linq-Embed-Mistral}: Elevating text retrieval with improved {GPT} data through task-specific control and quality refinement. Accessed: Oct. 11, 2025. [Online]. Available: \url{https://getlinq.com/blog/linq-embed-mistral/}
\BIBentrySTDinterwordspacing

\bibitem{wang2022text}
L.~Wang, N.~Yang, X.~Huang, B.~Jiao, L.~Yang, D.~Jiang, R.~Majumder, and F.~Wei, ``Text embeddings by weakly-supervised contrastive pre-training,'' \emph{arXiv preprint arXiv:2212.03533}, 2022.

\bibitem{warner2024smarter}
B.~Warner, A.~Chaffin, B.~Clavi{\'e}, O.~Weller, O.~Hallstr{\"o}m, S.~Taghadouini, A.~Gallagher, R.~Biswas, F.~Ladhak, T.~Aarsen \emph{et~al.}, ``Smarter, better, faster, longer: A modern bidirectional encoder for fast, memory efficient, and long context finetuning and inference,'' \emph{arXiv preprint arXiv:2412.13663}, 2024.

\bibitem{velivckovic2017graph}
P.~Veli{\v{c}}kovi{\'c}, G.~Cucurull, A.~Casanova, A.~Romero, P.~Lio, and Y.~Bengio, ``Graph attention networks,'' \emph{arXiv preprint arXiv:1710.10903}, 2017.

\bibitem{hamilton2017inductive}
W.~Hamilton, Z.~Ying, and J.~Leskovec, ``Inductive representation learning on large graphs,'' \emph{Adv. Neural Inf. Process. Syst.}, vol.~30, 2017.

\bibitem{zugner2018adversarial}
D.~Z{\"u}gner, A.~Akbarnejad, and S.~G{\"u}nnemann, ``Adversarial attacks on neural networks for graph data,'' in \emph{Proc. ACM SIGKDD Int. Conf. Knowl. Discovery Data Mining}, 2018, pp. 2847--2856.

\bibitem{li2021adversarial}
J.~Li, T.~Xie, L.~Chen, F.~Xie, X.~He, and Z.~Zheng, ``Adversarial attack on large scale graph,'' \emph{IEEE Trans. Knowl. Data Eng.}, vol.~35, no.~1, pp. 82--95, 2021.

\bibitem{zhao2023black}
S.~Zhao, W.~Wang, Z.~Du, J.~Chen, and Z.~Duan, ``A black-box adversarial attack method via nesterov accelerated gradient and rewiring towards attacking graph neural networks,'' \emph{IEEE Trans. Big Data}, 2023.

\bibitem{zugner_adversarial_2019}
D.~Z{\"u}gner and S.~G{\"u}nnemann, ``Adversarial attacks on graph neural networks via meta learning,'' in \emph{Proc. Int. Conf. Learn. Representations}, 2019.

\bibitem{waniek2018hiding}
M.~Waniek, T.~P. Michalak, M.~J. Wooldridge, and T.~Rahwan, ``Hiding individuals and communities in a social network,'' \emph{Nature Human Behaviour}, vol.~2, no.~2, pp. 139--147, 2018.

\bibitem{zhu2024simple}
G.~Zhu, M.~Chen, C.~Yuan, and Y.~Huang, ``Simple and efficient partial graph adversarial attack: A new perspective,'' \emph{IEEE Trans. Knowl. Data Eng.}, vol.~36, no.~8, pp. 4245--4259, 2024.

\bibitem{gao2018black}
J.~Gao, J.~Lanchantin, M.~L. Soffa, and Y.~Qi, ``Black-box generation of adversarial text sequences to evade deep learning classifiers,'' in \emph{Proc. IEEE Secur. Privacy Workshops}, 2018, pp. 50--56.

\bibitem{li2020bert}
L.~Li, R.~Ma, Q.~Guo, X.~Xue, and X.~Qiu, ``Bert-attack: Adversarial attack against bert using bert,'' in \emph{Proc. Conf. Empirical Methods Natural Lang. Process.}, 2020, pp. 6193--6202.

\bibitem{chen2021multi}
Y.~Chen, J.~Su, and W.~Wei, ``Multi-granularity textual adversarial attack with behavior cloning,'' in \emph{Proc. Conf. Empirical Methods Natural Lang. Process.}, 2021, pp. 4511--4526.

\bibitem{sun2022adversarial}
L.~Sun, Y.~Dou, C.~Yang, K.~Zhang, J.~Wang, P.~S. Yu, L.~He, and B.~Li, ``Adversarial attack and defense on graph data: A survey,'' \emph{IEEE Trans. Knowl. Data Eng.}, vol.~35, no.~8, pp. 7693--7711, 2022.

\bibitem{zhou2020graph}
J.~Zhou, G.~Cui, S.~Hu, Z.~Zhang, C.~Yang, Z.~Liu, L.~Wang, C.~Li, and M.~Sun, ``Graph neural networks: A review of methods and applications,'' \emph{AI Open}, vol.~1, pp. 57--81, 2020.

\bibitem{salton1988term}
G.~Salton and C.~Buckley, ``Term-weighting approaches in automatic text retrieval,'' \emph{Inf. Process. Manage.}, vol.~24, no.~5, pp. 513--523, 1988.

\bibitem{jin2020graph}
W.~Jin, Y.~Ma, X.~Liu, X.~Tang, S.~Wang, and J.~Tang, ``Graph structure learning for robust graph neural networks,'' in \emph{Proc. ACM SIGKDD Int. Conf. Knowl. Discovery Data Mining}, 2020, pp. 66--74.

\bibitem{geisler2021robustness}
S.~Geisler, T.~Schmidt, H.~{\c{S}}irin, D.~Z{\"u}gner, A.~Bojchevski, and S.~G{\"u}nnemann, ``Robustness of graph neural networks at scale,'' \emph{Adv. Neural Inf. Process. Syst.}, vol.~34, pp. 7637--7649, 2021.

\bibitem{li2024survey}
Y.~Li, Z.~Li, P.~Wang, J.~Li, X.~Sun, H.~Cheng, and J.~X. Yu, ``A survey of graph meets large language model: progress and future directions,'' in \emph{Proc. Int. Joint Conf. Artif. Intell.}, 2024, pp. 8123--8131.

\bibitem{achiam2023gpt}
J.~Achiam, S.~Adler, S.~Agarwal, L.~Ahmad, I.~Akkaya, F.~L. Aleman, D.~Almeida, J.~Altenschmidt, S.~Altman, S.~Anadkat \emph{et~al.}, ``Gpt-4 technical report,'' \emph{arXiv preprint arXiv:2303.08774}, 2023.

\bibitem{koroteev2021bert}
M.~V. Koroteev, ``Bert: a review of applications in natural language processing and understanding,'' \emph{arXiv preprint arXiv:2103.11943}, 2021.

\bibitem{liu2019roberta}
Y.~Liu, M.~Ott, N.~Goyal, J.~Du, M.~Joshi, D.~Chen, O.~Levy, M.~Lewis, L.~Zettlemoyer, and V.~Stoyanov, ``Roberta: A robustly optimized bert pretraining approach,'' \emph{arXiv preprint arXiv:1907.11692}, 2019.

\bibitem{brown2020language}
T.~Brown, B.~Mann, N.~Ryder, M.~Subbiah, J.~D. Kaplan, P.~Dhariwal, A.~Neelakantan, P.~Shyam, G.~Sastry, A.~Askell \emph{et~al.}, ``Language models are few-shot learners,'' \emph{Adv. Neural Inf. Process. Syst.}, vol.~33, pp. 1877--1901, 2020.

\bibitem{he2020deberta}
P.~He, X.~Liu, J.~Gao, and W.~Chen, ``Deberta: Decoding-enhanced bert with disentangled attention,'' \emph{arXiv preprint arXiv:2006.03654}, 2020.

\bibitem{devlin2019bert}
J.~Devlin, M.-W. Chang, K.~Lee, and K.~Toutanova, ``Bert: Pre-training of deep bidirectional transformers for language understanding,'' in \emph{Proc. Conf. North Amer. Assoc. Comput. Linguist. Human Lang. Technol.}, 2019, pp. 4171--4186.

\bibitem{xu2018how}
K.~Xu, W.~Hu, J.~Leskovec, and S.~Jegelka, ``How powerful are graph neural networks?'' in \emph{Proc. Int. Conf. Learn. Representations}, 2019.

\bibitem{brody2022attentive}
S.~Brody, U.~Alon, and E.~Yahav, ``How attentive are graph attention networks?'' in \emph{Proc. Int. Conf. Learn. Representations}, 2022.

\bibitem{maaten2008visualizing}
L.~v.~d. Maaten and G.~Hinton, ``Visualizing data using t-sne,'' \emph{J. Mach. Learn. Res.}, vol.~9, no. Nov, pp. 2579--2605, 2008.

\bibitem{davies2009cluster}
D.~L. Davies and D.~W. Bouldin, ``A cluster separation measure,'' \emph{IEEE Trans. Pattern Anal. Mach. Intell.}, no.~2, pp. 224--227, 2009.

\bibitem{rousseeuw1987silhouettes}
P.~J. Rousseeuw, ``Silhouettes: a graphical aid to the interpretation and validation of cluster analysis,'' \emph{J. Comput. Appl. Math.}, vol.~20, pp. 53--65, 1987.

\bibitem{hjelm2018learning}
R.~D. Hjelm, A.~Fedorov, S.~Lavoie-Marchildon, K.~Grewal, P.~Bachman, A.~Trischler, and Y.~Bengio, ``Learning deep representations by mutual information estimation and maximization,'' \emph{arXiv preprint arXiv:1808.06670}, 2018.

\bibitem{zhao2023deep}
W.~Zhao, G.~Xu, Z.~Cui, S.~Luo, C.~Long, and T.~Zhang, ``Deep graph structural infomax,'' in \emph{Proc. AAAI Conf. Artif. Intell.}, vol.~37, no.~4, 2023, pp. 4920--4928.

\bibitem{wang2020unifying}
H.~Wang and J.~Leskovec, ``Unifying graph convolutional neural networks and label propagation,'' \emph{arXiv preprint arXiv:2002.06755}, 2020.

\bibitem{rong2014word2vec}
X.~Rong, ``word2vec parameter learning explained,'' \emph{arXiv preprint arXiv:1411.2738}, 2014.

\bibitem{wolf2020transformers}
T.~Wolf, L.~Debut, V.~Sanh, J.~Chaumond, C.~Delangue, A.~Moi, P.~Cistac, T.~Rault, R.~Louf, M.~Funtowicz \emph{et~al.}, ``Transformers: State-of-the-art natural language processing,'' in \emph{Proc. Conf. Empirical Methods Natural Lang. Process. Syst. Demonstrations}, 2020, pp. 38--45.

\bibitem{li2020deeprobust}
Y.~Li, W.~Jin, H.~Xu, and J.~Tang, ``Deeprobust: A pytorch library for adversarial attacks and defenses,'' \emph{arXiv preprint arXiv:2005.06149}, 2020.

\bibitem{zeng2020openattack}
G.~Zeng, F.~Qi, Q.~Zhou, T.~Zhang, Z.~Ma, B.~Hou, Y.~Zang, Z.~Liu, and M.~Sun, ``Openattack: An open-source textual adversarial attack toolkit,'' \emph{arXiv preprint arXiv:2009.09191}, 2020.

\bibitem{zhu2019robust}
D.~Zhu, Z.~Zhang, P.~Cui, and W.~Zhu, ``Robust graph convolutional networks against adversarial attacks,'' in \emph{Proc. ACM SIGKDD Int. Conf. Knowl. Discovery Data Mining}, 2019, pp. 1399--1407.

\bibitem{jin2021node}
W.~Jin, T.~Derr, Y.~Wang, Y.~Ma, Z.~Liu, and J.~Tang, ``Node similarity preserving graph convolutional networks,'' in \emph{Proc. ACM Int. Conf. Web Search Data Mining}, 2021, pp. 148--156.

\bibitem{bojchevski2020efficient}
A.~Bojchevski, J.~Gasteiger, and S.~G{\"u}nnemann, ``Efficient robustness certificates for discrete data: sparsity-aware randomized smoothing for graphs, images and more,'' in \emph{Proc. Int. Conf. Mach. Learn.}, 2020, pp. 1003--1013.

\end{thebibliography}

\begin{table*}[!t]
\renewcommand\arraystretch{1.2}
\caption{Accuracy (\%) on clean graphs for targeted nodes.}
\label{tab:targeted-clean}
\centering
\resizebox{0.8\textwidth}{!}{
 \renewcommand{\arraystretch}{0.8}
\begin{tabular}{lccccccccccccc}
\toprule
\multirow{2}{*}{\textbf{Classification}} & \multirow{2}{*}{\textbf{Model}} & \multicolumn{3}{c}{\textbf{Cora}} & \multicolumn{3}{c}{\textbf{Pubmed}} & \multicolumn{3}{c}{\textbf{Ogbn-products}} & \multicolumn{3}{c}{\textbf{Tape-arxiv23}}  \\
\cmidrule(lr){3-5} \cmidrule(lr){6-8} \cmidrule(lr){9-11} \cmidrule(lr){12-14} 
& & GCN & GAT & SAGE  & GCN & GAT & SAGE  & GCN & GAT & SAGE  & GCN & GAT & SAGE \\
\midrule
& Shallow emb. & 85.46 & 81.06 & 82.07 & 89.49 & 89.38 & 87.76 & 84.67 & 83.91 & 79.52 & 68.49 & 64.26 & 69.70 \\

\midrule
\multirow{2}{*}{LLM-Explanation} & TAPE & 83.11 & 82.07 & 82.59 & 90.34 & \colorbox[HTML]{D4FAFC}{\underline{91.52}} & 90.88 & 88.06 & 88.07 & 88.12 & \colorbox[HTML]{D4FAFC}{\underline{76.91}} & \colorbox[HTML]{D4FAFC}{\underline{76.27}} & \colorbox[HTML]{B8E7E1}{\textbf{80.67}} \\

& KEA & 85.10 & \colorbox[HTML]{D4FAFC}{\underline{85.33}} & \colorbox[HTML]{D4FAFC}{\underline{86.13}} & 90.34 & 89.38 & 90.77 & 88.21 & 87.63 & 89.17 & 73.65 & 71.54 & 76.93 \\

\midrule
\multirow{3}{*}{LLM-Embedding}  & LLaMA &86.77 & \colorbox[HTML]{B8E7E1}{\textbf{85.48}} & 81.08 & 90.56 & 87.98 & 89.70 & 88.21 & 84.77 & 88.68 & 76.07 & 73.93 & 78.39 \\

&  TE3L &\colorbox[HTML]{B8E7E1}{\textbf{89.64}} & 82.66 & \colorbox[HTML]{B8E7E1}{\textbf{86.78}} & 90.02 & 90.13 & 91.20 & \colorbox[HTML]{D4FAFC}{\underline{88.72}} & \colorbox[HTML]{D4FAFC}{\underline{88.45}} & \colorbox[HTML]{B8E7E1}{\textbf{90.06}} & 74.42 & \colorbox[HTML]{B8E7E1}{\textbf{76.62}} & 75.85 \\

& Linq & \colorbox[HTML]{D4FAFC}{\underline{87.81}} & 83.23 & 84.92 & 90.56 & 89.05 & 90.88 & 87.92 & 87.05 & \colorbox[HTML]{D4FAFC}{\underline{89.68}} & 76.55 & 71.51 & \colorbox[HTML]{D4FAFC}{\underline{78.66}} \\

\midrule
\multirow{3}{*}{LM-Embedding}  & SimTeg &86.81 & 83.90 & 85.47 & \colorbox[HTML]{B8E7E1}{\textbf{91.20}} & 89.48 & 90.56 & 87.92 & 86.32 & 89.26 & 74.24 & 74.79 & 76.25 \\

& E5-Large & 87.29 & 82.10 & 80.00 & \colorbox[HTML]{D4FAFC}{\underline{91.09}} & \colorbox[HTML]{B8E7E1}{\textbf{91.85}} & \colorbox[HTML]{B8E7E1}{\textbf{92.38}} & \colorbox[HTML]{B8E7E1}{\textbf{88.78}} & \colorbox[HTML]{B8E7E1}{\textbf{88.63}} & 88.73 & \colorbox[HTML]{B8E7E1}{\textbf{77.77}} & 75.88 & 77.51 \\

&  ModernBert & 79.47 & 78.69 & 73.78 & 90.23 & 87.75 & \colorbox[HTML]{D4FAFC}{\underline{92.17}} & 88.21 & 84.02 & 88.65 & 73.01 & 74.51 & 77.29 \\

\bottomrule
\end{tabular}
}
\end{table*}

\begin{table*}[!t]
\renewcommand\arraystretch{1.2}
\caption{Accuracy (\%) on clean graphs for non-targeted nodes.}
\label{tab:global-clean}
\centering
\resizebox{0.8\textwidth}{!}{
 \renewcommand{\arraystretch}{0.8}
\begin{tabular}{lccccccccccccc}
\toprule
\multirow{2}{*}{\textbf{Classification}} & \multirow{2}{*}{\textbf{Model}} & \multicolumn{3}{c}{\textbf{Cora}} & \multicolumn{3}{c}{\textbf{Pubmed}} & \multicolumn{3}{c}{\textbf{Ogbn-products}} & \multicolumn{3}{c}{\textbf{Tape-arxiv23}}  \\
\cmidrule(lr){3-5} \cmidrule(lr){6-8} \cmidrule(lr){9-11} \cmidrule(lr){12-14} 
& & GCN & GAT & SAGE & GCN & GAT & SAGE & GCN & GAT & SAGE & GCN & GAT & SAGE \\
\midrule
& Shallow emb. &81.02 & 79.99 & 80.03 & 86.21 & 85.49 & 85.37 & 65.95 & 69.39 & 62.29 & 57.72 & 55.47 & 56.32 \\

\midrule
\multirow{2}{*}{LLM-Explanation} & TAPE & 81.74 & 80.48 & 81.82 & \colorbox[HTML]{B8E7E1}{\textbf{93.12}} & 89.77 & \colorbox[HTML]{B8E7E1}{\textbf{94.54}} & \colorbox[HTML]{B8E7E1}{\textbf{85.85}} & \colorbox[HTML]{B8E7E1}{\textbf{85.36}} & \colorbox[HTML]{B8E7E1}{\textbf{86.05}} & \colorbox[HTML]{B8E7E1}{\textbf{74.47}} & \colorbox[HTML]{B8E7E1}{\textbf{74.44}} & \colorbox[HTML]{B8E7E1}{\textbf{74.34}} \\
& KEA & 84.20 & \colorbox[HTML]{B8E7E1}{\textbf{84.47}} & \colorbox[HTML]{B8E7E1}{\textbf{84.85}} & 88.97 & 86.55 & 89.31 & 83.89 & 83.16 & 84.27 & 68.85 & 69.87 & 69.28 \\
\midrule
\multirow{3}{*}{LLM-Embedding} & LLaMA & 81.64 & 81.60 & 81.23 & 87.87 & 84.97 & 87.98 & 83.70 & 80.42 & 83.43 & 71.51 & 70.44 & 71.10 \\
& TE3L & 83.64 & \colorbox[HTML]{D4FAFC}{\underline{83.00}} & 83.89 & 88.76 & 87.09 & 89.04 & 84.08 & 82.83 & 84.55 & 70.17 & 68.80 & 69.86 \\
& Linq & \colorbox[HTML]{D4FAFC}{\underline{84.29}} & 82.57 & \colorbox[HTML]{D4FAFC}{\underline{84.42}} & 89.69 & 87.15 & 91.30 & 84.25 & 82.23 & 84.74 & 72.32 & 71.52 & 72.53 \\
\midrule
\multirow{3}{*}{LM-Embedding} & SimTeg & 83.29 & 82.89 & 83.77 & 89.28 & \colorbox[HTML]{D4FAFC}{\underline{93.06}} & 92.58 & 83.31 & 82.89 & 83.58 & 72.01 & 70.97 & 71.66 \\
& E5-Large & \colorbox[HTML]{B8E7E1}{\textbf{84.41}} & 82.00 & 83.79 & \colorbox[HTML]{D4FAFC}{\underline{93.06}} & 89.85 & \colorbox[HTML]{D4FAFC}{\underline{94.14}} &\colorbox[HTML]{D4FAFC}{\underline{84.96}} & \colorbox[HTML]{D4FAFC}{\underline{84.04}} & \colorbox[HTML]{D4FAFC}{\underline{84.99}} & \colorbox[HTML]{D4FAFC}{\underline{73.08}} & \colorbox[HTML]{D4FAFC}{\underline{71.67}} & \colorbox[HTML]{D4FAFC}{\underline{72.55}} \\
& ModernBert & 77.71 & 75.39 & 74.78 & 90.36 & \colorbox[HTML]{B8E7E1}{\textbf{94.32}} & 93.86 & 83.65 & 80.85 & 83.90 & 70.33 & 68.49 & 70.47 \\
\bottomrule
\end{tabular}
}
\end{table*}

\appendices

\section{Formulas of Embedding Properties} \label{appendix_formula}
\textbf{Embedding Homophily (Hom).} 
It is defined as:
\begin{equation}
    \text{Hom} = \frac{1}{N \cdot k} \sum_{i=1}^{N} \sum_{j \in \mathcal{N}_k(i)} \mathbf{1}[\,y_j = y_i\,],
\end{equation}
where $N$ is the number of nodes, $\mathcal{N}_k(i)$ denotes the set of $k$-nearest neighbors of node $i$ in the embedding space, $y_i$ is the label of node $i$, and $\mathbf{1}[\cdot]$ is an indicator function.

\textbf{Embedding-Label Mutual Information (ELMI).} 
Let $X$ and $Y$ be the random variables representing node embeddings and node labels, respectively. The mutual information between them is defined as:
\begin{equation}
    \text{ELMI} = \sum_{x \in X}\sum_{y \in Y} p(x,y)\,\log\frac{p(x,y)}{p(x)\,p(y)},
\end{equation}
where $p(x,y)$ is the joint distribution of embeddings and labels, and $p(x)$ and $p(y)$ are the corresponding marginal distributions.

\textbf{Embedding-Structural Mutual Information (ESMI).} 
Let $X$ and $H$ be the random variables representing node embeddings and local structural properties (e.g., degree and clustering coefficient), respectively. The mutual information between them is defined as:
\begin{equation}
    \text{ESMI} = \sum_{x \in X}\sum_{h \in H} p(x,h)\,\log\frac{p(x,h)}{p(x)\,p(h)},
\end{equation}
where $p(x,h)$ is the joint distribution of embeddings and local structural properties, and $p(x)$ and $p(h)$ are the corresponding marginal distributions.

\textbf{Neighbor Consistency (NCon).}
It is defined as:
\begin{equation}
    \text{NCon} = \frac{\sum_{i,j} \mathbf{A}_{ij}\,\cos{(x_{i},x_{j}})}{\sum_i \sum_{j} \mathbf{A}_{ij}},
\end{equation}
where $\mathbf{A}$ denotes an adjacency matrix, $\cos(\cdot)$ is the cosine similarity, and $x_i$ denotes the embedding of node $i$.

\begin{table*}[!t]
\renewcommand\arraystretch{1.2}
\caption{Robustness comparison against the \textbf{SGA} attack at a high perturbation level.}
\label{tab:high-sga}
\resizebox{1.0\textwidth}{!}{
\begin{tabular}{lcccccccccccccccccccccccc}
\toprule
\multirow{4}{*}{\textbf{Model}} & \multicolumn{6}{c}{\textbf{Cora}} & \multicolumn{6}{c}{\textbf{Pubmed}} & \multicolumn{6}{c}{\textbf{Ogbn-products}} & \multicolumn{6}{c}{\textbf{Tape-arxiv23}}  \\
\cmidrule(lr){2-7} \cmidrule(lr){8-13} \cmidrule(lr){14-19} \cmidrule(lr){20-25} 
& \multicolumn{2}{c}{GCN} & \multicolumn{2}{c}{GAT} &  \multicolumn{2}{c}{SAGE} & \multicolumn{2}{c}{GCN} & \multicolumn{2}{c}{GAT} &  \multicolumn{2}{c}{SAGE} & \multicolumn{2}{c}{GCN} & \multicolumn{2}{c}{GAT} &  \multicolumn{2}{c}{SAGE} & \multicolumn{2}{c}{GCN} & \multicolumn{2}{c}{GAT} &  \multicolumn{2}{c}{SAGE} \\
\cmidrule(lr){2-3} \cmidrule(lr){4-5} \cmidrule(lr){6-7} \cmidrule(lr){8-9} \cmidrule(lr){10-11} \cmidrule(lr){12-13} \cmidrule(lr){14-15} \cmidrule(lr){16-17} \cmidrule(lr){18-19}\cmidrule(lr){20-21}\cmidrule(lr){22-23}\cmidrule(lr){24-25}
& ACC $\uparrow$ & RDA $\downarrow$  & ACC $\uparrow$ & RDA $\downarrow$  & ACC $\uparrow$ & RDA $\downarrow$  & ACC $\uparrow$ & RDA $\downarrow$ & ACC $\uparrow$ & RDA $\downarrow$ & ACC $\uparrow$ & RDA $\downarrow$ & ACC $\uparrow$ & RDA $\downarrow$ & ACC $\uparrow$ & RDA $\downarrow$ & ACC $\uparrow$ & RDA $\downarrow$ & ACC $\uparrow$ & RDA $\downarrow$ & ACC $\uparrow$ & RDA $\downarrow$ & ACC $\uparrow$ & RDA $\downarrow$ \\
\midrule
Shallow emb. &60.81 & 28.84 & 54.42 & 32.86 & 65.97 & 19.62 & 74.78 & 16.44 & 83.37 & 6.72 & 83.26 & 5.13 & 75.93 & 10.32 & 75.21 & 10.37 & 72.20 & 9.21 & 24.75 & 63.86 & 50.26 & 21.79 & 60.99 & 12.50 \\

\midrule
TAPE &66.79 & \colorbox[HTML]{B8E7E1}{\textbf{19.64}} & \colorbox[HTML]{D4FAFC}{\underline{72.70}} & \colorbox[HTML]{D4FAFC}{\underline{11.42}} & 73.54 & 10.96 & \colorbox[HTML]{B8E7E1}{\textbf{83.04}} & \colorbox[HTML]{B8E7E1}{\textbf{8.08}} & \colorbox[HTML]{D4FAFC}{\underline{90.45}} & \colorbox[HTML]{B8E7E1}{\textbf{1.17}} & 90.77 & 0.12 & \colorbox[HTML]{D4FAFC}{\underline{81.77}} & \colorbox[HTML]{D4FAFC}{\underline{7.65}} & \colorbox[HTML]{D4FAFC}{\underline{82.89}} & \colorbox[HTML]{D4FAFC}{\underline{5.98}} & 86.03 & \colorbox[HTML]{D4FAFC}{\underline{2.47}} & \colorbox[HTML]{D4FAFC}{\underline{37.40}} & \colorbox[HTML]{D4FAFC}{\underline{51.37}} & \colorbox[HTML]{D4FAFC}{\underline{60.93}} & \colorbox[HTML]{D4FAFC}{\underline{20.11}} & \colorbox[HTML]{B8E7E1}{\textbf{73.64}} & \colorbox[HTML]{D4FAFC}{\underline{8.71}} \\

KEA &\colorbox[HTML]{D4FAFC}{\underline{67.28}} & \colorbox[HTML]{D4FAFC}{\underline{20.94}} & 71.95 & 15.68 & \colorbox[HTML]{B8E7E1}{\textbf{77.33}} & \colorbox[HTML]{D4FAFC}{\underline{10.22}} & 81.22 & 10.10 & 83.91 & 6.12 & 89.48 & 1.42 & 77.94 & 11.64 & 79.44 & 9.35 & 85.96 & 3.60 & 25.26 & 65.70 & 49.51 & 30.79 & 67.97 & 11.65 \\

\midrule
LLaMA &64.71 & 25.42 & 68.85 & 19.45 & 69.64 & 14.11 & 79.82 & 11.86 & 81.21 & 7.69 & 88.84 & 0.96 & 76.52 & 13.25 & 75.96 & 10.39 & 83.00 & 6.41 & 24.46 & 67.85 & 50.81 & 31.27 & 70.38 & 10.22 \\

TE3L &63.91 & 28.70 & 60.04 & 27.37 & 73.80 & 14.96 & 81.54 & \colorbox[HTML]{D4FAFC}{\underline{9.42}} & 84.55 & 6.19 & 89.27 & 2.12 & 78.88 & 11.09 & 77.14 & 12.79 & 85.46 & 5.11 & 29.45 & 60.43 & 53.83 & 29.74 & 67.90 & 10.48 \\

Linq &\colorbox[HTML]{B8E7E1}{\textbf{67.28}} & 23.38 & 64.91 & 22.01 & \colorbox[HTML]{D4FAFC}{\underline{74.03}} & 12.82 & 81.54 & 9.96 & 84.55 & 5.05 & 89.27 & 1.77 & 79.64 & 9.42 & 73.32 & 15.77 & \colorbox[HTML]{D4FAFC}{\underline{86.14}} & 3.95 & 28.13 & 63.25 & 54.79 & 23.38 & 71.45 & 9.17 \\

\midrule
 SimTeg &63.66 & 26.67 & 62.33 & 25.71 & 73.27 & 14.27 & 79.72 & 12.59 & 85.51 & 4.44 & 88.73 & 2.02 & 77.29 & 12.09 & 77.59 & 10.11 & 84.71 & 5.10 & 27.81 & 62.54 & 57.00 & 23.79 & 69.60 & 8.72 \\

E5-Large &66.53 & 23.78 & \colorbox[HTML]{B8E7E1}{\textbf{72.97}} & \colorbox[HTML]{B8E7E1}{\textbf{11.12}} & 72.46 & \colorbox[HTML]{B8E7E1}{\textbf{9.43}} & \colorbox[HTML]{D4FAFC}{\underline{81.64}} & 10.37 & \colorbox[HTML]{B8E7E1}{\textbf{90.77}} & \colorbox[HTML]{D4FAFC}{\underline{1.18}} & \colorbox[HTML]{D4FAFC}{\underline{92.27}} & \colorbox[HTML]{D4FAFC}{\underline{0.12}} & \colorbox[HTML]{B8E7E1}{\textbf{82.15}} & \colorbox[HTML]{B8E7E1}{\textbf{7.47}} & \colorbox[HTML]{B8E7E1}{\textbf{83.70}} & \colorbox[HTML]{B8E7E1}{\textbf{5.56}} & \colorbox[HTML]{B8E7E1}{\textbf{86.79}} & \colorbox[HTML]{B8E7E1}{\textbf{2.19}} & \colorbox[HTML]{B8E7E1}{\textbf{44.16}} & \colorbox[HTML]{B8E7E1}{\textbf{43.22}} & \colorbox[HTML]{B8E7E1}{\textbf{63.71}} & \colorbox[HTML]{B8E7E1}{\textbf{16.04}} & \colorbox[HTML]{D4FAFC}{\underline{72.50}} & \colorbox[HTML]{B8E7E1}{\textbf{6.46}} \\

ModernBert &61.55 & 22.55 & 59.01 & 25.01 & 62.87 & 14.79 & 80.35 & 10.95 & 85.83 & 2.19 & \colorbox[HTML]{B8E7E1}{\textbf{92.28}} & \colorbox[HTML]{B8E7E1}{\textbf{-0.12}} & 76.03 & 13.81 & 75.42 & 10.24 & 86.03 & 2.96 & 32.21 & 55.88 & 57.75 & 22.49 & 70.22 & 9.15 \\

\bottomrule
\end{tabular}}
\end{table*}

\begin{table*}[!t]
\renewcommand\arraystretch{1.2}
\caption{Robustness comparison against the \textbf{NAG-R} attack at a high perturbation level.}
\label{tab:high-nag}
\resizebox{1.0\textwidth}{!}{
\begin{tabular}{ccccccccccccccccccccccccc}
\toprule
\multirow{4}{*}{\textbf{Model}} & \multicolumn{6}{c}{\textbf{Cora}} & \multicolumn{6}{c}{\textbf{Pubmed}} & \multicolumn{6}{c}{\textbf{Ogbn-products}} & \multicolumn{6}{c}{\textbf{Tape-arxiv23}}  \\
\cmidrule(lr){2-7} \cmidrule(lr){8-13} \cmidrule(lr){14-19} \cmidrule(lr){20-25} 
& \multicolumn{2}{c}{GCN} & \multicolumn{2}{c}{GAT} &  \multicolumn{2}{c}{SAGE} & \multicolumn{2}{c}{GCN} & \multicolumn{2}{c}{GAT} &  \multicolumn{2}{c}{SAGE} & \multicolumn{2}{c}{GCN} & \multicolumn{2}{c}{GAT} &  \multicolumn{2}{c}{SAGE} & \multicolumn{2}{c}{GCN} & \multicolumn{2}{c}{GAT} &  \multicolumn{2}{c}{SAGE} \\
\cmidrule(lr){2-3} \cmidrule(lr){4-5} \cmidrule(lr){6-7} \cmidrule(lr){8-9} \cmidrule(lr){10-11} \cmidrule(lr){12-13} \cmidrule(lr){14-15} \cmidrule(lr){16-17} \cmidrule(lr){18-19}\cmidrule(lr){20-21}\cmidrule(lr){22-23}\cmidrule(lr){24-25}
& ACC $\uparrow$ & RDA $\downarrow$  & ACC $\uparrow$ & RDA $\downarrow$  & ACC $\uparrow$ & RDA $\downarrow$  & ACC $\uparrow$ & RDA $\downarrow$ & ACC $\uparrow$ & RDA $\downarrow$ & ACC $\uparrow$ & RDA $\downarrow$ & ACC $\uparrow$ & RDA $\downarrow$ & ACC $\uparrow$ & RDA $\downarrow$ & ACC $\uparrow$ & RDA $\downarrow$ & ACC $\uparrow$ & RDA $\downarrow$ & ACC $\uparrow$ & RDA $\downarrow$ & ACC $\uparrow$ & RDA $\downarrow$ \\
\midrule
Shallow emb. &70.93 & 17.00 & 60.78 & 25.02 & 73.22 & 10.78 & 83.58 & 6.60 & 87.23 & 2.41 & 85.41 & 2.68 & 77.67 & 8.27 & 76.96 & 8.28 & 70.90 & 10.84 & 48.76 & 28.81 & 58.15 & 9.51 & 68.56 & \colorbox[HTML]{B8E7E1}{\textbf{1.64}} \\

\midrule
TAPE &\colorbox[HTML]{B8E7E1}{\textbf{75.36}} & \colorbox[HTML]{B8E7E1}{\textbf{9.32}} & \colorbox[HTML]{B8E7E1}{\textbf{74.77}} & \colorbox[HTML]{B8E7E1}{\textbf{8.89}} & 78.13 & \colorbox[HTML]{D4FAFC}{\underline{5.40}} & \colorbox[HTML]{B8E7E1}{\textbf{91.52}} & \colorbox[HTML]{B8E7E1}{\textbf{-1.31}} & 91.31 & 0.23 & 91.09 & \colorbox[HTML]{D4FAFC}{\underline{-0.23}} & \colorbox[HTML]{D4FAFC}{\underline{86.83}} & \colorbox[HTML]{B8E7E1}{\textbf{1.40}} & \colorbox[HTML]{D4FAFC}{\underline{87.45}} & 0.70 & 87.39 & 0.83 & \colorbox[HTML]{D4FAFC}{\underline{67.36}} & \colorbox[HTML]{D4FAFC}{\underline{12.42}} & \colorbox[HTML]{D4FAFC}{\underline{69.05}} & \colorbox[HTML]{D4FAFC}{\underline{9.47}} & 73.06 & 9.43 \\

KEA &74.55 & 12.40 & \colorbox[HTML]{D4FAFC}{\underline{74.05}} & 13.22 & \colorbox[HTML]{B8E7E1}{\textbf{79.48}} & 7.72 & 88.31 & 2.25 & 87.98 & 1.57 & 90.02 & 0.83 & 86.02 & 2.48 & 85.42 & 2.52 & 88.41 & 0.85 & 55.22 & 25.02 & 61.69 & 13.77 & 69.41 & 9.78 \\

\midrule
LLaMA &73.56 & 15.22 & 71.89 & 15.90 & 76.92 & \colorbox[HTML]{B8E7E1}{\textbf{5.13}} & 87.55 & 3.32 & 85.19 & 3.17 & 90.23 & \colorbox[HTML]{B8E7E1}{\textbf{-0.59}} & 84.63 & 4.06 & 82.95 & 2.15 & 87.07 & 1.82 & 54.61 & 28.21 & 63.27 & 14.42 & \colorbox[HTML]{D4FAFC}{\underline{73.21}} & 6.61 \\

TE3L &\colorbox[HTML]{D4FAFC}{\underline{74.82}} & 16.53 & 71.99 & 12.91 & \colorbox[HTML]{D4FAFC}{\underline{78.69}} & 9.32 & 87.23 & 3.10 & \colorbox[HTML]{D4FAFC}{\underline{91.95}} & \colorbox[HTML]{D4FAFC}{\underline{-2.02}} & 90.98 & 0.24 & 85.59 & 3.53 & 85.60 & 3.22 & \colorbox[HTML]{B8E7E1}{\textbf{89.54}} & \colorbox[HTML]{D4FAFC}{\underline{0.58}} & 57.00 & 23.41 & 62.32 & 18.66 & 72.60 & \colorbox[HTML]{D4FAFC}{\underline{4.28}} \\

Linq &74.28 & 15.41 & 69.85 & 16.08 & 77.65 & 8.56 & 89.59 & 1.07 & 88.09 & 1.08 & 90.56 & 0.35 & 85.64 & 2.59 & 82.18 & 5.59 & \colorbox[HTML]{D4FAFC}{\underline{88.78}} & 1.00 & 59.91 & 21.74 & 56.78 & 20.60 & \colorbox[HTML]{B8E7E1}{\textbf{73.28}} & 6.84 \\

\midrule
SimTeg &74.53 & 14.15 & 72.96 & 13.04 & 78.15 & 8.56 & 88.62 & 2.83 & 90.55 & -1.20 & 89.91 & 0.72 & 84.50 & 3.89 & 85.70 & 0.72 & 87.51 & 1.96 & 56.63 & 23.72 & 66.55 & 11.02 & 69.80 & 8.46 \\

E5-Large &74.80 & 14.31 & 72.73 & 11.41 & 75.32 & 5.85 & \colorbox[HTML]{D4FAFC}{\underline{90.55}} & \colorbox[HTML]{D4FAFC}{\underline{0.59}} & \colorbox[HTML]{B8E7E1}{\textbf{92.28}} & -0.47 & \colorbox[HTML]{B8E7E1}{\textbf{92.49}} & -0.12 & \colorbox[HTML]{B8E7E1}{\textbf{87.35}} & \colorbox[HTML]{D4FAFC}{\underline{1.61}} & \colorbox[HTML]{B8E7E1}{\textbf{88.01}} & \colorbox[HTML]{D4FAFC}{\underline{0.70}} & 88.50 & \colorbox[HTML]{B8E7E1}{\textbf{0.26}} & \colorbox[HTML]{B8E7E1}{\textbf{68.63}} & \colorbox[HTML]{B8E7E1}{\textbf{11.75}} & \colorbox[HTML]{B8E7E1}{\textbf{70.08}} & \colorbox[HTML]{B8E7E1}{\textbf{7.64}} & 72.22 & 6.82 \\
ModernBert &71.11 & \colorbox[HTML]{D4FAFC}{\underline{10.52}} & 71.20 & \colorbox[HTML]{D4FAFC}{\underline{9.52}} & 66.53 & 9.83 & 89.37 & 0.95 & 90.02 & \colorbox[HTML]{B8E7E1}{\textbf{-2.59}} & \colorbox[HTML]{D4FAFC}{\underline{92.38}} & -0.23 & 84.83 & 3.83 & 84.01 & \colorbox[HTML]{B8E7E1}{\textbf{0.01}} & 87.89 & 0.86 & 59.61 & 18.35 & 66.33 & 10.98 & 70.09 & 9.32 \\

\bottomrule
\end{tabular}}
\end{table*}

\section{Results on Clean Graphs}
\label{appendix_clean}
We report the classification accuracy of LLM-enhanced GNN models on four datasets for both targeted nodes and non-targeted nodes in Tables~\ref{tab:targeted-clean} and~\ref{tab:global-clean}, respectively.

\section{More Results on Structural Attacks}
\label{appendix_structural}
Tables~\ref{tab:high-sga} and~\ref{tab:high-nag} present the performance of LLM-enhanced GNNs under SGA~\cite{li2021adversarial} and NAG-R~\cite{zhao2023black} (targeted attacks) at high perturbation levels, respectively. Tables~\ref{tab:high-dice} and~\ref{tab:high-pga} report the performance of LLM-enhanced GNNs under DICE~\cite{waniek2018hiding} and PGA~\cite{zhu2024simple} (non-targeted attacks) at high perturbation levels, respectively. Results for low perturbation levels are omitted due to space constraints and their limited effectiveness.

\begin{table*}[!tbp]
\renewcommand\arraystretch{1.2}
\caption{Robustness comparison against the \textbf{DICE} attack at a high perturbation level.}
\label{tab:high-dice}
\resizebox{1.0\textwidth}{!}{
\begin{tabular}{lcccccccccccccccccccccccc}
\toprule
\multirow{4}{*}{Model} & \multicolumn{6}{c}{\textbf{Cora}} & \multicolumn{6}{c}{\textbf{Pubmed}} & \multicolumn{6}{c}{\textbf{Ogbn-products}} & \multicolumn{6}{c}{\textbf{Tape-arxiv23}}  \\
\cmidrule(lr){2-7} \cmidrule(lr){8-13} \cmidrule(lr){14-19} \cmidrule(lr){20-25} 
& \multicolumn{2}{c}{GCN} & \multicolumn{2}{c}{GAT} &  \multicolumn{2}{c}{SAGE} & \multicolumn{2}{c}{GCN} & \multicolumn{2}{c}{GAT} &  \multicolumn{2}{c}{SAGE} & \multicolumn{2}{c}{GCN} & \multicolumn{2}{c}{GAT} & \multicolumn{2}{c}{SAGE} & \multicolumn{2}{c}{GCN} & \multicolumn{2}{c}{GAT} & \multicolumn{2}{c}{SAGE} \\
\cmidrule(lr){2-3} \cmidrule(lr){4-5} \cmidrule(lr){6-7} \cmidrule(lr){8-9} \cmidrule(lr){10-11} \cmidrule(lr){12-13} \cmidrule(lr){14-15} \cmidrule(lr){16-17} \cmidrule(lr){18-19}\cmidrule(lr){20-21}\cmidrule(lr){22-23}\cmidrule(lr){24-25}
& ACC $\uparrow$ & RDA $\downarrow$  & ACC $\uparrow$ & RDA $\downarrow$  & ACC $\uparrow$ & RDA $\downarrow$  & ACC $\uparrow$ & RDA $\downarrow$ & ACC $\uparrow$ & RDA $\downarrow$ & ACC $\uparrow$ & RDA $\downarrow$ & ACC $\uparrow$ & RDA $\downarrow$ & ACC $\uparrow$ & RDA $\downarrow$ & ACC $\uparrow$ & RDA $\downarrow$ & ACC $\uparrow$ & RDA $\downarrow$ & ACC $\uparrow$ & RDA $\downarrow$ & ACC $\uparrow$ & RDA $\downarrow$ \\
\midrule
Shallow emb. &68.63 & 15.29 & 65.17 & 18.53 & 67.93 & 15.12 & 74.36 & 13.75 & 74.17 & 13.24 & 82.82 & 2.99 & 55.41 & 15.98 & 55.40 & 20.16 & 50.90 & 18.29 & 50.10 & 13.20 & 49.71 & \colorbox[HTML]{B8E7E1}{\textbf{10.38}} & 53.97 & 4.17 \\

\midrule
TAPE &70.82 & 13.36 & 69.28 & 13.92 & 71.78 & 12.27 & \colorbox[HTML]{B8E7E1}{\textbf{85.73}} & \colorbox[HTML]{D4FAFC}{\underline{7.94}} & \colorbox[HTML]{B8E7E1}{\textbf{80.84}} & \colorbox[HTML]{B8E7E1}{\textbf{9.95}} & \colorbox[HTML]{B8E7E1}{\textbf{94.30}} & 0.25 & \colorbox[HTML]{B8E7E1}{\textbf{81.03}} & \colorbox[HTML]{B8E7E1}{\textbf{5.60}} & \colorbox[HTML]{B8E7E1}{\textbf{81.07}} & \colorbox[HTML]{B8E7E1}{\textbf{5.03}} & \colorbox[HTML]{B8E7E1}{\textbf{85.10}} & \colorbox[HTML]{B8E7E1}{\textbf{1.10}} & \colorbox[HTML]{B8E7E1}{\textbf{67.13}} & \colorbox[HTML]{B8E7E1}{\textbf{9.86}} & \colorbox[HTML]{B8E7E1}{\textbf{64.85}} & 12.88 & \colorbox[HTML]{B8E7E1}{\textbf{73.17}} & 1.57 \\

KEA &75.78 & 10.00 & \colorbox[HTML]{B8E7E1}{\textbf{75.07}} & \colorbox[HTML]{D4FAFC}{\underline{11.13}} & \colorbox[HTML]{B8E7E1}{\textbf{78.73}} & \colorbox[HTML]{D4FAFC}{\underline{7.21}} & 78.33 & 11.96 & 77.09 & 10.93 & 89.14 & 0.19 & 77.80 & 7.26 & 76.49 & 8.02 & 80.72 & 4.21 & 59.75 & 13.22 & 54.59 & 21.87 & 67.16 & 3.06 \\

\midrule
LLaMA &72.93 & 10.67 & 69.83 & 14.42 & 73.19 & 9.90 & 77.11 & 12.25 & 72.96 & 14.13 & 88.27 & \colorbox[HTML]{D4FAFC}{\underline{-0.33}} & 77.03 & 7.97 & 71.99 & 10.48 & 80.28 & 3.78 & 63.04 & 11.84 & 56.43 & 19.89 & 69.46 & 2.31 \\

TE3L &74.30 & 11.17 & 70.74 & 14.77 & 77.29 & 7.87 & 78.41 & 11.66 & 76.74 & 11.88 & 89.57 & \colorbox[HTML]{B8E7E1}{\textbf{-0.60}} & 78.15 & 7.05 & 75.74 & 8.56 & 82.35 & 2.60 & 61.25 & 12.71 & 57.75 & 16.06 & 67.75 & 3.02 \\

Linq &\colorbox[HTML]{D4FAFC}{\underline{75.99}} & \colorbox[HTML]{D4FAFC}{\underline{9.85}} & 70.04 & 15.18 & 78.00 & 7.60 & 79.09 & 11.82 & 76.05 & 12.74 & 91.57 & -0.30 & 78.76 & 6.52 & 73.61 & 10.48 & 82.31 & 2.87 & 63.24 & 12.56 & 59.13 & 17.32 & 70.26 & 3.13 \\

\midrule
SimTeg &73.74 & 11.47 & 73.11 & 11.80 & 76.46 & 8.73 & 78.95 & 11.57 & 77.18 & 17.06 & 90.36 & 2.40 & 76.69 & 7.95 & 75.57 & 8.83 & 80.24 & 4.00 & 63.78 & 11.43 & 60.88 & 14.22 & 69.79 & 2.61 \\

E5-Large &\colorbox[HTML]{B8E7E1}{\textbf{76.31}} & \colorbox[HTML]{B8E7E1}{\textbf{9.60}} & \colorbox[HTML]{D4FAFC}{\underline{74.25}} & \colorbox[HTML]{B8E7E1}{\textbf{9.45}} & \colorbox[HTML]{D4FAFC}{\underline{78.37}} & \colorbox[HTML]{B8E7E1}{\textbf{6.47}} & \colorbox[HTML]{D4FAFC}{\underline{85.48}} & 8.15 & \colorbox[HTML]{D4FAFC}{\underline{80.80}} & \colorbox[HTML]{D4FAFC}{\underline{10.07}} & \colorbox[HTML]{D4FAFC}{\underline{94.13}} & 0.01 & \colorbox[HTML]{D4FAFC}{\underline{79.83}} & \colorbox[HTML]{D4FAFC}{\underline{6.04}} & \colorbox[HTML]{D4FAFC}{\underline{78.40}} & \colorbox[HTML]{D4FAFC}{\underline{6.71}} & \colorbox[HTML]{D4FAFC}{\underline{83.18}} & \colorbox[HTML]{D4FAFC}{\underline{2.13}} & \colorbox[HTML]{D4FAFC}{\underline{65.39}} & \colorbox[HTML]{D4FAFC}{\underline{10.52}} & \colorbox[HTML]{D4FAFC}{\underline{62.75}} & \colorbox[HTML]{D4FAFC}{\underline{12.45}} & \colorbox[HTML]{D4FAFC}{\underline{71.61}} & \colorbox[HTML]{D4FAFC}{\underline{1.30}} \\

ModernBert &65.25 & 16.03 & 61.34 & 18.64 & 62.63 & 16.25 & 83.74 & \colorbox[HTML]{B8E7E1}{\textbf{7.33}} & 77.09 & 18.27 & 93.83 & 0.03 & 78.48 & 6.18 & 73.00 & 9.71 & 81.66 & 2.67 & 62.82 & 10.68 & 57.29 & 16.35 & 69.70 & \colorbox[HTML]{B8E7E1}{\textbf{1.09}} \\

\bottomrule
\end{tabular}}
\end{table*}

\begin{table*}[!t]
\renewcommand\arraystretch{1.2}
\caption{Robustness comparison against the \textbf{PGA} attack at a high perturbation level.}
\label{tab:high-pga}
\resizebox{1.0\textwidth}{!}{
\begin{tabular}{ccccccccccccccccccccccccc}
\toprule
\multirow{4}{*}{\textbf{Model}} & \multicolumn{6}{c}{\textbf{Cora}} & \multicolumn{6}{c}{\textbf{Pubmed}} & \multicolumn{6}{c}{\textbf{Ogbn-products}} & \multicolumn{6}{c}{\textbf{Tape-arxiv23}}  \\
\cmidrule(lr){2-7} \cmidrule(lr){8-13} \cmidrule(lr){14-19} \cmidrule(lr){20-25} 
& \multicolumn{2}{c}{GCN} & \multicolumn{2}{c}{GAT} &  \multicolumn{2}{c}{SAGE} & \multicolumn{2}{c}{GCN} & \multicolumn{2}{c}{GAT} &  \multicolumn{2}{c}{SAGE} & \multicolumn{2}{c}{GCN} & \multicolumn{2}{c}{GAT} &  \multicolumn{2}{c}{SAGE} & \multicolumn{2}{c}{GCN} & \multicolumn{2}{c}{GAT} &  \multicolumn{2}{c}{SAGE} \\
\cmidrule(lr){2-3} \cmidrule(lr){4-5} \cmidrule(lr){6-7} \cmidrule(lr){8-9} \cmidrule(lr){10-11} \cmidrule(lr){12-13} \cmidrule(lr){14-15} \cmidrule(lr){16-17} \cmidrule(lr){18-19}\cmidrule(lr){20-21}\cmidrule(lr){22-23}\cmidrule(lr){24-25}
& ACC $\uparrow$ & RDA $\downarrow$  & ACC $\uparrow$ & RDA $\downarrow$  & ACC $\uparrow$ & RDA $\downarrow$  & ACC $\uparrow$ & RDA $\downarrow$ & ACC $\uparrow$ & RDA $\downarrow$ & ACC $\uparrow$ & RDA $\downarrow$ & ACC $\uparrow$ & RDA $\downarrow$ & ACC $\uparrow$ & RDA $\downarrow$ & ACC $\uparrow$ & RDA $\downarrow$ & ACC $\uparrow$ & RDA $\downarrow$ & ACC $\uparrow$ & RDA $\downarrow$ & ACC $\uparrow$ & RDA $\downarrow$ \\
\midrule
Shallow emb. &71.89 & 11.27 & 68.91 & 13.85 & 73.30 & 8.41 & 86.22 & \colorbox[HTML]{D4FAFC}{\underline{-0.01}} & 86.10 & \colorbox[HTML]{D4FAFC}{\underline{-0.71}} & 85.70 & \colorbox[HTML]{D4FAFC}{\underline{-0.39}} & 65.27 & \colorbox[HTML]{B8E7E1}{\textbf{1.03}} & 67.63 & 2.54 & 59.33 & 4.75 & 56.39 & 2.30 & 52.18 & 5.93 & 56.11 & \colorbox[HTML]{B8E7E1}{\textbf{0.37}} \\
\midrule
TAPE &75.71 & 7.38 & 74.37 & \colorbox[HTML]{B8E7E1}{\textbf{7.59}} & 77.32 & 5.50 & \colorbox[HTML]{D4FAFC}{\underline{92.82}} & 0.32 & \colorbox[HTML]{D4FAFC}{\underline{88.34}} & 1.59 & \colorbox[HTML]{B8E7E1}{\textbf{94.49}} & 0.05 & \colorbox[HTML]{B8E7E1}{\textbf{84.57}} & 1.48 & \colorbox[HTML]{D4FAFC}{\underline{84.54}} & \colorbox[HTML]{D4FAFC}{\underline{0.96}} & \colorbox[HTML]{B8E7E1}{\textbf{85.46}} & \colorbox[HTML]{D4FAFC}{\underline{0.69}} & \colorbox[HTML]{B8E7E1}{\textbf{72.89}} & 2.12 & \colorbox[HTML]{B8E7E1}{\textbf{72.20}} & 3.01 & \colorbox[HTML]{B8E7E1}{\textbf{73.12}} & 1.64 \\
KEA &77.14 & 8.38 & \colorbox[HTML]{B8E7E1}{\textbf{76.64}} & 9.27 & \colorbox[HTML]{D4FAFC}{\underline{80.40}} & 5.24 & 88.61 & 0.40 & 87.28 & \colorbox[HTML]{B8E7E1}{\textbf{-0.84}} & 89.32 & -0.01 & 82.55 & 1.60 & 81.95 & 1.46 & 83.32 & 1.13 & 66.92 & 2.80 & 67.40 & 3.54 & 68.36 & \colorbox[HTML]{D4FAFC}{\underline{1.33}} \\
\midrule
LLaMA &75.46 & 7.57 & 70.74 & 13.31 & 76.12 & 6.29 & 87.84 & 0.03 & 85.54 & -0.67 & 88.45 & \colorbox[HTML]{B8E7E1}{\textbf{-0.53}} & 82.15 & 1.85 & 78.71 & 2.13 & 82.43 & 1.20 & 69.62 & 2.64 & \colorbox[HTML]{D4FAFC}{\underline{68.98}} & \colorbox[HTML]{B8E7E1}{\textbf{2.07}} & 68.52 & 3.63 \\
TE3L &77.13 & 7.78 & 73.62 & 11.30 & 79.69 & 5.01 & 88.50 & 0.29 & 87.44 & -0.40 & 88.56 & 0.54 & 82.82 & 1.50 & 82.00 & 1.00 & 83.67 & 1.04 & 68.85 & 1.88 & 65.20 & 5.23 & 67.21 & 3.79 \\
Linq &\colorbox[HTML]{D4FAFC}{\underline{78.18}} & \colorbox[HTML]{D4FAFC}{\underline{7.25}} & 72.55 & 12.14 & \colorbox[HTML]{B8E7E1}{\textbf{80.76}} & \colorbox[HTML]{D4FAFC}{\underline{4.34}} & 89.34 & 0.39 & 87.42 & -0.31 & 90.00 & 1.42 & 83.15 & 1.31 & 80.15 & 2.53 & 84.03 & 0.84 & 69.77 & 3.53 & 68.67 & 3.98 & 70.24 & 3.16 \\
\midrule
SimTeg &76.58 & 8.06 & 75.38 & 9.06 & 79.81 & 4.73 & 88.55 & 0.82 & 87.78 & 5.67 & 90.23 & 2.54 & 81.96 & 1.62 & 81.62 & 1.53 & 82.64 & 1.12 & 70.34 & 2.32 & 68.36 & 3.68 & 70.32 & 1.87 \\
E5-Large &\colorbox[HTML]{B8E7E1}{\textbf{78.55}} & \colorbox[HTML]{B8E7E1}{\textbf{6.94}} & \colorbox[HTML]{D4FAFC}{\underline{75.56}} & \colorbox[HTML]{D4FAFC}{\underline{7.85}} & 80.18 & \colorbox[HTML]{B8E7E1}{\textbf{4.31}} & \colorbox[HTML]{B8E7E1}{\textbf{92.84}} & 0.24 & 87.59 & 2.52 & \colorbox[HTML]{D4FAFC}{\underline{94.27}} & -0.14 & \colorbox[HTML]{D4FAFC}{\underline{83.74}} & 1.44 & \colorbox[HTML]{B8E7E1}{\textbf{85.12}} & \colorbox[HTML]{B8E7E1}{\textbf{-1.29}} & \colorbox[HTML]{D4FAFC}{\underline{84.43}} & \colorbox[HTML]{B8E7E1}{\textbf{0.66}} & \colorbox[HTML]{D4FAFC}{\underline{72.60}} & \colorbox[HTML]{B8E7E1}{\textbf{0.66}} & 68.07 & 5.02 & \colorbox[HTML]{D4FAFC}{\underline{71.53}} & 1.41 \\
ModernBert &71.12 & 8.48 & 67.62 & 10.31 & 69.50 & 7.06 & 92.27 & \colorbox[HTML]{B8E7E1}{\textbf{-2.11}} & \colorbox[HTML]{B8E7E1}{\textbf{88.78}} & 5.87 & 93.81 & 0.05 & 82.60 & \colorbox[HTML]{D4FAFC}{\underline{1.26}} & 78.43 & 2.99 & 83.13 & 0.92 & 69.14 & \colorbox[HTML]{D4FAFC}{\underline{1.69}} & 67.04 & \colorbox[HTML]{D4FAFC}{\underline{2.12}} & 68.92 & 2.20 \\
\bottomrule
\end{tabular}}
\end{table*}

\begin{table*}[!t]
\renewcommand\arraystretch{1.2}
\caption{Robustness comparison against the \textbf{DWord} attack.}
\label{tab:text-dword}
\resizebox{1.0\textwidth}{!}{
\begin{tabular}{ccccccccccccccccccccccccc}
\toprule
\multirow{4}{*}{\textbf{Model}} & \multicolumn{6}{c}{\textbf{Cora}} & \multicolumn{6}{c}{\textbf{Pubmed}} & \multicolumn{6}{c}{\textbf{Ogbn-products}} & \multicolumn{6}{c}{\textbf{Tape-arxiv23}}  \\
\cmidrule(lr){2-7} \cmidrule(lr){8-13} \cmidrule(lr){14-19} \cmidrule(lr){20-25} 
& \multicolumn{2}{c}{GCN} & \multicolumn{2}{c}{GAT} &  \multicolumn{2}{c}{SAGE} & \multicolumn{2}{c}{GCN} & \multicolumn{2}{c}{GAT} &  \multicolumn{2}{c}{SAGE} & \multicolumn{2}{c}{GCN} & \multicolumn{2}{c}{GAT} &  \multicolumn{2}{c}{SAGE} & \multicolumn{2}{c}{GCN} & \multicolumn{2}{c}{GAT} &  \multicolumn{2}{c}{SAGE} \\
\cmidrule(lr){2-3} \cmidrule(lr){4-5} \cmidrule(lr){6-7} \cmidrule(lr){8-9} \cmidrule(lr){10-11} \cmidrule(lr){12-13} \cmidrule(lr){14-15} \cmidrule(lr){16-17} \cmidrule(lr){18-19}\cmidrule(lr){20-21}\cmidrule(lr){22-23}\cmidrule(lr){24-25}
& ACC $\uparrow$ & RDA $\downarrow$  & ACC $\uparrow$ & RDA $\downarrow$  & ACC $\uparrow$ & RDA $\downarrow$  & ACC $\uparrow$ & RDA $\downarrow$ & ACC $\uparrow$ & RDA $\downarrow$ & ACC $\uparrow$ & RDA $\downarrow$ & ACC $\uparrow$ & RDA $\downarrow$ & ACC $\uparrow$ & RDA $\downarrow$ & ACC $\uparrow$ & RDA $\downarrow$ & ACC $\uparrow$ & RDA $\downarrow$ & ACC $\uparrow$ & RDA $\downarrow$ & ACC $\uparrow$ & RDA $\downarrow$ \\
\midrule
Shallow emb. &80.83 & 0.23 & 79.63 & 0.45 & 79.97 & 0.07 & 86.35 & -0.16 & 85.64 & \colorbox[HTML]{D4FAFC}{\underline{-0.18}} & 85.25 & 0.14 & 66.10 & \colorbox[HTML]{B8E7E1}{\textbf{-0.23}} & 68.70 & 0.99 & 62.05 & 0.39 & 57.68 & 0.07 & 55.65 & -0.32 & 56.80 & \colorbox[HTML]{B8E7E1}{\textbf{-0.85}} \\
\midrule
TAPE &81.21 & 0.65 & 80.90 & -0.52 & 81.62 & 0.24 & \colorbox[HTML]{D4FAFC}{\underline{93.03}} & 0.10 & \colorbox[HTML]{B8E7E1}{\textbf{89.82}} & -0.06 & \colorbox[HTML]{B8E7E1}{\textbf{94.46}} & 0.08 & \colorbox[HTML]{B8E7E1}{\textbf{85.86}} & -0.02 & \colorbox[HTML]{B8E7E1}{\textbf{85.70}} & \colorbox[HTML]{B8E7E1}{\textbf{-0.40}} & \colorbox[HTML]{B8E7E1}{\textbf{86.30}} & \colorbox[HTML]{B8E7E1}{\textbf{-0.29}} & \colorbox[HTML]{B8E7E1}{\textbf{74.41}} & 0.08 & \colorbox[HTML]{B8E7E1}{\textbf{74.30}} & 0.19 & \colorbox[HTML]{B8E7E1}{\textbf{74.36}} & \colorbox[HTML]{D4FAFC}{\underline{-0.03}} \\
KEA &84.14 & 0.07 & \colorbox[HTML]{B8E7E1}{\textbf{84.02}} & 0.53 & \colorbox[HTML]{D4FAFC}{\underline{84.43}} & 0.49 & 88.80 & 0.19 & 86.80 & \colorbox[HTML]{B8E7E1}{\textbf{-0.29}} & 89.16 & 0.17 & 83.69 & 0.24 & \colorbox[HTML]{D4FAFC}{\underline{83.49}} & \colorbox[HTML]{D4FAFC}{\underline{-0.40}} & 84.14 & 0.15 & 68.60 & 0.36 & 69.47 & 0.57 & 69.02 & 0.38 \\
\midrule
LLaMA &81.62 & 0.02 & 81.24 & 0.44 & 81.47 & -0.30 & 87.85 & 0.02 & 84.66 & 0.36 & 87.94 & 0.05 & 83.53 & 0.20 & 79.90 & 0.65 & 83.41 & 0.02 & 71.36 & 0.21 & 70.83 & \colorbox[HTML]{D4FAFC}{\underline{-0.55}} & 71.09 & 0.01 \\
TE3L &83.68 & \colorbox[HTML]{D4FAFC}{\underline{-0.05}} & 82.32 & 0.82 & 83.94 & -0.06 & 88.54 & 0.25 & 86.73 & 0.41 & 89.36 & \colorbox[HTML]{D4FAFC}{\underline{-0.36}} & 83.94 & 0.17 & 82.70 & 0.16 & 84.45 & 0.12 & 70.20 & \colorbox[HTML]{D4FAFC}{\underline{-0.04}} & 68.58 & 0.32 & 69.73 & 0.19 \\
Linq &\colorbox[HTML]{B8E7E1}{\textbf{84.32}} & -0.04 & 82.07 & 0.61 & \colorbox[HTML]{B8E7E1}{\textbf{84.51}} & -0.11 & 89.64 & 0.06 & 87.10 & 0.06 & 91.15 & 0.16 & 84.30 & -0.06 & 81.68 & 0.67 & \colorbox[HTML]{D4FAFC}{\underline{84.61}} & 0.15 & 72.33 & -0.01 & \colorbox[HTML]{D4FAFC}{\underline{71.47}} & 0.07 & \colorbox[HTML]{D4FAFC}{\underline{72.41}} & 0.17 \\
\midrule
SimTeg &83.27 & 0.02 & 82.72 & 0.21 & 83.77 & 0.00 & 90.93 & \colorbox[HTML]{D4FAFC}{\underline{-1.85}} & 88.32 & 5.09 & 92.96 & \colorbox[HTML]{B8E7E1}{\textbf{-0.41}} & 83.37 & \colorbox[HTML]{D4FAFC}{\underline{-0.07}} & 82.63 & 0.31 & 83.75 & \colorbox[HTML]{D4FAFC}{\underline{-0.20}} & 72.10 & \colorbox[HTML]{B8E7E1}{\textbf{-0.12}} & 71.14 & -0.24 & 71.62 & 0.06 \\
E5-Large &\colorbox[HTML]{D4FAFC}{\underline{84.17}} & 0.28 & \colorbox[HTML]{D4FAFC}{\underline{82.84}} & \colorbox[HTML]{D4FAFC}{\underline{-1.02}} & 84.28 & \colorbox[HTML]{D4FAFC}{\underline{-0.58}} & \colorbox[HTML]{B8E7E1}{\textbf{93.14}} & -0.09 & \colorbox[HTML]{D4FAFC}{\underline{89.70}} & 0.17 & \colorbox[HTML]{D4FAFC}{\underline{94.35}} & -0.22 & \colorbox[HTML]{D4FAFC}{\underline{84.61}} & 0.41 & 83.15 & 1.06 & 84.36 & 0.74 & \colorbox[HTML]{D4FAFC}{\underline{72.83}} & 0.34 & 71.36 & 0.43 & 72.39 & 0.22 \\
ModernBert &78.61 & \colorbox[HTML]{B8E7E1}{\textbf{-1.16}} & 76.23 & \colorbox[HTML]{B8E7E1}{\textbf{-1.11}} & 77.35 & \colorbox[HTML]{B8E7E1}{\textbf{-3.44}} & 92.40 & \colorbox[HTML]{B8E7E1}{\textbf{-2.26}} & 86.53 & 8.26 & 93.96 & -0.11 & 83.58 & 0.08 & 80.73 & 0.15 & 83.23 & 0.80 & 69.76 & 0.81 & 69.10 & \colorbox[HTML]{B8E7E1}{\textbf{-0.89}} & 70.30 & 0.24 \\
\bottomrule
\end{tabular}}
\end{table*}

\begin{table*}[!t]
\renewcommand\arraystretch{1.2}
\caption{Robustness comparison against the \textbf{BertAtk} attack.}
\label{tab:text-bertatk}
\resizebox{1.0\textwidth}{!}{
\begin{tabular}{ccccccccccccccccccccccccc}
\toprule
\multirow{4}{*}{\textbf{Model}} & \multicolumn{6}{c}{\textbf{Cora}} & \multicolumn{6}{c}{\textbf{Pubmed}} & \multicolumn{6}{c}{\textbf{Ogbn-products}} & \multicolumn{6}{c}{\textbf{Tape-arxiv23}}  \\
\cmidrule(lr){2-7} \cmidrule(lr){8-13} \cmidrule(lr){14-19} \cmidrule(lr){20-25} 
& \multicolumn{2}{c}{GCN} & \multicolumn{2}{c}{GAT} &  \multicolumn{2}{c}{SAGE} & \multicolumn{2}{c}{GCN} & \multicolumn{2}{c}{GAT} &  \multicolumn{2}{c}{SAGE} & \multicolumn{2}{c}{GCN} & \multicolumn{2}{c}{GAT} &  \multicolumn{2}{c}{SAGE} & \multicolumn{2}{c}{GCN} & \multicolumn{2}{c}{GAT} &  \multicolumn{2}{c}{SAGE} \\
\cmidrule(lr){2-3} \cmidrule(lr){4-5} \cmidrule(lr){6-7} \cmidrule(lr){8-9} \cmidrule(lr){10-11} \cmidrule(lr){12-13} \cmidrule(lr){14-15} \cmidrule(lr){16-17} \cmidrule(lr){18-19}\cmidrule(lr){20-21}\cmidrule(lr){22-23}\cmidrule(lr){24-25}
& ACC $\uparrow$ & RDA $\downarrow$  & ACC $\uparrow$ & RDA $\downarrow$  & ACC $\uparrow$ & RDA $\downarrow$  & ACC $\uparrow$ & RDA $\downarrow$ & ACC $\uparrow$ & RDA $\downarrow$ & ACC $\uparrow$ & RDA $\downarrow$ & ACC $\uparrow$ & RDA $\downarrow$ & ACC $\uparrow$ & RDA $\downarrow$ & ACC $\uparrow$ & RDA $\downarrow$ & ACC $\uparrow$ & RDA $\downarrow$ & ACC $\uparrow$ & RDA $\downarrow$ & ACC $\uparrow$ & RDA $\downarrow$ \\
\midrule
Shallow emb. &78.46 & 3.16 & 78.74 & 1.56 & 78.19 & 2.30 & 86.01 & \colorbox[HTML]{D4FAFC}{\underline{0.23}} & 85.50 & \colorbox[HTML]{D4FAFC}{\underline{-0.01}} & 84.78 & 0.69 & 64.37 & 2.40 & 67.75 & 2.36 & 59.83 & 3.95 & 55.97 & 3.03 & 53.78 & 3.05 & 53.58 & 4.87 \\
\midrule
TAPE &81.81 & \colorbox[HTML]{D4FAFC}{\underline{-0.09}} & 80.29 & 0.24 & 80.21 & 1.97 & \colorbox[HTML]{D4FAFC}{\underline{92.33}} & 0.85 & \colorbox[HTML]{D4FAFC}{\underline{89.18}} & 0.66 & \colorbox[HTML]{D4FAFC}{\underline{93.26}} & 1.35 & \colorbox[HTML]{B8E7E1}{\textbf{85.63}} & 0.24 & \colorbox[HTML]{B8E7E1}{\textbf{85.36}} & \colorbox[HTML]{D4FAFC}{\underline{0.00}} & \colorbox[HTML]{B8E7E1}{\textbf{85.43}} & 0.72 & \colorbox[HTML]{B8E7E1}{\textbf{73.05}} & 1.91 & \colorbox[HTML]{B8E7E1}{\textbf{73.29}} & \colorbox[HTML]{D4FAFC}{\underline{1.54}} & \colorbox[HTML]{B8E7E1}{\textbf{73.13}} & 1.63 \\
KEA &\colorbox[HTML]{B8E7E1}{\textbf{84.35}} & \colorbox[HTML]{B8E7E1}{\textbf{-0.18}} & \colorbox[HTML]{B8E7E1}{\textbf{83.99}} & 0.57 & \colorbox[HTML]{D4FAFC}{\underline{83.86}} & 1.17 & 88.46 & 0.57 & 86.53 & 0.02 & 87.65 & 1.86 & 83.73 & 0.19 & 83.11 & 0.06 & 83.78 & \colorbox[HTML]{D4FAFC}{\underline{0.58}} & 68.03 & \colorbox[HTML]{D4FAFC}{\underline{1.19}} & 66.16 & 5.31 & 68.29 & \colorbox[HTML]{D4FAFC}{\underline{1.43}} \\
\midrule
LLaMA &80.62 & 1.25 & 81.16 & 0.54 & 79.67 & 1.92 & 87.04 & 0.94 & 84.33 & 0.75 & 86.36 & 1.84 & 83.11 & 0.70 & 81.06 & \colorbox[HTML]{B8E7E1}{\textbf{-0.80}} & 82.57 & 1.03 & 69.33 & 3.05 & 69.30 & 1.62 & 69.06 & 2.87 \\
TE3L &83.28 & 0.43 & 80.78 & 2.67 & 82.98 & 1.08 & 88.11 & 0.73 & 86.76 & 0.38 & 87.41 & 1.83 & 83.74 & 0.40 & 82.65 & 0.22 & 83.96 & 0.70 & 68.25 & 2.74 & 66.04 & 4.01 & 67.90 & 2.81 \\
Linq &\colorbox[HTML]{D4FAFC}{\underline{84.08}} & 0.25 & 82.66 & \colorbox[HTML]{D4FAFC}{\underline{-0.11}} & \colorbox[HTML]{B8E7E1}{\textbf{84.04}} & \colorbox[HTML]{B8E7E1}{\textbf{0.45}} & 89.20 & 0.55 & 87.10 & 0.06 & 89.59 & 1.87 & \colorbox[HTML]{D4FAFC}{\underline{84.28}} & \colorbox[HTML]{D4FAFC}{\underline{-0.04}} & 81.83 & 0.49 & \colorbox[HTML]{D4FAFC}{\underline{84.10}} & 0.76 & 71.16 & 1.60 & 70.37 & 1.61 & 70.72 & 2.50 \\
\midrule
SimTeg &83.28 & 0.01 & \colorbox[HTML]{D4FAFC}{\underline{82.96}} & -0.08 & 83.30 & \colorbox[HTML]{D4FAFC}{\underline{0.56}} & 91.02 & \colorbox[HTML]{B8E7E1}{\textbf{-1.95}} & 88.05 & 5.38 & 92.92 & \colorbox[HTML]{B8E7E1}{\textbf{-0.37}} & 83.48 & \colorbox[HTML]{B8E7E1}{\textbf{-0.20}} & 82.78 & 0.13 & 83.68 & \colorbox[HTML]{B8E7E1}{\textbf{-0.12}} & \colorbox[HTML]{D4FAFC}{\underline{72.18}} & \colorbox[HTML]{B8E7E1}{\textbf{-0.24}} & \colorbox[HTML]{D4FAFC}{\underline{70.79}} & \colorbox[HTML]{B8E7E1}{\textbf{0.25}} & \colorbox[HTML]{D4FAFC}{\underline{71.77}} & \colorbox[HTML]{B8E7E1}{\textbf{-0.15}} \\
E5-Large &83.60 & 0.96 & 82.76 & \colorbox[HTML]{B8E7E1}{\textbf{-0.93}} & 83.11 & 0.81 & \colorbox[HTML]{B8E7E1}{\textbf{92.77}} & 0.31 & \colorbox[HTML]{B8E7E1}{\textbf{89.91}} & \colorbox[HTML]{B8E7E1}{\textbf{-0.07}} & \colorbox[HTML]{B8E7E1}{\textbf{94.09}} & \colorbox[HTML]{D4FAFC}{\underline{0.05}} & 84.13 & 0.98 & \colorbox[HTML]{D4FAFC}{\underline{83.52}} & 0.62 & 83.77 & 1.44 & 71.93 & 1.57 & 69.99 & 2.34 & 71.43 & 1.54 \\
ModernBert &74.69 & 3.89 & 69.86 & 7.34 & 72.82 & 2.62 & 90.03 & 0.37 & 84.23 & 10.70 & 88.82 & 5.37 & 82.81 & 1.00 & 78.57 & 2.82 & 82.67 & 1.47 & 68.67 & 2.36 & 66.63 & 2.72 & 68.26 & 3.14 \\
\bottomrule
\end{tabular}}
\end{table*}

\section{More Results on Textual Attacks}
\label{appendix_textual}
\begin{figure}[!t]
  \centering
  \captionsetup[sub]{aboveskip=0pt,belowskip=0pt,skip=4pt}
  \subfloat[ACC]{\includegraphics[width=0.5\columnwidth]{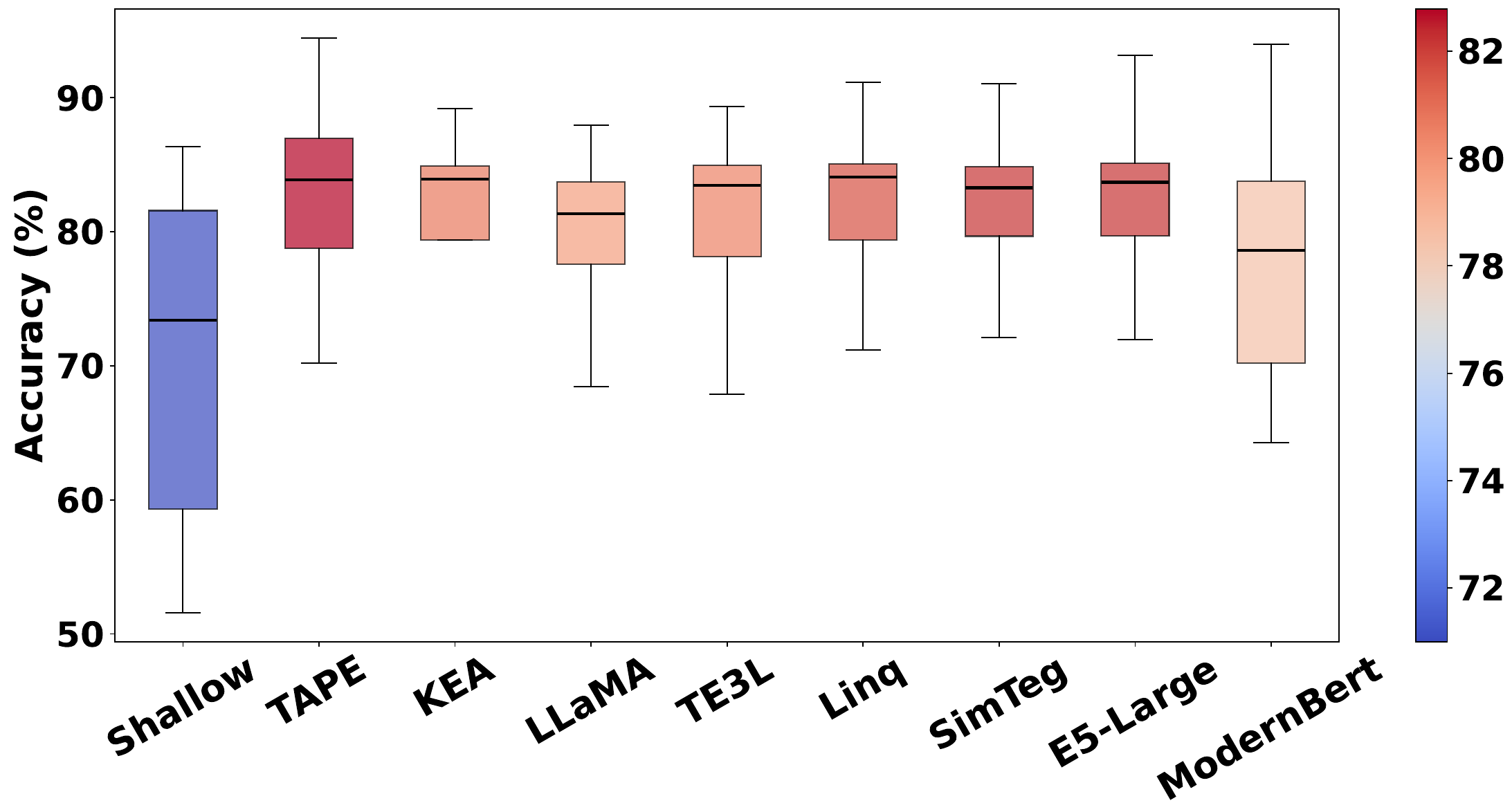}
\label{fig:textaccbox}}
 \subfloat[RDA]{\includegraphics[width=0.5\columnwidth]{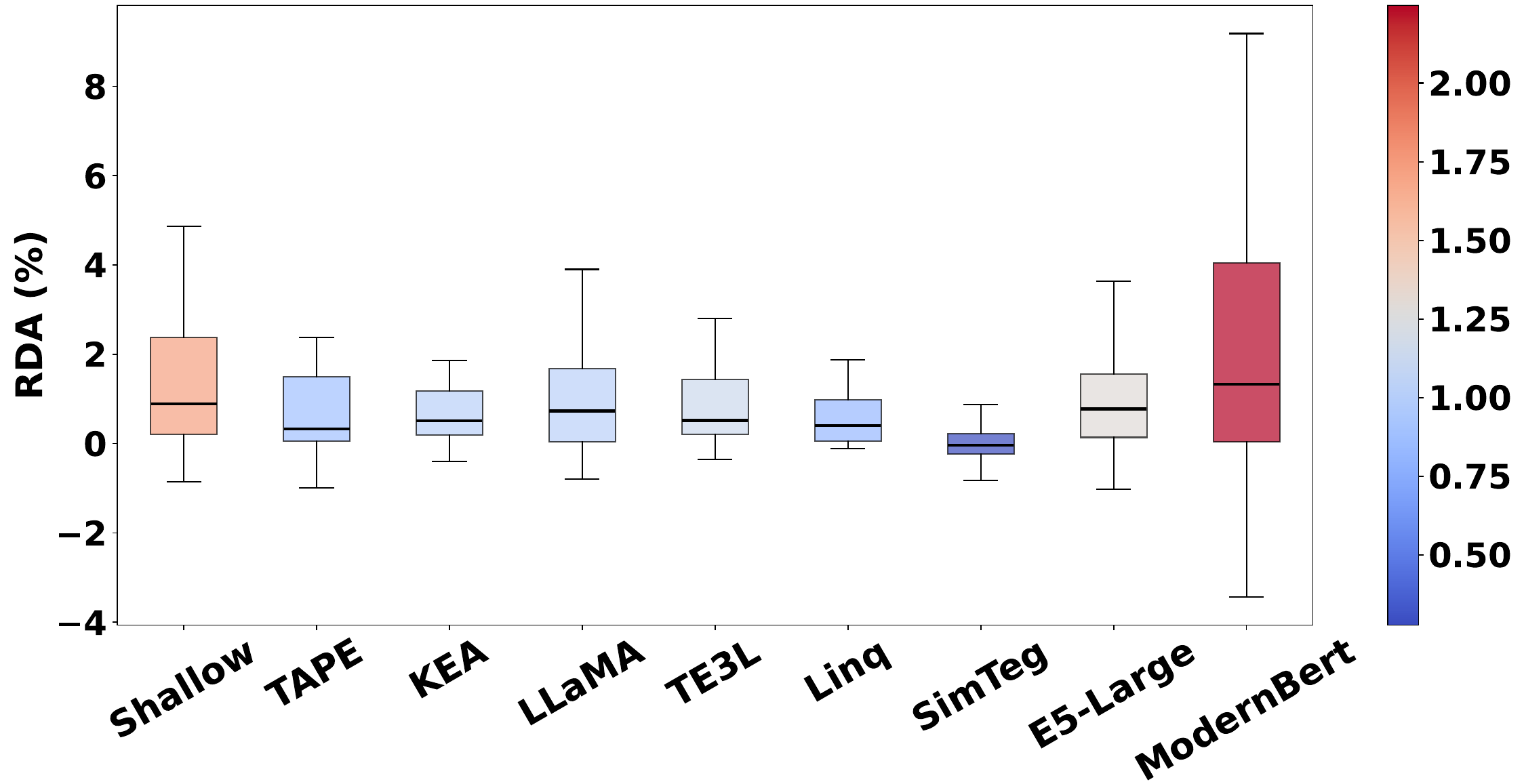}
\label{fig:textrdabox}}
\caption{\label{fig:boxplots_textual}ACC and RDA under the textual attacks.}
\vspace{-2mm}
\end{figure}

Tables~\ref{tab:text-dword} and~\ref{tab:text-bertatk} report the performance of LLM-enhanced GNNs under DWord~\cite{gao2018black} and BertAtk~\cite{li2020bert}, respectively. Fig.~\ref{fig:boxplots_textual} compares the robustness of different models under three textual attacks. 

\begin{table}[!t]
\renewcommand\arraystretch{1.2}
\caption{Results of certified robustness.}
\label{tab:cert-robust}
\resizebox{1.0\linewidth}{!}{
\begin{tabular}{ccccccccc}
\toprule
\multirow{2}{*}{\textbf{Model}} & \multicolumn{2}{c}{\textbf{Cora}} & \multicolumn{2}{c}{\textbf{Pubmed}} & \multicolumn{2}{c}{\textbf{Ogbn-products}} & \multicolumn{2}{c}{\textbf{Tape-arxiv23}}  \\
\cmidrule(lr){2-3} \cmidrule(lr){4-5} \cmidrule(lr){6-7} \cmidrule(lr){8-9}
& CA $\uparrow$ & MCR $\uparrow$ & CA $\uparrow$ & MCR $\uparrow$ & CA $\uparrow$ & MCR $\uparrow$ & CA $\uparrow$ & MCR $\uparrow$ \\
\midrule
Shallow emb. &51.36 &1.74 &34.23 &0.97 &34.23 &0.86 &28.23 &0.93 \\
TAPE &\colorbox[HTML]{B8E7E1}{\textbf{64.47}} &\colorbox[HTML]{D4FAFC}{\underline{2.33}} &\colorbox[HTML]{B8E7E1}{\textbf{68.87}} &\colorbox[HTML]{D4FAFC}{\underline{1.90}} &\colorbox[HTML]{B8E7E1}{\textbf{73.44}} &\colorbox[HTML]{D4FAFC}{\underline{2.01}} & \colorbox[HTML]{B8E7E1}{\textbf{54.24}} &\colorbox[HTML]{D4FAFC}{\underline{1.08}} \\
TE3L &60.08 &2.05 &41.74 &1.07 &65.39 &1.97 &34.36 &1.01 \\
E5-Large &\colorbox[HTML]{D4FAFC}{\underline{63.54}} &\colorbox[HTML]{B8E7E1}{\textbf{2.36}} &\colorbox[HTML]{D4FAFC}{\underline{68.16}} &\colorbox[HTML]{B8E7E1}{\textbf{2.17}} &\colorbox[HTML]{D4FAFC}{\underline{70.31}} &\colorbox[HTML]{B8E7E1}{\textbf{2.09}} &\colorbox[HTML]{D4FAFC}{\underline{46.47}} &\colorbox[HTML]{B8E7E1}{\textbf{1.22}} \\
\bottomrule
\end{tabular}}
\end{table}

\section{Certified Robustness}
\label{appendix_certify}

To assess certified robustness after structural poisoning attacks, we evaluated three types of LLM-enhanced GNNs using the randomized smoothing framework~\cite{bojchevski2020efficient} under high-perturbation Mettack poisoning. Certified lower bounds were computed via the Clopper-Pearson confidence interval (95\% confidence, 10,000 Monte Carlo samples). For each sample, we generated a perturbed graph by independently deleting each existing edge with probability $p_{\text{del}}=0.4$, without introducing spurious edges ($p_{\text{add}}=0$). We report \textbf{Certified Accuracy (CA)}, the proportion of correctly classified nodes whose predictions are provably robust to adversarial edge deletions at radius $r \ge 1$, and \textbf{Mean Certified Radius (MCR)}, the average maximum number of edge deletions that can be tolerated without changing the prediction. As shown in Table~\ref{tab:cert-robust}, LLM-enhanced models consistently outperform the shallow embedding-based baseline across all datasets in terms of CA and MCR. These results indicate that LLM/LM-generated embeddings not only improve empirical robustness but also strengthen certified robustness after structural poisoning attacks.
\end{document}